\documentclass[10pt]{article}
\usepackage[utf8]{inputenc}
\usepackage{fancyhdr}
\usepackage{extramarks}
\usepackage{amsmath}
\usepackage{amsthm}
\usepackage{amsfonts}
\usepackage{siunitx}
 \usepackage{tikz-cd}
\usepackage{tikz}
\usepackage{stmaryrd}
\usepackage[plain]{algorithm}
\usepackage{algpseudocode}
\usepackage{multirow}
\usepackage{booktabs}
\usepackage{palatino}
\usepackage{graphicx}
\usepackage{subcaption}
\usepackage[colorlinks,linkcolor=black,anchorcolor=black,citecolor=black,urlcolor=blue]{hyperref}
\usepackage{soul}
\usepackage{amsmath,bm}
\usepackage{booktabs}
\usepackage{mathtools}
\usepackage{amssymb}
\usepackage{caption}
\usepackage{capt-of}
\usepackage{mciteplus}
\usepackage{cite}
\usepackage{mathrsfs}
\usepackage[title,titletoc,toc]{appendix}
\usepackage{xr}
\usepackage{parskip}
\usepackage{textcomp}
\usepackage[colaction]{multicol}
\usepackage[switch]{lineno}
\usepackage{lipsum}
\usepackage{etoolbox}
\usepackage{longtable}
\usepackage{array}
\usepackage{tablefootnote}
\usepackage[T1]{fontenc}

\usepackage{algpseudocode}

\newcolumntype{C}[1]{>{\centering\arraybackslash}p{#1}}
\usetikzlibrary{automata,positioning}
\DeclareMathOperator{\im}{im}

\usetikzlibrary{automata,positioning}

\newcommand{\cplx}{\mathbb C}

\begin{document}

\title{
CCP: Correlated Clustering and Projection for Dimensionality Reduction
}
\author{Yuta Hozumi$^1$, Rui Wang$^1$, and Guo-Wei Wei$^{1,2,3}$\footnote{
		Corresponding author.		E-mail: weig@msu.edu} \\
	$^1$ Department of Mathematics, \\
	Michigan State University, MI 48824, USA.\\
	$^2$ Department of Electrical and Computer Engineering,\\
	Michigan State University, MI 48824, USA. \\
	$^3$ Department of Biochemistry and Molecular Biology,\\
	Michigan State University, MI 48824, USA. \\
}

\date{} 

\maketitle
\begin{abstract}
Most dimensionality reduction methods employ frequency domain representations obtained from matrix diagonalization and may not be efficient for large datasets with relatively high intrinsic dimensions. To address this challenge, Correlated Clustering and Projection (CCP) offers a novel data domain strategy that does not need to solve any matrix. CCP partitions high-dimensional features into correlated clusters and then projects correlated features in each cluster into a one-dimensional representation based on sample correlations. Residue-Similarity (R-S) scores and indexes, the shape of data in Riemannian manifolds, and algebraic topology-based persistent Laplacian are introduced for visualization and analysis. Proposed methods are validated with benchmark datasets associated with various machine learning algorithms.
 \end{abstract}


\pagebreak
%
 {\setcounter{tocdepth}{4} \tableofcontents}
 \newpage

\setcounter{page}{1}
\renewcommand{\thepage}{{\arabic{page}}}
\section{Introduction}
Technological advances have fueled exponential growth in high-dimensional data. In biological science, high dimensional data are ubiquitous in genomics, epigenomics, transcriptomics, proteomics, metabolomics, and phenomics. For example, the sample dimension in single-cell RNA-Seq analysis is the number of genes \cite{luecken2019current}, which can be very large. In image science, an image of  moderate size, i.e., $1024\times1024$, gives rise to a 1,048,576-dimensional vector.  
The rapid increase in the size and complexity of scientific data has made the problem of the ``curse of dimensionality" more challenging than ever before in data sciences \cite{bellman1966dynamic}. In machine learning, this problem is associated with the phenomenon that the average predictive power of a well-trained model first increases as the feature size increases but starts to deteriorate beyond a certain dimensionality \cite{trunk1979problem, chandrasekaran1974quantization}. Moreover, data with an enormous volume of the feature space will become sparse, which is problematic for the statistical analysis in finding the statistical significance and principal variables. Furthermore, it is challenging to visualize data in high dimensions unless one can reduce the dimension to two or three. Therefore, it is desirable to reduce the dimensionality of high-dimensional data for the sake of prediction, analysis, and visualization. These challenges have been driving the development of many dimensionality reduction (DR) methods that can capture the intrinsic correlations in the original data in a low-dimensional representation \cite{murphy2012machine}. 

 Dimensionality reduction can be achieved through various deep neural networks (DNN), such as graph neural networks, autoencoders, transformers, etc. However, most DNN methods may not work well with excessively high-dimensional data. Commonly used dimensionality reduction algorithms fall into two categories: linear and nonlinear with respect to a certain distance metric. Principle component analysis (PCA) \cite{dunteman1989principal} is a basic linear DR algorithm that focuses on finding the principal components by creating new uncorrelated variables that successively maximize variances \cite{jolliffe2016principal}. Specifically, the first principal component is a vector that maximizes the variance of the projected data, while the $i$th principal component is a vector that is orthogonal to the first $(i-1)$ principal components, leading to the maximization of the variance of the projected data. Linear discriminant analysis (LDA) is another linear DR method proposed by Sir Ronald Fisher in 1936 \cite{fisher1936use}. As a generalization of Fisher's linear discriminant, LDA aims to find a linear combination of features that maximizes the separability of classes and minimizes the inter-class variance for the multi-class classification problem \cite{xanthopoulos2013linear}. Another category of dimensionality reduction methods contains many nonlinear algorithms, which can be classified into two groups: those that favor the preservation of the global pairwise distance and those that seek to retain local distance instead of global distance. Algorithms such as kernel principal component analysis (kernel PCA), Sammon mapping, and spectral embedding fall within the former category, while Isomap, LargeVis, Laplacian eigenmaps, locally linear embedding (LLE), diffusion maps \cite{coifman2005geometric,dos2022grassmannian}, t-SNE, and UMAP fall into the latter category. Kernel PCA \cite{scholkopf1998nonlinear} is an extension of PCA. Standard PCA typically has poor performance if the data has complicated algebraic structures that cannot be well-represented in a linear space. Therefore, kernel PCA is designed by applying the kernel functions in a reproducing kernel Hilbert space in 1998. In 1969, John W. Sammon first proposed the Sammon mapping  \cite{sammon1969nonlinear}, which aims to conserve the structure of inter-point distances by minimizing Sammon's error and attempts to ensure the mapping does not affect the underlying topology \cite{henderson1997sammon}. Spectral embedding computes the full Laplacian graph and uses graph eigenvectors, which allows for the preservation of the original global graph structure in the lower dimensional space. Although kernel PCA, Sammon mapping, and spectral embedding preserve the pairwise distance structure amongst all the data, they fail to capture the local relationship between data points. Therefore, nonlinear algorithms are essential to incorporate the local structure in low-dimensional space and better describe the local information of original data.  
%

A quantitative survey of dimensionality reduction techniques is given in Ref. \cite{espadoto2019toward}. Several widely used nonlinear DR algorithms are briefly discussed in the following. Isomap \cite{tenenbaum2000global} is a nonlinear method that aims to preserve the geodesic distance between samples while reducing the dimension. Isomap is actually an extension of multidimensional scaling (MDS) \cite{mead1992review}, which replaces the Euclidean distance in MDS with geodesic distance (estimated by Dijkstra's distance in graph theory). Moreover, Isomap is a local method as it estimates the intrinsic geometry of a data manifold by roughly estimating each sample’s neighbors, which ensures its efficiency \cite{anowar2021conceptual}. Laplacian Eigenmap (LE), introduced in 2003 \cite{belkin2003laplacian}, is another unsupervised-nonlinear algorithm that looks for low-dimensional representation by maintaining the local properties of a weighted graph Laplacian. The reduction procedure of LE is as follows: It first constructs a neighborhood graph where each data point is linked to its nearest neighbors. Then, the weight of each edge is estimated on the Gaussian kernel function. After solving the eigenvectors of the matrix generalized by the weighted neighborhood graph, one leaves out the eigenvectors associated with  0 eigenvalues and uses the subsequent $k$ eigenvectors (smallest) for embedding in $k$-dimensional space. Moreover, t-Distributed Stochastic Neighbor Embedding (t-SNE) \cite{hinton2002stochastic,van2008visualizing} is a nonlinear, manifold-based method for dimensionality reduction, which is well suited for reducing high dimensional data into  a two- or three-dimensional space for visualization. Based on stochastic neighbor embedding (SNE), t-SNE first represents similarities for every pair of data by constructing conditional probability distribution over pairs of data. Afterward, the `student t-distribution'' is applied to obtain the probability distribution in the embedded space. By minimizing the Kullback-Leibler (KL) divergence between these two probabilities in the original and embedded space, t-SNE preserves the significant structure of the data and is accessible for analyzing and visualizing high-dimensional data \cite{li2019survey}. Furthermore, a  state-of-art nonlinear dimensionality reduction algorithm is uniform manifold approximation and projection (UMAP) \cite{mcinnes2018umap}, a graph-based algorithm that builds on the Laplacian eigenmaps and performs great visualization and feature extraction. Three assumptions make UMAP stand out among the other dimensionality reduction algorithms: (1) Data is uniformly distributed on a Riemann manifold, (2) Riemannian metric is locally constant, and  (3) The manifold is locally connected. UMAP creates $k$-dimensional weighted graph representations based on the $k$-nearest neighborhood searching and intents to minimize the edge-wise cross-entropy between the embedded low-dimensional weighted graph representation in teams of a fuzzy set cross-entropy loss function via the stochastic gradient descent. Specifically, UMAP constructs a weighted directed adjacency matrix $A$, where $A(i,j)$ represents the connection between the $i$th node and the $j$th node when the $j$th node is one of the $k$ nearest neighbors. Next, a normalized sparse Laplacian matrix can be derived from $A$ with the implementation of the cross-entropy loss involved, and the $k$-dimensional eigenvectors of this normalized Laplacian will be used to represent each of the original data points in a low-dimension space. 

All of the dimensionality reduction algorithms mentioned above have broad applications in science and technology. Howbeit, they depend on frequency domain representations obtained from matrix diagonalization.  Generally speaking, the computational complexity of the eigenvalue decomposition for a full matrix is $O(M^3)$ if the number of samples is $M$, which forms an $M\times M$ matrix. Fast solvers are available but render low accuracy for datasets with relatively high intrinsic dimensions \cite{espadoto2019toward}.  In addition, for data with a large number of features $I$, where $M << I$, the dependence of matrix diagonalization limits their performance. Moreover, most methods rely on computing the distance between data entries (samples), which might be problematic in a high dimension. Especially, for methods exploiting nearest neighbors, such as UMAP and t-SNE, comparing samples by distance may result in instability for datasets with moderately high intrinsic dimensions, which was outlined in the ``curse of dimensionality''\cite{bellman1966dynamic}. 

A somewhat related but different problem is tensor-based dimensionality reduction
\cite{chen2009pca,wang2008tensor}. It involves data with certain internal structures of geometric,  topological, algebraic, and/or physical origins. Methods dealing with tonsorial structures, such as  Tucker decomposition, are often used \cite{tucker1966some,li2016mr},  in addition to the aforementioned dimensionality reduction approaches. These methods are often used for videos, X-ray Computed Tomography (X-ray CT), and Magnetic Resonance Image (MRI) data. 

Other related issues concern feature evaluation, ranking,  clustering, extraction, and selection for unlabeled data. Feature evolution and ranking can be performed through filtering or embedding, while feature clustering and selection can be carried out by $K$-means, $K$-means++, $K$-medoids, etc. These methods can be used for the pre-processing of dimensionality reduction. For labeled data, various supervised learning methods can be used for feature selection or extraction. 

In this work,  we propose a two-step data-domain method that seeks an optimal clustering in terms of a distance describing intrinsic feature correlations among samples to divide  $I$ feature vectors into $N$ correlated clusters and then, non-linearly project correlated features in each cluster into a single descriptor by using Flexibility Rigidity Index (FRI) \cite{xia2013multiscale}, which results in  a    low-dimensional representation of the original data.  Additionally, the complex global correlations among samples are embedded into samples' local representations during the FRI-based nonlinear projection $\mathbb{R}^{I}\rightarrow \mathbb{R}^{N}$. To gain computational efficiency, one may further compute the pairwise correlation matrix of samples and impose a cutoff distance to avoid the global summation during the projection. 
The resulting method, called Correlated Clustering and Projection (CCP),  precedes the other DR algorithms in the following aspects. 
(1) Instead of solving a  matrix to reduce the dimensionality, CCP does not involve matrix diagonalization and thus can handle the dimensionality reduction of large sample sizes.  
(2) CCP exploits statistical measures, such as (distance) covariances, to quantify the high-level dependence between random feature vectors, rendering a stable algorithm  when dealing with high dimensional data.  
(3) CCP is flexible with respect to targeted dimension $N$ because the partition of features is based on $N$, whereas other methods may rely on $\min(M, N)$. The performance of CCP is stable with respect to the increase of $N$, which is important for datasets with high or moderately high intrinsic dimensions. In contrast, many existing methods stop working when the intrinsic data dimension is moderately high.   
(4) CCP is stable with respect to subsampling, which allows continuously adding new samples into a pre-existing dataset without the need to restart the calculation from the very beginning and thus, is advantageous for  continuous  data acquisition, collection, and analysis. This capability is valuable when the transient data are too expensive to be kept permanently, e.g., molecular dynamics simulations. Additionally, this  subsampling property enables parameter optimization using a small amount of data in case of large data size.  
(5) As a data-domain method, CCP can be combined with a frequency-domain method, such as UMAP or t-SNE, for a secondary dimensionality reduction to better preserve global structures of data and achieve higher accuracy.    
(6) Finally, the performance of CCP is validated on several benchmark classification datasets: Leukemia, Carcinoma, ALL-AML,  TCGA-PANCAN, 
Coil-20,  Coil-100, and Smallnorb based on various  traditional algorithms such as  $k$-NN, support vector machine, random forest and gradient boost decision tree. In all cases, CCP is very competitive with the state-of-art algorithms. 

Additionally, we have also proposed a new method, called Residue-Similarity (R-S) scores or R-S plot, for the performance visualization of unsupervised clustering and  supervised classification algorithms. Although Receiving Operation Characteristic curve (ROC) and Area Under the ROC Curve (AUC) are typically used for the performance visualization of binary classes, they are not convenient for multiple classes. The proposed R-S scores can be used for visualizing the performance in an arbitrary number of classes.  Finally, R index, S index,  R-S disparity, and total R-S index are proposed to characterize clustering and classification results.    

 Recent years have witnessed the growth of Topological Data Analysis (TDA) via persistent homology \cite{frosini1992measuring,edelsbrunner2000topological,zomorodian2005computing,carlsson2009topology, mischaikow2013morse,KLXia:2014c,townsend2020representation} in data sciences. It can be used to analyze the topological invariant of the R-S scores. However,  persistent homology is insensitive to the homotopic shape evolution of data during filtration. 
We introduce a topological Laplacian,  Persistent Spectral Graph (PSG)    \cite{wang2020persistent}, to capture the homotopic shape of data, in addition to topological invariants. Note that TDA and PSG are dimensionality reduction algorithms that can generate low dimensional representations of the original high-dimensional data \cite{chen2022persistent,nguyen2020review}. 

To further analyze the shape of data,  we transform point cloud data into a Grassmann manifold representation by using FRI \cite{xia2013multiscale}. When $N=3$, the 2-manifold shape of data can be directly visualized.  
Such shape of data can be further analyzed by differential geometry apparatuses, including curvatures \cite{nguyen2019dg},   Hodge decomposition \cite{zhao2020rham} and evolutionary de Rham-Hodge theory \cite{chen2021evolutionary}.

\section{Methods and Algorithms}
Let $\mathcal{Z}  := \{z_{m}^i\}_{m=1, i=1}^{M, I}$ with 
 $M$ and $I$ being the number of input data entries (i.e., samples) and the number of features for each data entry, respectively. 
Our goal is to find an $N$-dimensional representation of the original data, denoted as 
 $\mathcal{X}  := \{x_{m}^i\}_{m=1, i=1}^{M, N}$,  
such that $1 \le N << I$, by using a data-domain two-step  clustering-projection strategy. 
%

\subsection{Feature clustering }
Let $\mathcal{Z}  = \{\mathbf{z}^1, ..., \mathbf{z}^i, ..., \mathbf{z}^I\}$ be the set of  data, where  $\mathbf{z}^i \in \mathbb{R}^M$ represents the $i$th feature vector for the data. 
The objective is to partition the feature vector into $N$ parts, where $1\le N <<I$ is a preselected  reduced feature dimension. To this end, we   find an optimal disjoint partition of the data $\displaystyle \mathcal{Z}  := \uplus_{n=1}^N \mathcal{Z}^{n}$, for a given $N$, where $\mathcal{Z}^{n}$ is the $n$th partition (cluster) of the features.

To seek the optimal partition, we first analyze the correlation among feature vectors $  \mathbf{z}^i $. A variety of correlation measures can be used for this purpose. We discuss two standard approaches. 

\paragraph{Covariance  distance}

First, we consider an $I\times I$ normalized covariance matrix with component 
\begin{align}
    {\rho}(\mathbf{z}^i,\mathbf{z}^j)=
	\frac{\text{Cov}(\mathbf{z}^i, \mathbf{z}^j)}{\sigma(\mathbf{z}^i)\sigma(\mathbf{z}^j)}, \quad 1 \le i,j \le I,
\end{align}
where $\text{Cov}(\mathbf{z}^i, \mathbf{z}^j)$ is the covariance of $\mathbf{z}^i$ and $\mathbf{z}^j$, and $\sigma(\mathbf{z}^i)$ and $\sigma(\mathbf{z}^i)$ are the variances of $\mathbf{z}^i$ and $\mathbf{z}^j$, respectively. 

We set  negative correlations to zero and subtract from 1 to obtain a {\it covariance distance} between feature vectors
\begin{align}
	 \|\mathbf{z}^i-\mathbf{z}^j\|_{\rm dCov}= \begin{cases}
			1 - \rho(\mathbf{z}^i, \mathbf{z}^j), & \rho(\mathbf{z}^i, \mathbf{z}^j) > 0 \\
			0, & \text{otherwise.}
	\end{cases}
\end{align}
Note that  covariance distances have the range of $0 <   \|\mathbf{z}^i-\mathbf{z}^j\|_{\rm dCov} < 1$, for all pairs of vectors, $\mathbf{z}^i$ and $\mathbf{z}^j$. Highly correlated feature vectors will have their covariance distances close to 0, while the uncorrelated feature vectors will have their covariance distances close to 1.

\paragraph{Correlation distance}

 Alternatively, one may also consider the correlation distance defined via the distance correlation \cite{szekely2007measuring}. 
First, one computes a distance matrix for each vector $\mathbf{z}^i, i=1,2,...,I$
\begin{equation}
	a^i_{jk} = \|z^i_m- z^i_k\|, m,k=1,2,...,M, 
\end{equation}
where $\|\cdot\|$ denotes the Euclidean norm. 
Define  doubly centered distance for vector $\mathbf{z}^i$,
\begin{equation}
	A^i_{jk} :=a_{jk}-\bar{a}_{j\cdot} -\bar{a}_{ \cdot k}+ \bar{a}_{\cdot\cdot},
\end{equation}
where  $ \bar{a}_{j\cdot}$ is the $j$th row mean, $\bar{a}_{ \cdot k}$ is the $k$th column mean,  and $\bar{a}_{\cdot\cdot}$ is the grand mean of the distance matrix for  vector $\mathbf{z}^i$.

For a pair of vectors $(\mathbf{z}^i,\mathbf{z}^j)$, the squared   distance covariance is given by 
\begin{equation}
	{\rm dCov}^2(\mathbf{z}^i,\mathbf{z}^j):=\frac{1}{M^2}\sum_{j}\sum_k A^i_{jk} A^j_{jk}.  
\end{equation}
The distance correlation between vectors $(\mathbf{z}^i,\mathbf{z}^j)$ is given by 
 \begin{equation}
	{\rm dCor}(\mathbf{z}^i,\mathbf{z}^j):=\frac{ {\rm dCov}^2(\mathbf{z}^i,\mathbf{z}^j)}{{\rm dCov}(\mathbf{z}^i, \mathbf{z}^i){\rm dCov}(\mathbf{z}^j, \mathbf{z}^j)}.    
\end{equation}
We define a {\it correlation distance} between vectors  $\mathbf{z}^i$ and $\mathbf{z}^j$ as 
 \begin{equation}
	\| \mathbf{z}^i-\mathbf{z}^j\|_{\rm dCor}=1- {\rm dCor}(\mathbf{z}^i,\mathbf{z}^j).  
\end{equation}
The correlation distance has values in  range $0 \leq \| \mathbf{z}^i-\mathbf{z}^j\|_{\rm dCor}\leq 1$.   It gives   $ \| \mathbf{z}^i-\mathbf{z}^j\|_{\rm dCor}=1$ if $\mathbf{z}^i$ and $\mathbf{z}^j$ are uncorrelated or independent. When $\mathbf{z}^i$ and $\mathbf{z}^j$ are linearly depends on each other, one has $ \| \mathbf{z}^i-\mathbf{z}^j\|_{\rm dCor}=0$. 

\paragraph{Correlated clustering} 

Feature partition can be achieved with a variety of clustering methods. Here, as an example, we utilize a modified $K$-medoids method to perform the partition in a minimization process. Certainly, other $K$-means type of algorithms, including BFR algorithm, centroidal Voronoi tessellation, $k$ $q$-flats, $K$-means++, etc., can be utilized for our feature partition as well.

For a pre-selected $N$, we begin by randomly selecting $N$ medoids $\{\mathbf{m}^n   \}_{n=1}^N$ and assign each vector to its nearest medoid, which gives rise to the initial partition $\{  \mathcal{Z}^n \}_{n=1}^N$. Second, we denote the closest vector to the center of the $n$th partition $ \mathcal{Z}^n$ as the new medoid $\{\mathbf{m}^n\in \mathcal{Z}^n \}_{n=1}^N$. We reassign each vector into its nearest medoid, resulting in a new partition $\{\mathcal{Z}^n \}_{n=1}^N$ to minimize the loss function or the accumulated distance. The process is repeated until $\{\mathcal{Z}^n\}_{n=1}^N$ is optimized with respect to a specific distance definition, 
\begin{align}\label{argmin}
\displaystyle		{\rm arg~min}_{\{\mathcal{Z}^1,...,  \mathcal{Z}^n, ...,\mathcal{Z}^N \}} \sum_{n=1}^N \sum_{\mathbf{z}^i \in \mathcal{Z}^n} \|\mathbf{z}^i-\mathbf{m}^n\|,
\end{align}
 where $\|\cdot\|$ is either the covariance distance or the correlation distance. In comparison, covariance distance is easy to compute, while correlation distance can deal with complex nonlinear high-level correlations among feature vectors and samples. Note that many other metrics can be used too. For a given $N$, the minimization partitions similar feature vectors into $N$ clusters, which provides the foundation for further projections.

Our next goal is to project the original $I$-dimensional dataset $\mathcal{Z}$ into an $N$-dimensional representation $\mathcal{X}$ according to the partition result.

\subsection{Feature projection} 

\paragraph{Flexibility Rigidity Index (FRI)}

In this section, we review  Flexibility Rigidity Index (FRI)   \cite{xia2013multiscale}. 

Let $\{\mathbf{z}_1, ...,\mathbf{z}_m, ..., \mathbf{z}_M\}$ be the input dataset, where $\mathbf{z}_m \in \mathbb{R}^I$. Denote $\|\mathbf{z}_i - \mathbf{z}_j\|$ some   metric between $i$th and $j$th data entries, and the correlations between data entries are computed as 
\begin{align}
	C_{ij} = \Phi(\|\mathbf{z}_i - \mathbf{z}_j\|; \tau, \eta, \kappa), \quad 1 \le i,j \le M
\end{align}
where $\Phi$ is the correlation kernel, and $\tau,  \eta,  \kappa > 0$ are the parameters for the kernel. Commonly used metrics include  the Euclidean distance,  the Manhattan distance, the Wasserstein distance, etc. 

The correlation kernel is a real-valued smooth monotonically decreasing function, satisfying the two properties
\begin{align*}
	& \Phi(\|\mathbf{z}_i - \mathbf{z}_i\|; \tau,  \eta, \kappa) = 1 , \quad \text{as $|\mathbf{z}_i-\mathbf{z}_j| \to 0$} \\
	& \Phi(\|\mathbf{z}_i - \mathbf{z}_j\|; \tau,  \eta,  \kappa) = 0, \quad \text{as $|\mathbf{z}_i-\mathbf{z}_j| \to \infty$}
\end{align*}
A popular choice for such functions is a radial basis function. For example, one may use the generalized exponential function
\begin{align}
	\Phi(\|\mathbf{z}_i - \mathbf{z}_j\|; \tau,  \eta, \kappa) = \begin{cases} e^{-(\|\mathbf{z}_i - \mathbf{z}_j\|/ \tau \eta)^\kappa}, & \|\mathbf{z}_j - \mathbf{z}_j\| < r_c \\
		0, & \text{otherwise}
	\end{cases}
\end{align}
or the generalized Lorentz kernel
\begin{align}
	\Phi(\|\mathbf{z}_i - \mathbf{z}_j\|; \tau,  \eta, \kappa) = \begin{cases}\frac{1}{1 + \left(\|\mathbf{z}_i - \mathbf{z}_j\|/ \tau\eta\right)^\kappa}& \|\mathbf{z}_j - \mathbf{z}_j\| < r_c, \\
		0 & \text{otherwise}
	\end{cases}
\end{align}
 where $\kappa$ is the power, $\tau$ is the multiscale parameter, $\eta$ is the scale to be computed from the given data, and $r_c$ is the cutoff distance, which is useful in a certain data structure to reduce the computational complexity \cite{opron2014fast}.  In the context of t-SNE, $\eta$ would be the perplexity, and in UMAP, $\eta$ would define the geodesic of the Riemannian metric.

 We can construct a correlation matrix $C = \{C_{ij}\}$, which reveals the topological connectivity between samples \cite{xia2013multiscale}. We can also view such correlation map as a weighted graph \cite{wang2020persistent,nguyen2019agl}, where $r_c$ is the cutoff function. In order to understand the connectivity, we choose $\eta$ to be the average minimum distance  between the data entries
\begin{align*}
\displaystyle	\eta = \frac{\sum_{m=1}^M \min_{\mathbf{z}_j}\|\mathbf{z}_m - \mathbf{z}_j\|}{M}.
\end{align*}

Using the correlation function, we can define the  rigidity of $\mathbf{x}_i$ as 
\begin{align*}
	\mu_i = \sum_{m=1}^M \omega_{im}\Phi(\|\mathbf{z}_i - \mathbf{z}_m\|; \tau,  \eta, \kappa),
\end{align*}
where $\omega_{im}$ are the weights. Here, we set $\omega_{im} = 1$ for all $i$ and $m$. From the graph perspective, one can also view $\mu_i$ as the degree matrix of node $\mathbf{x}_i$.

\paragraph{Correlated  projection}

In this subsection, we employ FRI for the correlative dimensional reduction of input dataset $\{\mathbf{z}_1, ..., \mathbf{z}_m, ... ,\mathbf{z}_M\}$, where
 $\mathbf{z}_m \in \mathbb{R}^I$, 
 leading to a    low-dimensional representation $\{\mathbf{x}_1, ..., \mathbf{x}_m, ... ,\mathbf{x}_M\}$, with 
 $\mathbf{x}_m \in \mathbb{R}^N$. The FRI reduction captures the intrinsic correlation among samples. 

Recall that  $\{\mathcal{Z}^n\}_{n=1}^{N}$ are optimal partition of feature vectors from the $K$-medoids or another clustering method. Let $\mathcal{S} = \{1, ..., I\}$ be the whole set of indices of the feature vectors, and $S = \uplus_{n=1}^N \mathcal{S}^n$, where $\mathcal{S}^n = \{i | \mathbf{z}^i \in \mathcal{Z}^n\}$.    
 We can  define $\mathbf{z}_m^{\mathcal{S}^n}$ as $m$th input data with the $n$th collection of feature indices $\mathcal{S}^n$, i.e., $\mathbf{z}_m^{\mathcal{S}^n}:=\{z_m^i |i\in \mathcal{S}^n \}$.

We can now define the $n$th correlation matrix $\{C^{n}_{ij}\}_{i,j=1,...,M}$ associated with subset $\mathcal{S}^n $ of features
\begin{align}\label{weights}
	C^{n}_{ij} = \Phi^{n}(\|\mathbf{z}_i^{\mathcal{S}^n} - \mathbf{z}_j^{\mathcal{S}^n}\|; \tau,  \eta,  \kappa), \quad 1 \le i,j \le M, \quad 1 \le n \le N,
\end{align}
where $\Phi^{n}$ is the radial basis kernel for the $k$th grouping. For example, one may choose 
\begin{align}
	\Phi^{n}(\|\mathbf{z}^{S_n}_i - \mathbf{z}^{S_n}_j\|; \tau, \eta^n,  \kappa) &= \begin{cases} e^{-(\|\mathbf{z}^{\mathcal{S}^n}_i - \mathbf{z}^{\mathcal{S}^n}_j\|/ \tau \eta^n)^\kappa}, & \|\mathbf{z}^{\mathcal{S}^n}_i - \mathbf{z}^{\mathcal{S}^n}_j\| < r^n_c \\
		0, & \text{otherwise},  
	\end{cases}~~  {\rm or} \\
	\Phi^{n}(\|\mathbf{z}^{\mathcal{S}^n}_i - \mathbf{z}^{\mathcal{S}^n}_j\|; \tau, \eta^n,\kappa) &= \begin{cases}\frac{1}{1 + \left(\|\mathbf{z}^{\mathcal{S}^n}_i - \mathbf{z}^{\mathcal{S}^n}_j\|/ \tau\eta^n\right)^\kappa}, & \|\mathbf{z}^{\mathcal{S}^n}_i - \mathbf{z}^{\mathcal{S}^n}_j\| < r^n_c \\
		0, & \text{otherwise},
	\end{cases}
	\label{eq: fri}
\end{align}
where truncation distance ($r^n_c$) can be set to 2 or 3-standard deviations, and $\eta^n$ is set to 
\begin{align*}
\displaystyle	\eta^n = \frac{\sum_{m=1}^M \min_{\mathbf{z}_j^{\mathcal{S}^n}}\|\mathbf{z}_m^{\mathcal{S}^n} - \mathbf{z}_j^{\mathcal{S}^n}\|}{M}.
\end{align*}
Then, we can project the data to an $N$-dimensional space representation by taking the   rigidity function defined by correlation kernels, 
\begin{align}\label{xcomponent}
	x_i^n = \sum_{m=1}^M \omega_{im}^{n}\Phi^{n}(\|\mathbf{z}_i^{\mathcal{S}^n} - \mathbf{z}_m^{\mathcal{S}^n}\|; \tau,  \eta^n, \kappa), ~ n=1,2,...,N; i=1,2,...,M
\end{align}
where $\omega_{im}^{n}$ are the weights associated with $\Phi^{n}$ for the $n$th cluster and can be set to 1. Moreover, the $m$th data  in the reduced $n$-dimension representation is a vector  $\mathbf{x}_m = (x_m^1, ..., x_m^n, ..., x_m^N)^T$. 

We also propose to improve the computational efficiency in Eq. (\ref{xcomponent}) by avoiding the global summation. This can be easily done as follows. First, we construct an $M\times M$ global distance matrix of the samples to obtain the nearest neighbors of each sample. Then, we use the cell lists algorithm with the cutoff value to replace the global summations   Eq. (\ref{xcomponent}) by considering only a few nearest neighbors \cite{opron2014fast}. This approach significantly reduces the memory requirement.  
Since the projections of $x_i^n$ for different $i$ and $n$ are independent of each other, massively parallel computations can be used for large datasets.  
\subsection{Visualization and analysis }

\paragraph{The shape of  data}
 Continuous FRI was defined to offer the shape of $M$ data entries in $\mathbb{R}^3$   \cite{xia2013multiscale}. A similar idea was used to define interactive differentiable Riemannian manifolds \cite{nguyen2019dg}. Here, we extend these ideas to construct Grassmann manifolds Gr($N-1, I$). 

Let $\mathcal{X}= \{\mathbf{x}_1, ..., \mathbf{x}_m, ..., \mathbf{x}_M\}$ be a finite set of $M$ data entries. Denote $\mathbf{x}_m \in \mathbb{R}^N$ be the feature vector for the $m$th sample, and $|\mathbf{x} - \mathbf{x}_m|$ be the Euclidean distance between a point $ \mathbf{x} \in \mathbb{R}^N$ to the $j$th sample. Let $\displaystyle \eta = \frac{1}{M}\sum_{m=1}^M \min_{\mathbf{x}_j} \|\mathbf{x}_m - \mathbf{x}_j\|$ be the average minimum pairwise distance of the  input data. 
Then, the unnormalized rigidity density at point $\mathbf{x} \in \mathbb{R}^N $ is given by  
\begin{align}\label{density}
	\mu(\mathbf{x}) = \sum_{m=1}^M \omega_m \Phi(\|\mathbf{x} - \mathbf{x}_m\|; \tau, \eta,\kappa),
\end{align}
where $\omega_m = 1$, and $\tau $ and $ \kappa$ are the hyperparameters of the correlation kernel $\Phi$. Notice that we can choose an isosurface $	\mu(\mathbf{x}) = c \mu_{\max}$, which defines an $(N-1)-$ dimensional Riemannian manifold by the collection of   points 
\begin{align}
\{\mathbf{x}| \mathbf{x} \in \mathbb{R}^N, 	\mu(\mathbf{x}) = c \mu_{\max}\}, 
\end{align}
 where $c \in (0,1)$ and $ \displaystyle \mu_{\max} = \max_{\mathbf{x}}  \mu(\mathbf{x}) $. 
 The shape of data can be directly visualized for $2\leq N\leq 3$ as shown in Ref. \cite{nguyen2019dg}. 

One can restrict $\mathbf{x}_m$ to a given subset in Eq. (\ref{density}) to compare the shape of data in different classes when the class labels are known.  

For further analysis,  one can obtain $(N-1)$ independent curvatures via fundamental forms \cite{nguyen2019dg}. Additionally, Hodge decomposition can be applied to analyze topological connectivity (i.e., Betti numbers associated with the harmonic spectra) and non-harmonic spectra of the Hodge Laplacians of the data \cite{zhao2020rham}. For evolving manifolds, the evolutionary de Rham-Hodge theory can be used to analyze the geometry and topology of data \cite{chen2021evolutionary}.

\paragraph{Residue-Similarity (R-S) scores and indexes}

Traditionally, the visualization of data is enabled from reducing the data into 2 or 3 feature components. However, such aggressive reduction leads to poor representations for data with high intrinsic dimensions, despite of nice visualization effects. For classification problems with 2 classes, receiving operation characteristic (ROC) curve and Area Under the ROC Curve (AUC) curve can be used to show the performance. However, not all classification is binary. In this section, we introduce a new visualization tool called the Residue-Similarity (R-S) scores or R-S plots, which can be applied to an arbitrary number of classes.

An R-S plot consists of two components, residue and similarity scores. Assume that the data is $\{(\mathbf{x}_m, y_m)| \mathbf{x}_m \in \mathbb{R}^N, y_m \in \mathbb{Z}_L\}^M_{m=1}$, where $\mathbf{x}_m$ is the $m$th data. 
For classification problems, $y_m$ is the ground truth, and for clustering problems, $y_m$ is the cluster label.
Here, $N$ is the number of features and $M$ is the number of samples. $L$ is the number of classes, that is $y_m\in [0,1,...,L-1]$. 
We can partition $\mathcal{X}= \{\mathbf{x}_m\}_{m=1}^M$ into $L$ classes by taking $\mathcal{C}_l = \{\mathbf{x}_m \in \mathcal{X}| y_m = l\}$. Note that $\uplus_{l=0}^{L-1}\mathcal{C}_l = \mathcal{X}$.

The residue score is defined as the inter-class sum of distances. Suppose $y_m = l$. Then, the residue score for $\mathbf{x}_m$ is given by
\begin{align}
	R_m:= R(\mathbf{x}_m) = \frac{1}{  R_{\max}}\sum_{\mathbf{x}_j \not\in \mathcal{C}_l} \|\mathbf{x}_m - \mathbf{x}_j\|,
\end{align}
where $\|\cdot\|$ is the  distance between a pair of vectors and $\displaystyle R_{\max} = \max_{\mathbf{x}_m\in \mathcal{X}} R(\mathbf{x}_m)$ is the maximal residue score. The similarity score is given by taking the average intra-class score. That is, for $y_m = l$, 
\begin{align}
	S_m:=S(\mathbf{x}_m) = \frac{1}{|\mathcal{C}_l|} \sum_{\mathbf{x}_j \in \mathcal{C}_l} \left( 1-\frac{\|\mathbf{x}_m - \mathbf{x}_j\|}{d_{\max}} \right),
\end{align}
where $ \displaystyle d_{\max} = \max_{\mathbf{x}_i,  \mathbf{x}_j \in \mathcal{X}} \|\mathbf{x}_i - \mathbf{x}_j\|$ is the maximal pairwise distance of the dataset. Note that by scaling, $0 \le R(\mathbf{x}_m) \le 1$ and $0 \le S(\mathbf{x}_m) \le 1$ for all $\mathbf{x}_m$. In this work, we employ the Euclidean distance in our R-S scores. However, other distance metrics can be similarly used as well. 
In general, a large $R(\mathbf{x}_m)$ indicates that the data is far from other classes, and a large $S(\mathbf{x}_m)$ indicates that the data is well clustered. Since $ R_{\max}$ and $d_{\max}$ are for the whole dataset, residue and similarity scores in different classes can be compared. 

The residue score and similarity score can be used to visualize each class separately, where $R(\mathbf{\mathbf{x}})$ is the $x$-axis, and $S(\mathbf{x})$ is the $y$-axis. In the case of classification, define $\{(\mathbf{x}_m, y_m, \hat{y}_m) | x_m \in \mathbb{R}^N, y_m \in \mathbb{Z}_L, \hat{y}_m \in \mathbb{Z}_L\}_{m=1}^M$, where $\hat{y}_m$ is the predicted label for the $m$th sample. Then, we can repeat the above process by using the ground truth  and visualize each class separately. By coloring the data point with the predicted label $\hat{y}_m$, we get the R-S score visualization of the classification.

Class  residue  index (CRI) and class  similarity  index (CSI) can be easily defined for the $l$th class as ${\rm CRI}_l =\frac{1}{| \mathcal{C}_l|}\sum_m R_m  $ and  ${\rm CSI}_l =\frac{1}{| \mathcal{C}_l|}\sum_m S_m$, respectively. Such   indexes can be used to compare the distributions in different classes obtained by different methods. 

The above indices depend on clusters or classes. It is more useful to construct class-independent global indices. To this end, we first  define residue index (RI) and similarity index (SI)  as ${\rm RI} =\frac{1}{L}\sum_l {\rm CRI}_l $ and ${\rm SI} =\frac{1}{L}\sum_l {\rm CSI}_l $, respectively.    All of these indexes have the range of [0,1] and the larger the better for a given dataset. 
Additionally, we define R-S disparity (RSD) as ${\rm RSD}={\rm RI}-{\rm SI}$. RSD ranges [-1,1]. Finally, we define R-S index (RSI) as ${\rm RSI=1-|{\rm RI}-{\rm SI}|}$. R-S index has the range of  [0,1]. 

The Rand index is known to correlate with accuracy \cite{rand1971objective}.  
We speculate that the R-S disparity may correlate with the convergence of clustering and the R-S index may correlate with the accuracy of classification. 
R-S disparity and R-S index can be used to measure the performance of different methods.

\paragraph{Persistent Spectral Graph (PSG)}

 Further analysis of point cloud data or the points in the R-S plot can be carried out with Topology Data Analysis (TDA).  Persistent homology \cite{frosini1992measuring,edelsbrunner2000topological,zomorodian2005computing,carlsson2009topology, mischaikow2013morse,KLXia:2014c,townsend2020representation} is an algebraic topology technique and the main workhorse of TDA. It introduces a filtration process to generate a family of topological spaces so that the original data can be analyzed in multiscales. However, it cannot detect the homotopic shape evolution of data during filtration. Topological Laplacians, such as persistent spectral graph (aka persistent Laplacian) \cite{wang2021hermes,memoli2020persistent} and evolutionary  Hodge Laplacian \cite{chen2021evolutionary} are designed to preserve full topological persistence and capture homotopic shape evolution of data during a filtration.  The persistent spectral graph returns the same multiscale topological invariants in its kernels of various dimensions and scales but offers additional homotopic shape information in its non-harmonic spectra.

Considering two boundary operators  $\partial_q^t: C_q(K_t) \mapsto C_{q-1}(K_{t})$ and $\partial_{q+1}^{t+p}: C_{q+1}(K_{t+p}) \mapsto C_{q}(K_{t+p})$, where $ C_{q}(K_{t+p})$ is a chain group and $K_{t}\subset K_{t+p}$ are simplicial complexes generated by a filtration. 
Denote $\partial^{t+p}_{q+1}\rvert_{\cplx^{t,p}_{q+1}}$ as $\text{\dh}_{q+1}^{t,p}$ such as 
\begin{align}
    \cplx^{t,p}_{q+1} = \{\alpha \in C_{q+1}^{t+p} \mid \partial_{q+1}^{t+p}(\alpha) \in C_{q}^t \}.
\end{align}
 Namely, $\cplx^{t,p}_{q+1}$ consists of elements whose images under $\partial_{q+1}^{t+p}$ are in $C_{q}^t$.
The $p$-persistent $q$-combinatorial Laplacian operator \cite{wang2020persistent} is given by 
\begin{equation}\label{PLaplacian}
    \Delta_q^{t,p} = \text{\dh}_{q+1}^{t,p} (\text{\dh}_{q+1}^{t,p})^{\ast} + (\partial_q^t)^*\partial_q^t.
\end{equation}
 
The topological invariants of the corresponding persistent homology defined by the same filtration are recovered from the kernel of the persistent Laplacian Eq. (\ref{PLaplacian}) \cite{wang2020persistent},
 \begin{equation}
     \beta_q^{t,p} = \dim \ker \partial_q^t - \dim \im \text{\dh}_{q+1}^{t,p} = \dim \ker \Delta_q^{t,p}.
 \end{equation}
State differently, the zero eigenvalues  of the persistent Laplacian operator Eq. (\ref{PLaplacian}) give rise to the entire topological variants of the persistent homology. Then,  the non-harmonic part of the spectra (i.e., the non-zero eigenvalues  of the persistent Laplacian) and associated eigenvectors  offer additional shape information of the underlying data. 

Note that for small-sized high-dimensional datasets, PSG can be directly employed to reduce the dimensionality in terms of the statistical quantities of the data spectra. The resulting spectra or their statistics can be directly used to represent the original datasets.

\section{Results}
 
In this section, we numerically explore CCP's performance on a variety of high dimensional benchmark test datasets. For each dataset, we use  10 random seeds for 5-fold or 10-fold cross-validations, depending on the number of samples of the data.

In order to validate the effectiveness of CCP, we compare it with Uniform Manifold Approximation and Projection (UMAP), Principle Components Analysis (PCA), Locally Linear Embedding (LLE), and Isomap. 

For metric-based embedding, the Euclidean distance was used. All parameters were set to default, according to the packages outlined in \autoref{tab: version}. In order to test the effectiveness of the dimensionality reduction, a $k$-nearest neighbor classifer ($k$NN) was used. \autoref{tab: version} shows the versions of the packages used in our comparison. 

In order to visualize the accuracy, R-S scores were utilized. In R-S plots, the residue and similarity scores were represented as the $x$ and $y$ axes, respectively, and the data points were colored according to the predicted labels from classification results.

\begin{table}[H]
	\centering 
	\begin{tabular}{l } \hline
		Package  \\ \hline
		Python v3.8.5 \\
		Numpy v1.19.2 \\
		Scikit-learn v0.23.2 \\
		Scikit-learn-exta v0.2.0 \\
		Sklearn v0.0 \\
		umap-learn v0.5.1 \\\hline
	\end{tabular}
	\caption{Python packages used for the dimensionality reduction and benchmark tests.}
	\label{tab: version}
\end{table}

\subsection{Datasets}
We test CCP and several other commonly used algorithms on benchmark datasets, including 	Leukemia ALL-AML, Carcinoma, Arcene, TCGA-PANCAN, 
Coil-20 and Coil-100, and 	Smallnorb. \autoref{tab: dataset} summarizes the datasets used in the present work.

\begin{table}[H]
	\caption{datasets used in the benchmark tests}
	\label{tab: dataset}
	\begin{tabular}{l | l  p{10cm}} \hline
		dataset [ref] & ($M$, $I$, $L$) & Description \\ \hline
	
				Leukemia \cite{ding2005minimum}  &(72, 7070, 2) & Microarray dataset of Leukemia. The data contains 72 samples, each with 7070 gene expressions.\\
		
		Carcinoma \cite{su2001molecular, yang2006stable} & (174, 9182, 11) & Microarray dataset of human carcionmas. Original data \cite{su2001molecular} contains 12533 genes, which were processed to 9182 dimensions in Ref.  \cite{yang2006stable}. \\

		ALL-AML \cite{golub1999molecular} & (72, 7129, 2) & Cancer classification dataset based on gene expressions by DNA microarrays of acute myeloid Leukemia (AML) and acute lymphoblastic Leukemia (ALL). \\

		TCGA-PANCAN \cite{chang2013cancer} & (801, 20531, 5) & Gene expression dataset. Part of the RNA-seq (HiSeq) PANCAN data, where expressions of 5 different types of tumors were extracted. \\
		
		Coil-20 \cite{nene1996columbia20} & (1440, 16384, 20) & Image classification dataset with 1440 images. Each image has size $128 \times 128=16384$, where 20 objects are captured at 72 angles. Each image was treated as a vector of length 16384. \\
		
		Coil-100 \cite{nene1996columbia100} & (7200, 49152, 100) & Image classification dataset of 7200 images. Each image has size $128 \times 128=16384$ with 3 channels, where 100 objects are captured at 72 angles. Each image was treated as a vector of length 49152. \\ 
		
Smallnorb \cite{LeCun2004LearningMF} & (24300,18432,5) & Image classification dataset with 5 generic categories: four-legged animals, human figures, airplanes, trucks, and cars. Each object was taken from a variety of radial and azimuthal angles. Each sample consists of 2 images, the left and the right views, both of size 96$\times$96. Both images for flattened to create a vector of length 2$\times$96$\times$96.\\ \hline
	\end{tabular}
\end{table}

\subsection{Validation}
\subsubsection{Clustering analysis}
Since CCP uses clustering to partition features based on correlations, it is vital to analyze clustering effectiveness.

One of the best methods to evaluate the effectiveness of the clustering from $k$-medoids and/or $k$-means is to compute the loss function with respect to the varying number $k$ of clusters. Then, using the elbow method, one finds the inflection point of the loss function and sets the corresponding $k$ to be the optimal number of clusters. 

\begin{figure}[H]
	\centering
	\includegraphics[width=1.0\textwidth]{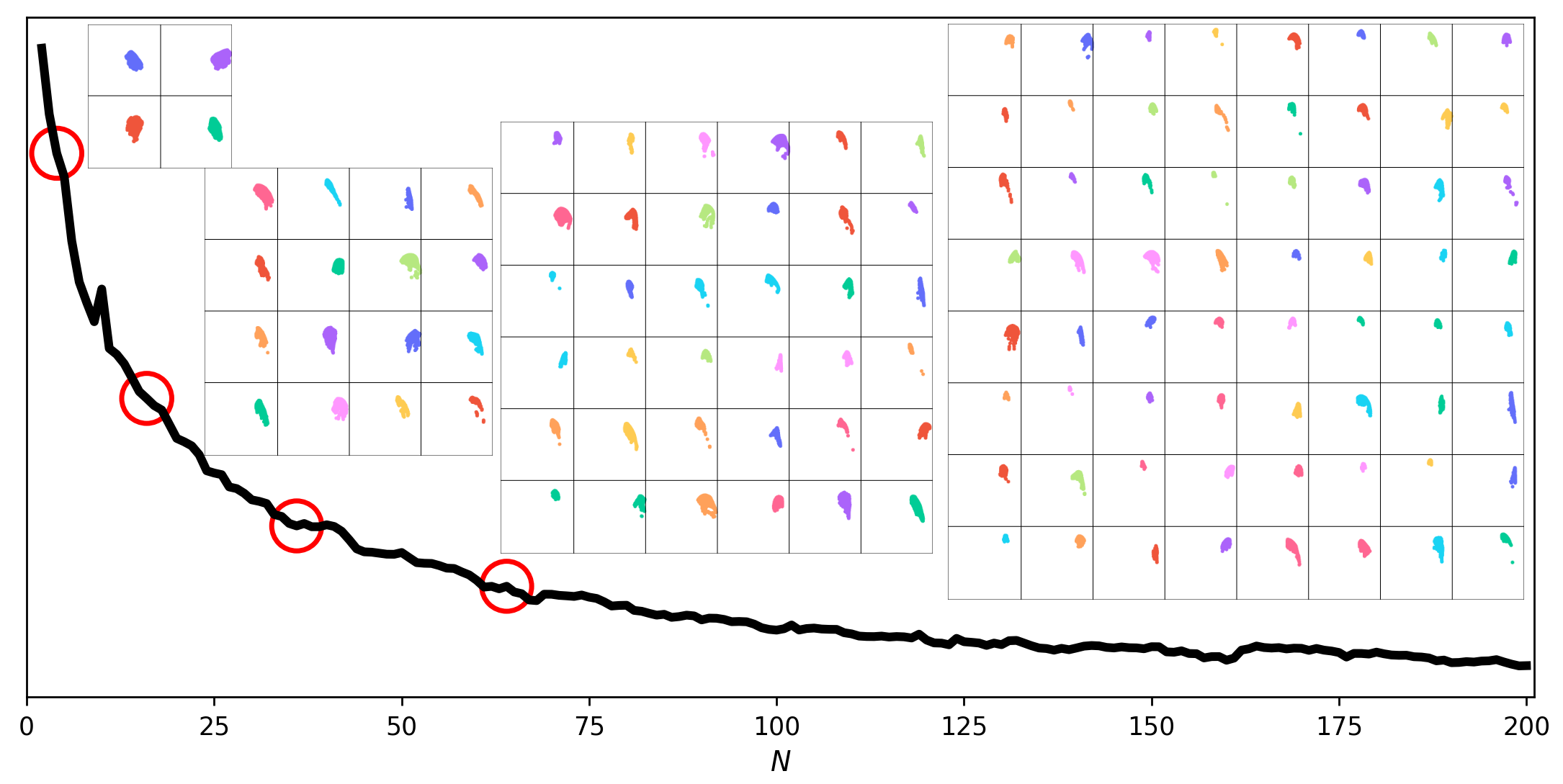}
	\caption{Illustration of the partition and  R-S visualization of 16384 features of the Coil-20 dataset into different numbers ($N$) of clusters. The line is the loss function defined in Eq. (\ref{argmin}) with respect to different $N$ values ranging from  2 to 200. Four red circles indicate $N=4,16,36,$ and $64$, for which the corresponding   R-S charts are given in charts from the left to the right. Each section of the charts represents an individual feature cluster with the $x$ and $y$ axes are the residue and similarity scores, respectively.  The R-S indices for $N = 4, 16,36$ and $64$ are $0.842, 0.887, 0.952$ and $0.992$, respectively.}
	\label{fig: kmedoids visual}
\end{figure}

In this work, we present another method for visualizing the feature clusters using our R-S scores. \autoref{fig: kmedoids visual} shows the loss function of the $k$-medoids feature clustering given in Eq. (\ref{argmin}) for Coil-20 dataset for $N=2$ to $200$. In addition, for $N=4,16,36$, and $64$, the clustering results were visualized using R-S  scores. The red circles on the loss function correspond to $N=4,16,36,$ and $64$, plotted in four charts.  Each section in a given chart represents one cluster, and the $x$ and the $y$-axes represent the residue and similarity scores, respectively for the cluster. For $N=4$, the cluster colored in blue has a low similarity score, indicating that the data is spread out within the cluster. It indicates that a larger $N$ is needed. From $N=36$ to $N = 64$, it is seen that the loss function has an elbow, indicating that these numbers of clusters are appropriate. The R-S plots in these two charts are more compact, indicating that the clustering performs well.

As speculated earlier,  the R-S disparity may correlate with the convergence of clustering. To verify this speculation, we have computed R-S disparity for the feature clustering results at $N = 4, 16, 36$, and $64$.  We found that ${\rm RSD}_{N=4}=0.158$,  ${\rm RSD}_{N=16}=0.113$, ${\rm RSD}_{N=36}=0.048$, and ${\rm RSD}_{N=64}=-0.008$.  Therefore, R-S disparity correlates strongly with the loss function of the $k$-medoids clustering shown in  \autoref{fig: kmedoids visual}.

One of the advantages of the $N$-medoids method is that the cluster centers will always be one of the feature vectors. In addition, since each medoid is always one of feature vectors, the method is much more efficient when dealing with high dimensional data. Other clustering methods, such as $k$-means, spectral clustering, DBSCAN, and hierarchical clustering, can be utilized for the clustering as well.
 
\subsubsection{Partition scheme evaluation}
In order to explore the effectiveness of different partitions, we compare results obtained with three feature partitions: correlation partition, random equal partition, and equal variance partition.

 In the random equal partition, the features are randomly shuffled and split into $N$ equal-sized clusters (i.e., $N$ dimensions in the CCP). Therefore, each cluster has the same number of features, which will be projected into a one-dimensional representation in CCP. In equal variance partition, the features are normalized with respect to the largest one and ordered, and then,  are split into $N$ clusters such that all clusters have a similar amount of variance. In this partition, the first cluster contains the largest number of low-variance features, whereas the last cluster, cluster $N$, contains the least number of high-variance features. Notice that in the correlation partition, the numbers of features in clusters also vary and are determined by minimization according to Eq. (\ref{argmin}).  

The Leukemia and Carcinoma datasets were used to compare the 3 feature partition schemes. For both tests, 5-fold cross-validations with 10 random seeds were used for the dimensionality reduction, and $k$-NN was used to obtain classification accuracy. For each fold of partition, all results attained from 10 seeds were included to evaluate partition schemes.

\begin{figure}[H]
	\centering
	\begin{subfigure}{\textwidth}
	\centering
	\includegraphics[scale=0.4]{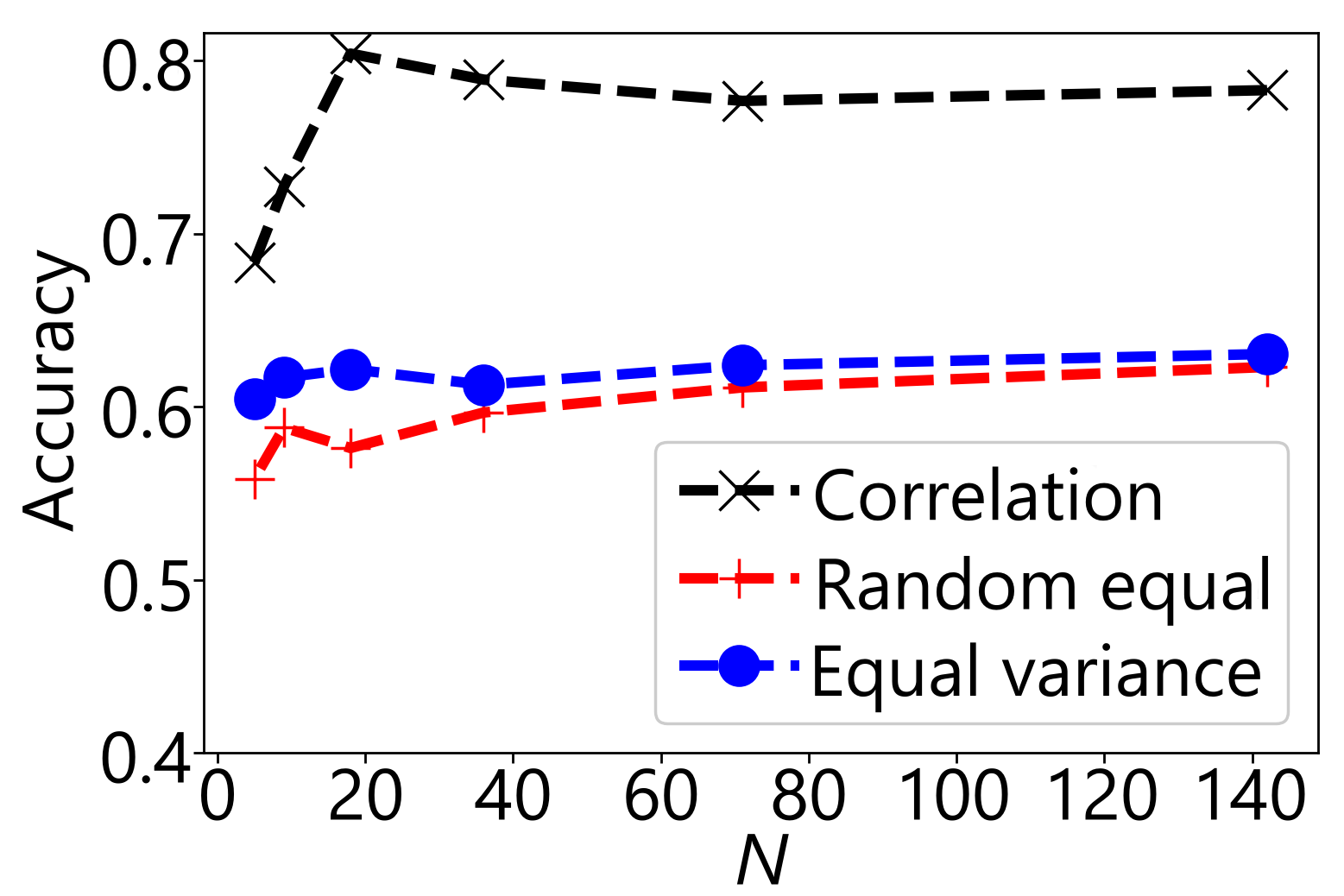}
	\caption{Comparison of the correlation partition, random equal partition, and equal variance partition-based dimensionality reduction and classification for the Leukemia dataset of 7070 features. The accuracies are computed from $k$NN classifications using the reduced $N$ features generated from CCP with three partitions. 
	}
	\end{subfigure}
	\begin{subfigure}{\textwidth}
		\centering
	\includegraphics[scale = 0.4]{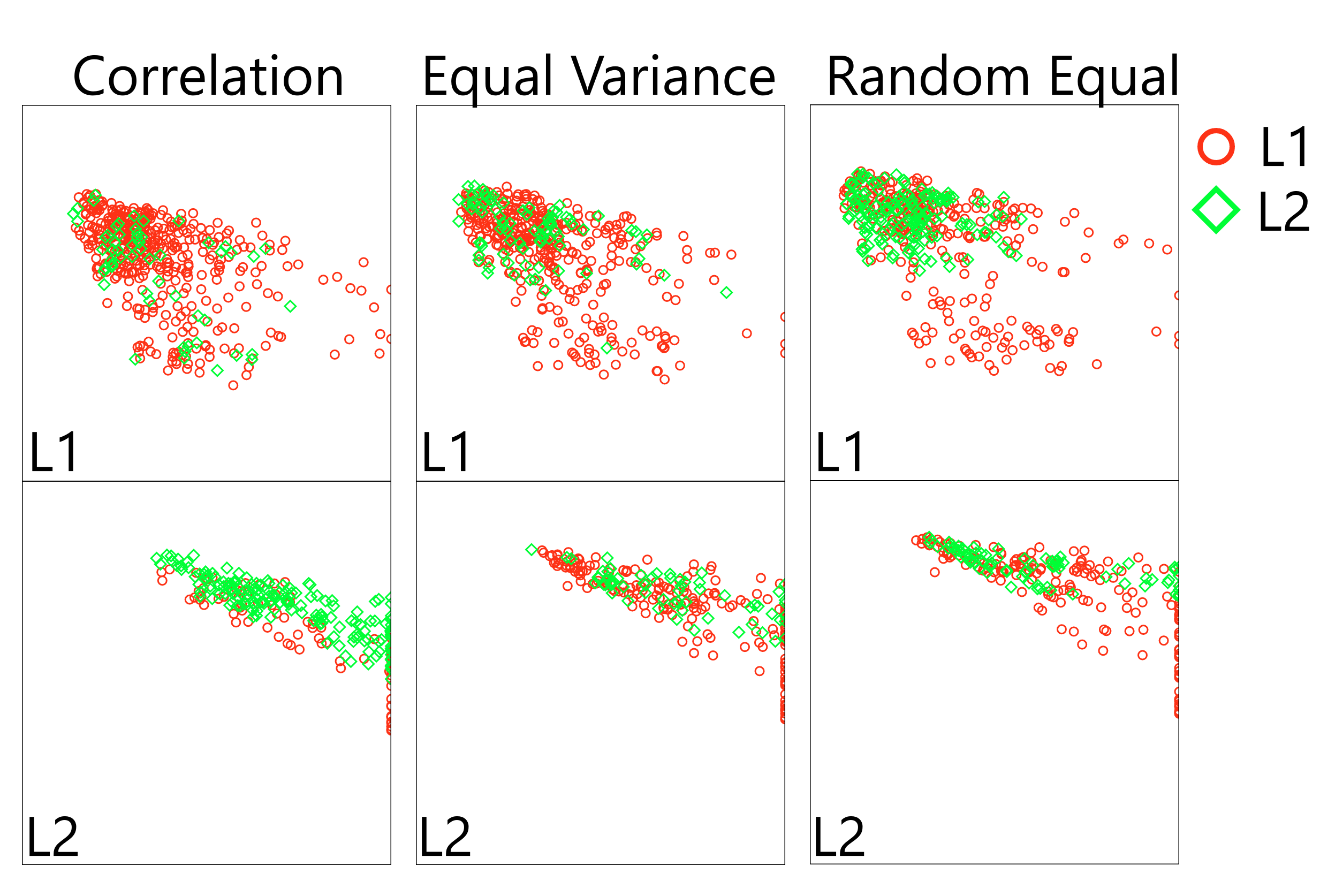}
	\caption{R-S plots of clusters generated from CCP-based classification using correlation partition, equal variance partition, and random equal partition of the Leukemia dataset at $N=18$.}
	\end{subfigure}
	\caption{Comparing the CCP-based classification effectiveness using correlation partition, equal random partition, and equal variance partition of the features of the leukemia dataset. For FRI, exponential kernel with $\kappa = 2$ and $\tau = 1.0$ was used. For each test, results from all 10 seeds were plotted. Form left to right: R-S plots of correlation partition, equal variance partition, and random equal partition. The $x$-axis is the residual score, and the $y$-axis is the similarity score. Each section corresponds to one cluster and the data was colored according to the predicted labels from the $k$-NN classification.}
	\label{fig:leukima}
\end{figure}

\autoref{fig:leukima}(a) shows the accuracy of CCP-based classification of  the Leukemia dataset under various CCP reduced dimensions $N$ equipped with 3 feature partition schemes. The correlation partition outperforms both equal random and variance partitions over all $N$ values. In addition, as shown in \autoref{fig:leukima}(b),  for   R-S plots,  correlation partition outperforms  other two partitions in each cluster at $N=18$. In particular, equal random partition and equal variance partition do not work well in classifying label 2.

\begin{figure}[H]
	\centering
	
	\begin{subfigure}{\textwidth}
		\centering
		\includegraphics[scale=0.4]{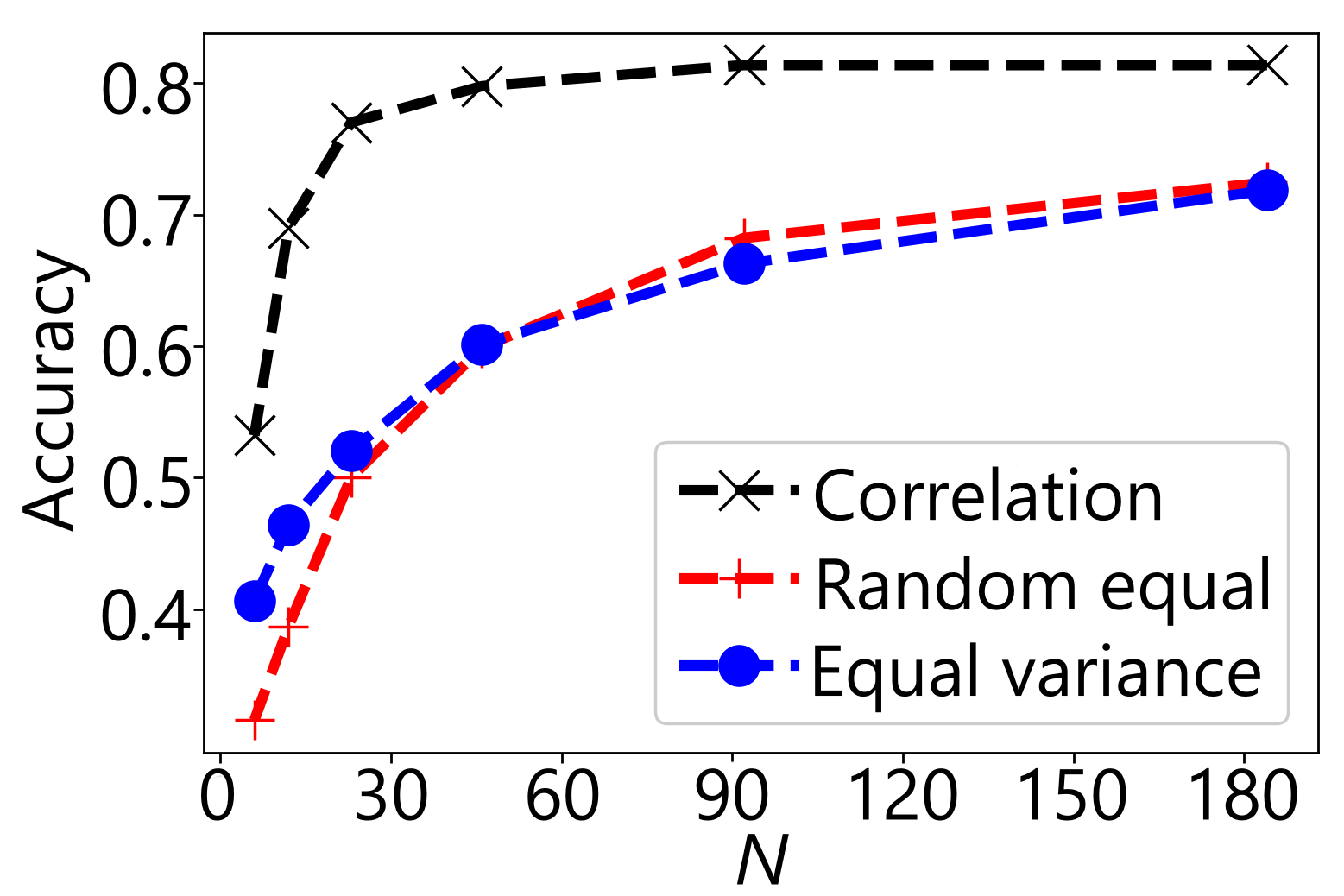}
		\caption{Comparison of the correlation partition, random equal partition, and equal variance partition-based dimensionality reduction and classification for the  Carcinoma dataset of 7070 features. The accuracies are computed from $k$NN classifications using the reduced $N$ features generated from CCP with three partitions.   }
	\end{subfigure}

	\begin{subfigure}{\textwidth}
		\centering
		\includegraphics[scale = 0.4]{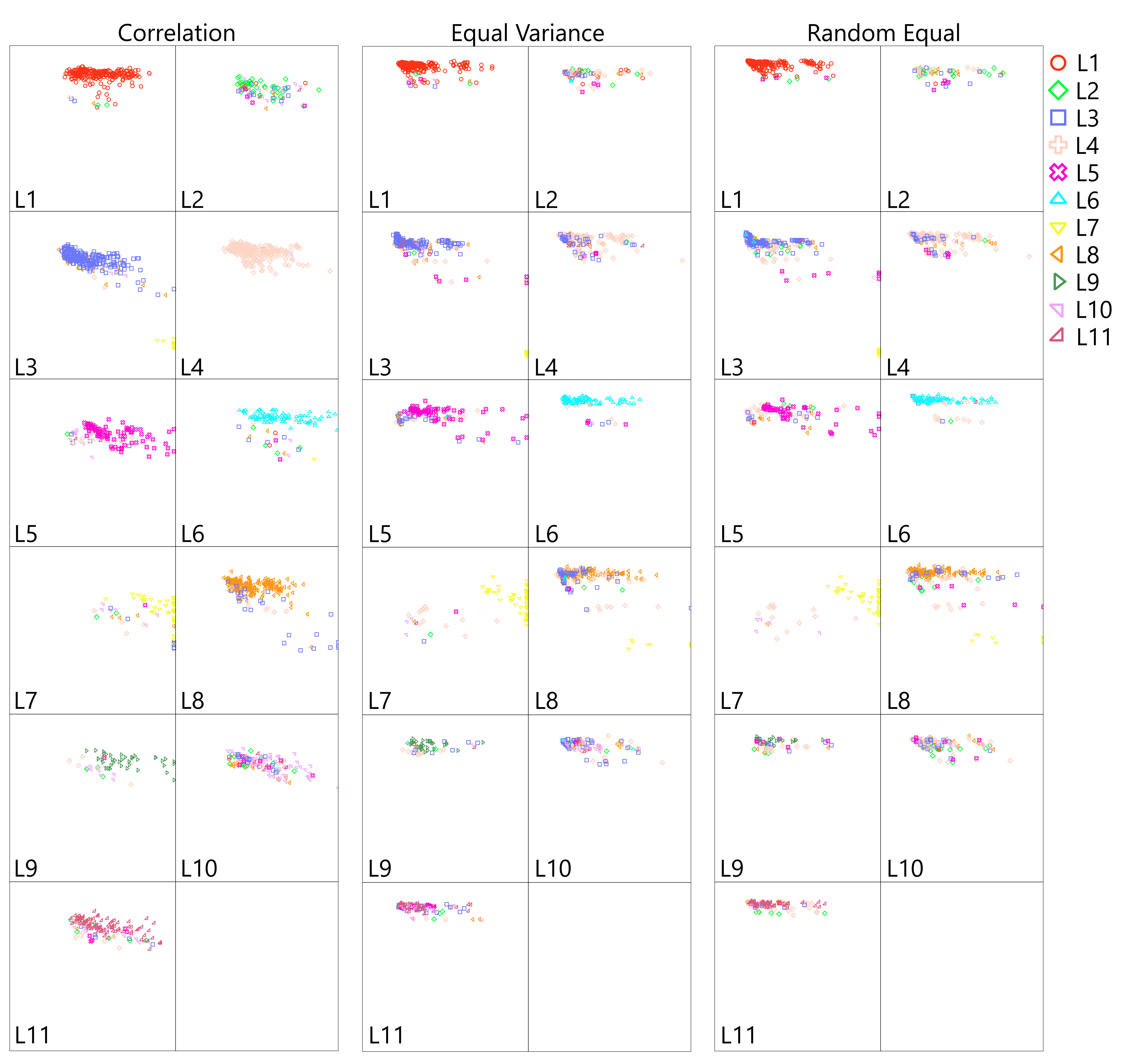}
		\caption{R-S plots of clusters generated from CCP-based classification using correlation partition, equal variance partition, and random equal partition of the Carcinoma dataset at $N=46$.}
	\end{subfigure}
	\caption{ Comparing the CCP-based classification effectiveness using correlation partition, equal random partition, and equal variance partition of the features of the Carcinoma dataset. For FRI, exponential kernel with $\kappa = 2$ and $\tau = 2.0$ was used. For each test, the results of all 10 seeds were plotted. From left to right: R-S plots of correlation partition, equal variance partition, and random equal partition. The $x$-axis is the residual score, and the $y$-axis is the similarity score. Each section corresponds to one cluster and the data were colored according to the predicted labels from $k$-NN.}
	\label{fig:carcinom}
\end{figure}

\autoref{fig:carcinom}(a) shows the accuracies of CCP-based classifications of  the Carcinoma dataset under various CCP reduced dimensions $N$ equipped with 3 feature partition schemes. The correlation partition outperforms both equal random and equal variance partitions over all $N$ values. In addition, as shown in \autoref{fig:carcinom}(b), 
for the  R-S plots, the correlation partition outperforms the other two partitions in each cluster at $N=46$.  
	
\subsubsection{Geometric shape, Residue-Similarity (R-S), and topological analysis }

In this section, we compare the 3D shape, R-S plot, and topological persistence of the TCGA-PANCAN data. For the comparison, CCP was used to reduce the data to $N=3$ components. The data were divided according to their true labels into 5 classes. The 5-fold cross-validation was used to obtain the predicted labels for visualization (coloring). 

For the 3D shape visualization, after the dimensionality reduction, the  Gaussian surface was used to generate the volumetric representation. The Chimera \cite{pettersen2021ucsf} was used to visualize the shape of data at the isovalue of 0.1. The surface was colored according to the predicted labels. 

For the persistence plot, after the dimensionality reduction, the data was divided according to their true labels. The HERMES package \cite{wang2021hermes} with the  $\alpha$ complex  was used to generate topological dimensions 0 (Betti-0), 1 (Betti-1), and 2 (Betti-2) curves and the corresponding smallest non-zero eigenvalue curves. 
Note that persistent Laplacian itself offers low-dimensional geometric and topological representations of the original high-dimensional data \cite{chen2022persistent}.

\begin{figure}[H]
	\centering
	\begin{minipage}{0.48\linewidth}
		\begin{subfigure}{\linewidth}
			\centering
			\includegraphics[height=5cm,width=6cm]{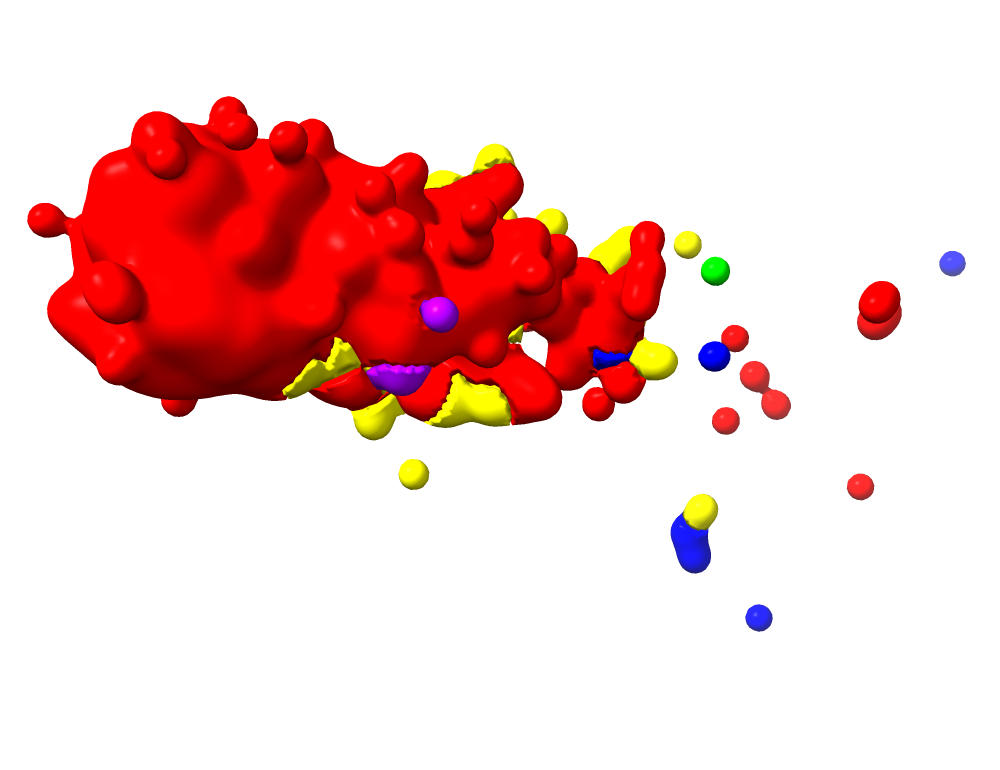}
			\caption{Gaussian surface of TCGA-PANCAN class 1.}
		\end{subfigure}
		
		\begin{subfigure}{\linewidth}
			\centering
			\includegraphics[height=5cm,width=6cm]{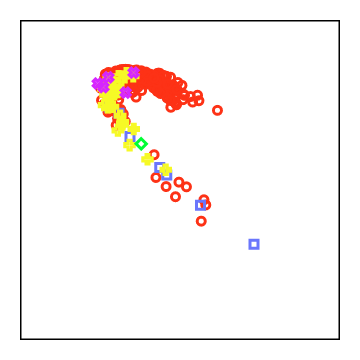}
			\caption{R-S plot of TCGA-PANCAN class 1.}
		\end{subfigure}
	\end{minipage}
	\begin{minipage}{0.48\linewidth}
		\begin{subfigure}{\linewidth}
			\centering
			\includegraphics[height=10.7 cm,width=8cm]{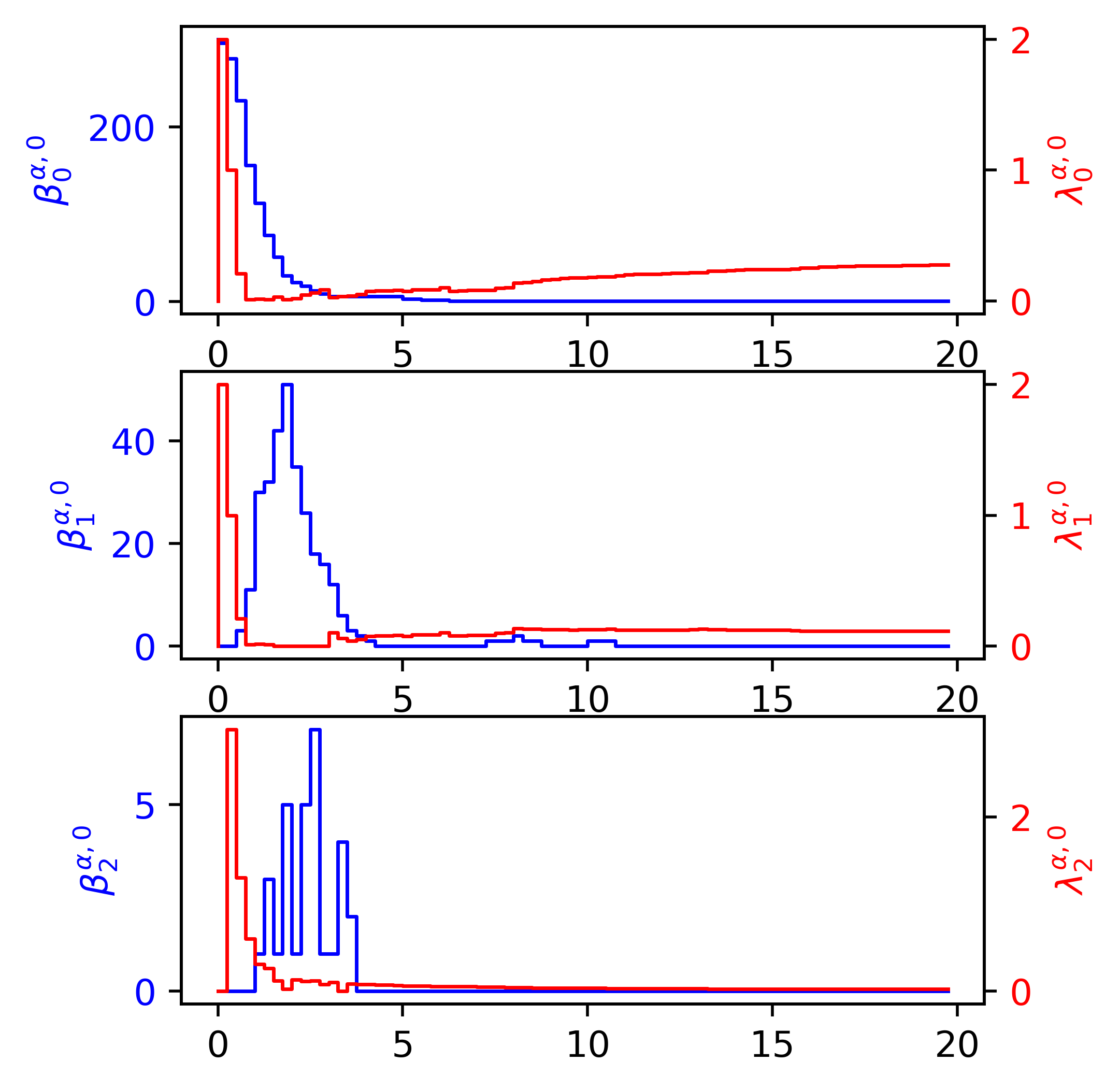}
			\caption{Persistence of TCGA-PANCAN class 1.}
	 	\end{subfigure}
	\end{minipage}
	\caption{Shape of data, R-S and persistence visualization of TCGA-PANCAN class 1 data. CCP was used to reduce the data to $N=3$. (a) Shape of data  was visualized with isovalue  0.1 in ChimeraX \cite{pettersen2021ucsf}. Red color indicates the correctly classified data. (b) R-S plot of class 1. Red circle is the correct label. The $x$ and $y$-axes correspond to the residue and similarity scores, respectively. (c) Visualization of the smallest non-zero eigenvalue curves along the filtration (indicated by red color) $\lambda_0^{\alpha, 0}, \lambda_1^{\alpha, 0},$ and $\lambda_2^{\alpha, 0}$ , and the harmonic spectral curves (indicated by blue color) $\beta_0^{\alpha, 0}, \beta_1^{\alpha, 0},$ and $  \beta_2^{\alpha, 0}$ for    class 1. HERMES package \cite{wang2021hermes} with the $\alpha$ complex was used to calculate the harmonic and non-harmonic spectra. The $x$-axis is the filtration radius. The left $y$-axis corresponds to the $\beta_0^{\alpha, 0}, \beta_1^{\alpha, 0}, $and $ \beta_2^{\alpha, 0}$ from top to bottom, and the right $y$-axis corresponds to  $\lambda_0^{\alpha, 0}, \lambda_1^{\alpha, 0}, $ and $ \lambda_2^{\alpha, 0}$ from top to bottom.}
	\label{fig: tcgapancan class1}
\end{figure}

\autoref{fig: tcgapancan class1} shows the 3 different visualizations of class 1. Notice that the shape analysis shows predominately red regions or dots mixed with misclassified labels. We can see this mixing in the R-S plot as well.  The yellow points have lower R scores, indicating that these samples are more likely to be mislabeled in machine learning. We can see that blue points with low S scores are isolated in the shape visualization.

 Note that  $\beta_0^{\alpha, 0}$,  $\beta_1^{\alpha, 0}$ and  $\beta_2^{\alpha, 0}$ offer the same information as persistent homology does for the data. The $\beta_0^{\alpha, 0}$ shows there are about 290 samples in this class that become fully connected at radius 6 ( $\beta_0^{\alpha, 0}=1$). The $\beta^{\alpha, 0}_1$ shows there are many cycles in the sample. The $\beta^{\alpha, 0}_1$ indicates there are at most 7 cavities. There are no topological changes in the data after radius=11. However, the smallest non-zero eigenvalue ($\lambda^{\alpha, 0}_0$) keeps changing as the filtration radius increases, indicating that persistent Laplacian reveals more information about the data than persistent homology does.  

Finally, it is noticed that most misclassified samples have relatively low R-S scores. This observation indicates the effectiveness of our R-S scores and indexes.  

\begin{figure}[H]
	\centering
	\begin{minipage}{0.48\linewidth}
		\begin{subfigure}{\linewidth}
			\centering
			\includegraphics[height=5cm,width=6cm]{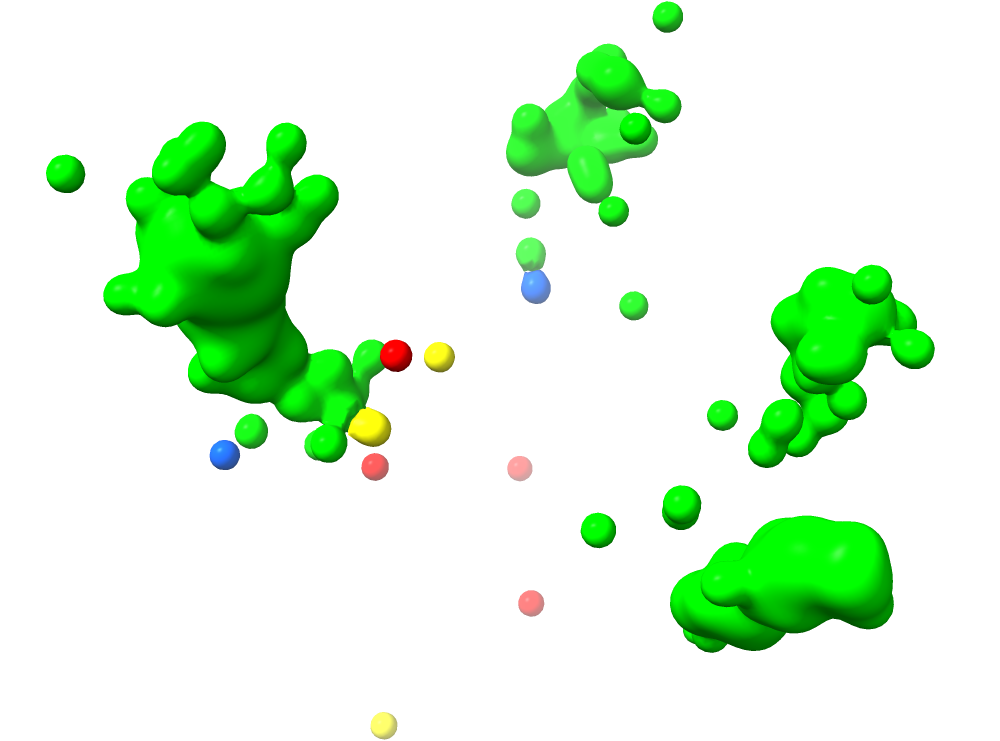}
			\caption{Gaussian surface of TCGA-PANCAN class 2.}
		\end{subfigure}
		
		\begin{subfigure}{\linewidth}
			\centering
			\includegraphics[height=5cm,width=6cm]{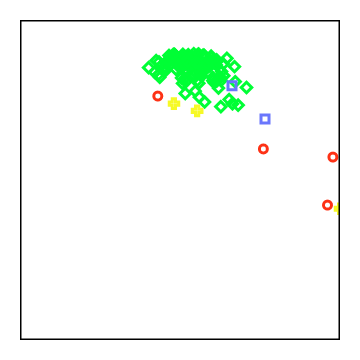}
			\caption{R-S plot of TCGA-PANCAN class 2.}
		\end{subfigure}
	\end{minipage}
	\begin{minipage}{0.48\linewidth}
		 \begin{subfigure}{\linewidth}
			\centering
			\includegraphics[height=10.7 cm,width=8cm]{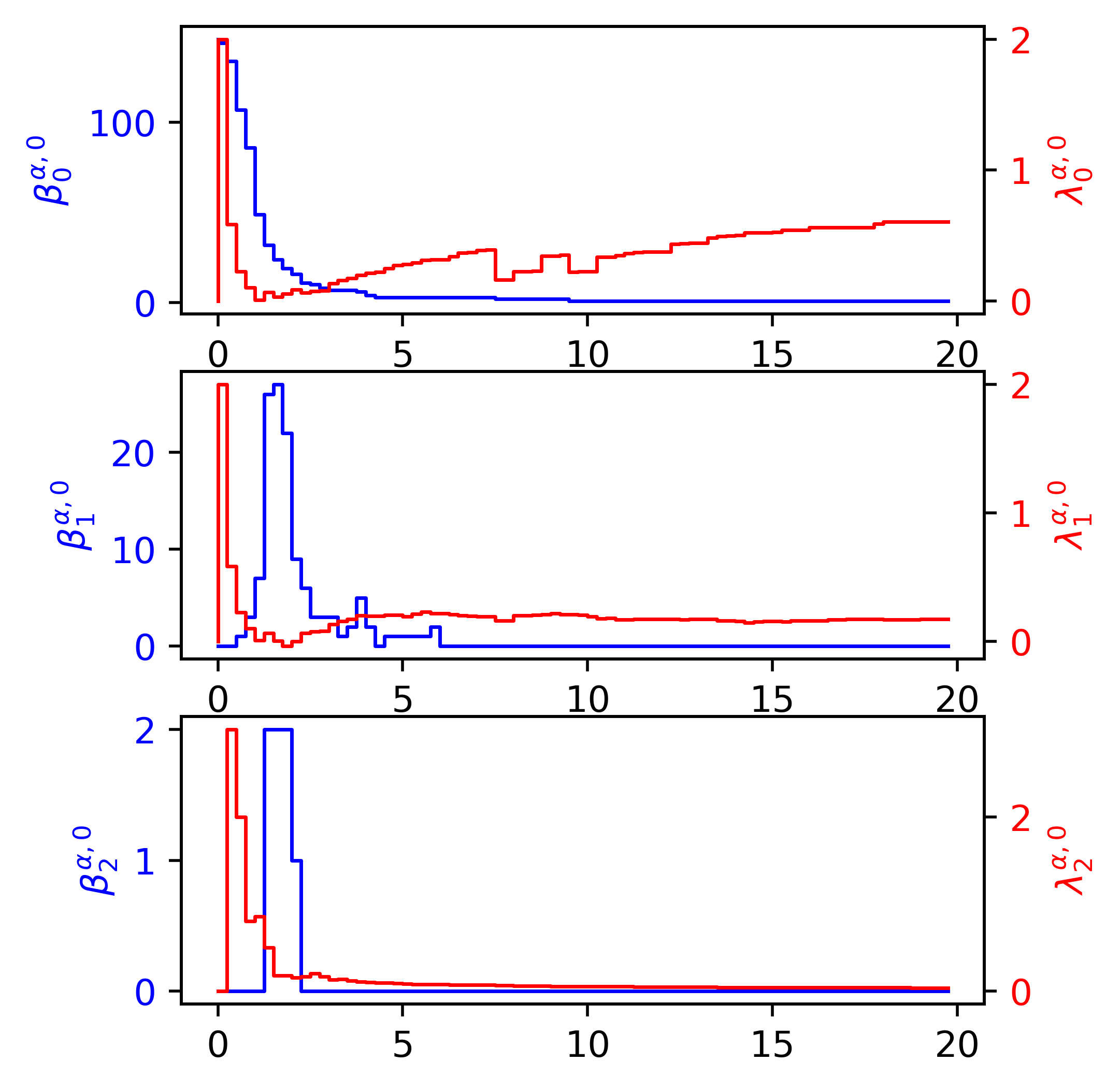}
			\caption{Persistence of TCGA-PANCAN class 2.}
	 \end{subfigure}
	\end{minipage}
	\caption{Shape of data, R-S and persistence visualization of TCGA-PANCAN class 2 data. CCP was used to reduce the data to $N=3$. (a) Shape of data  was visualized with isovalue  0.1 in ChimeraX \cite{pettersen2021ucsf}.  Green color indicates the correctly classified data. (b) R-S plot of class 2.  Green circle is the correct label. The $x$ and $y$-axes correspond to the residue and similarity scores, respectively. (c) Visualization of the smallest non-zero eigenvalue curves along the filtration (indicated by red color) $\lambda_0^{\alpha, 0}, \lambda_1^{\alpha, 0},$ and $\lambda_2^{\alpha, 0}$ , and the harmonic spectral curves (indicated by blue color) $\beta_0^{\alpha, 0}, \beta_1^{\alpha, 0},$ and $  \beta_2^{\alpha, 0}$ for    class 2. HERMES package \cite{wang2021hermes} with the $\alpha$ complex was used to calculate the harmonic and non-harmonic spectra. The $x$-axis is the filtration radius. The left $y$-axis corresponds to the $\beta_0^{\alpha, 0}, \beta_1^{\alpha, 0}, $and $ \beta_2^{\alpha, 0}$ from top to bottom, and the right $y$-axis corresponds to  $\lambda_0^{\alpha, 0}, \lambda_1^{\alpha, 0}, $ and $ \lambda_2^{\alpha, 0}$ from top to bottom.}
	\label{fig: tcgapancan class2}
\end{figure}

 The shape, R-S, and topological analysis of class 2 are given in \autoref{fig: tcgapancan class2}.  The $\beta_0^{\alpha, 0}$ indicates class 2 has about 130 samples, which become fully connected near radius=10. The $\lambda_0^{\alpha, 0}$ curve shows a significant discontinuity at radius near radius=10.  The $\beta_1^{\alpha, 0}$ shows about 28 cycles at its peak value. At most two cavities in data have been found in $\beta_2^{\alpha, 0}$ at a given filtration. The shape shows four major pieces and only a few samples were mislabeled by machine learning. R-S plots should have four different labels in this class. Most of the samples were correctly predicted, which is consistent with the shape analysis. Most class 1 labels (red ones) have lower S scores, indicating that they do not belong.

Most misclassified samples have low R-S scores and are disconnected from other samples in the class.


\begin{figure}[H]
	\centering
	\begin{minipage}{0.48\linewidth}
		\begin{subfigure}{\linewidth}
			\centering
			\includegraphics[height=5cm,width=6cm]{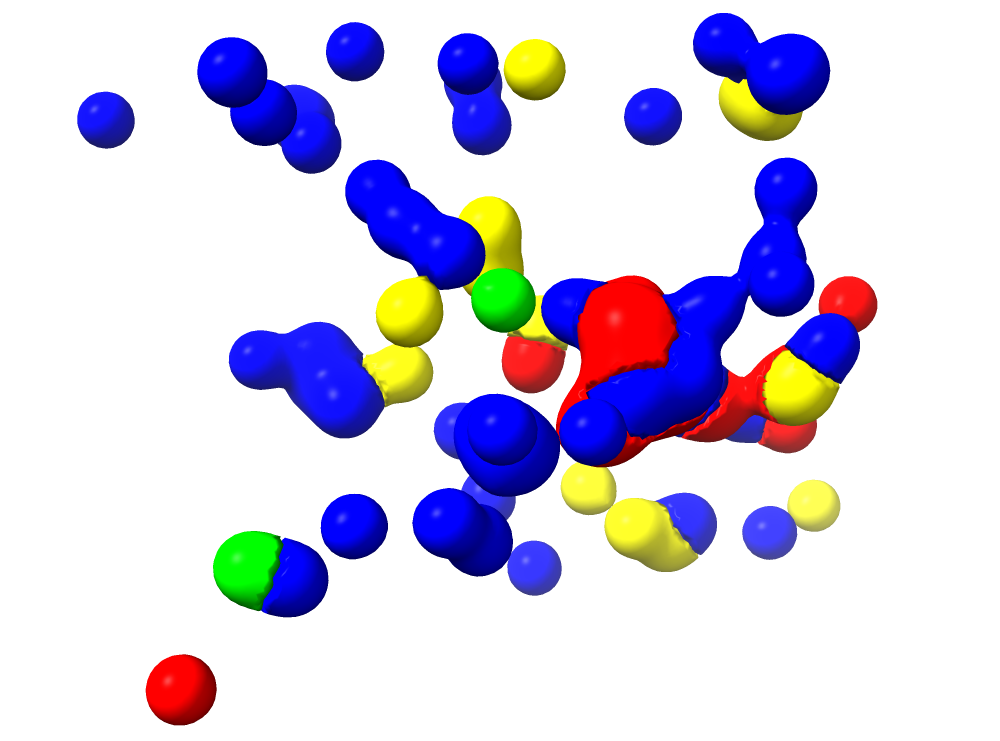}
			\caption{Gaussian surface of TCGA-PANCAN class 3.}
		\end{subfigure}
		
		\begin{subfigure}{\linewidth}
			\centering
			\includegraphics[height=5cm,width=6cm]{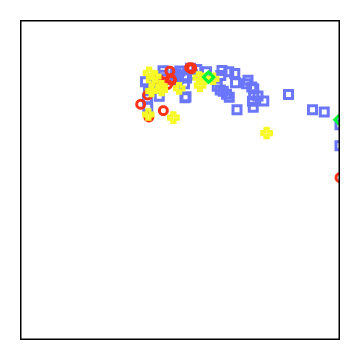}
			\caption{R-S plot of TCGA-PANCAN class 3.}
		\end{subfigure}
	\end{minipage}
	\begin{minipage}{0.48\linewidth}
		 \begin{subfigure}{\linewidth}
			\centering
			\includegraphics[height=10.7cm,width=8cm]{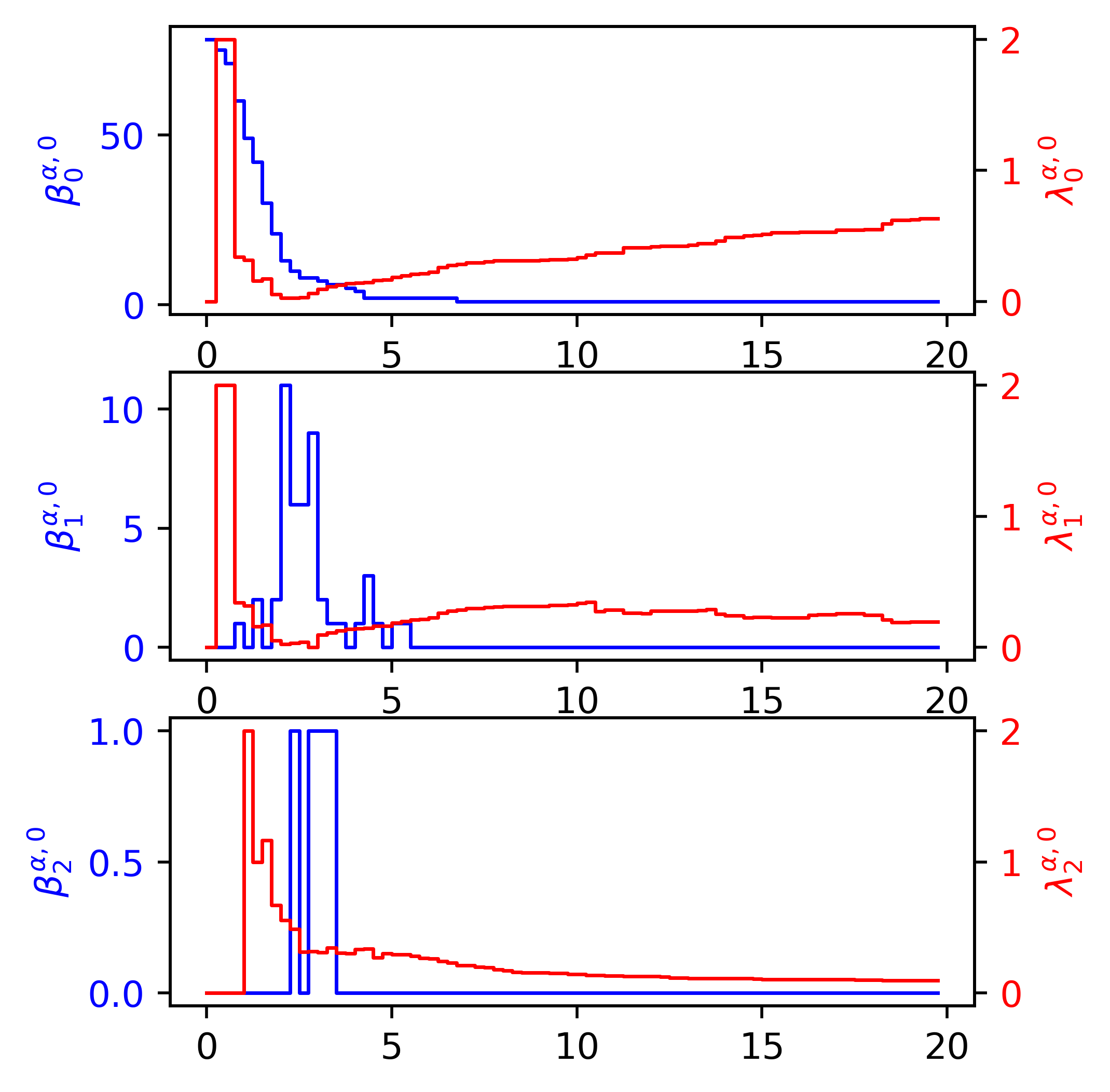}
			\caption{Persistence of TCGA-PANCAN class 3.}
	 	\end{subfigure}
	\end{minipage}
	\caption{Shape of data, R-S and persistence visualization of TCGA-PANCAN class 3 data. CCP was used to reduce the data to $N=3$. (a) Shape of data  was visualized with isovalue  0.1 in ChimeraX \cite{pettersen2021ucsf}.   Blue  color indicates the correctly classified data. (b) R-S plot of class 3.   Blue  circle is the correct label. The $x$ and $y$-axes correspond to the residue and similarity scores, respectively. (c) Visualization of the smallest non-zero eigenvalue curves along the filtration (indicated by red color) $\lambda_0^{\alpha, 0}, \lambda_1^{\alpha, 0},$ and $\lambda_2^{\alpha, 0}$ , and the harmonic spectral curves (indicated by blue color) $\beta_0^{\alpha, 0}, \beta_1^{\alpha, 0},$ and $  \beta_2^{\alpha, 0}$ for    class 3. HERMES package \cite{wang2021hermes} with the $\alpha$ complex was used to calculate the harmonic and non-harmonic spectra. The $x$-axis is the filtration radius. The left $y$-axis corresponds to the $\beta_0^{\alpha, 0}, \beta_1^{\alpha, 0}, $and $ \beta_2^{\alpha, 0}$ from top to bottom, and the right $y$-axis corresponds to  $\lambda_0^{\alpha, 0}, \lambda_1^{\alpha, 0}, $ and $ \lambda_2^{\alpha, 0}$ from top to bottom.}
	\label{fig: tcgapancan class3}
\end{figure}

 \autoref{fig: tcgapancan class3} gives three types of analyses for class 3. 
This class has only about 78 samples, as shown by the $\beta_0^{\alpha, 0}$ curve. At filtration radius=2, there were 11 one-dimensional holes in the data. There were only two cavities found by $\beta_2^{\alpha, 0}$. The shape plot indicates most samples are disconnected at isovalue 0.1 but merge at radius 7 as detected by $\beta_0^{\alpha, 0}$. The yellow labels are close to the red ones, as shown by the shape and R-S plots.   
The $\lambda_1^{\alpha, 0}$ curve demonstrates a few discontinuities after topological persistence ($\beta_1^{\alpha, 0}$) becomes flat, indicating important homotopic events in the data. As in other classes,  most misclassified samples have relatively low R-S scores.


\begin{figure}[H]
	\centering
	\begin{minipage}{0.48\linewidth}
		\begin{subfigure}{\linewidth}
			\centering
			\includegraphics[height=5cm,width=6cm]{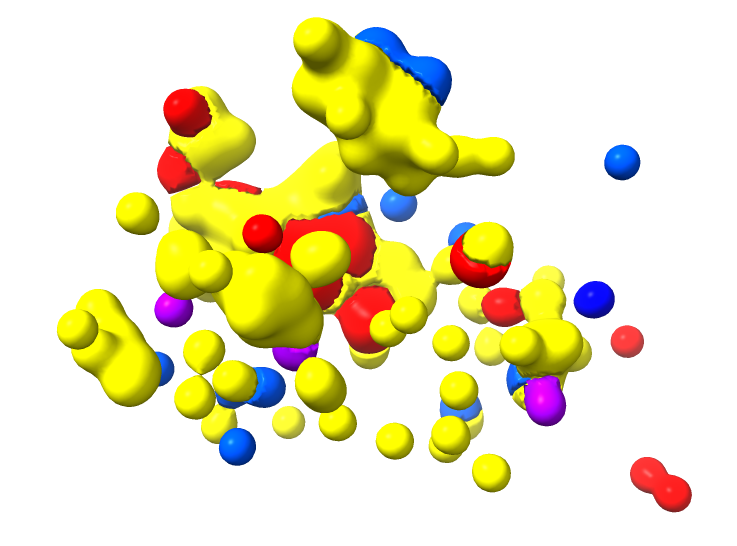}
			\caption{Gaussian surface of TCGA-PANCAN class 4.}
		\end{subfigure}
		
		\begin{subfigure}{\linewidth}
			\centering
			\includegraphics[height=5cm,width=6cm]{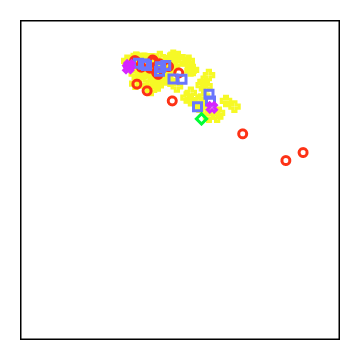}
			\caption{R-S plot of TCGA-PANCAN class 4.}
		\end{subfigure}
	\end{minipage}
	\begin{minipage}{0.48\linewidth}
		 \begin{subfigure}{\linewidth}
			\centering
			\includegraphics[height=10.7 cm,width=8cm]{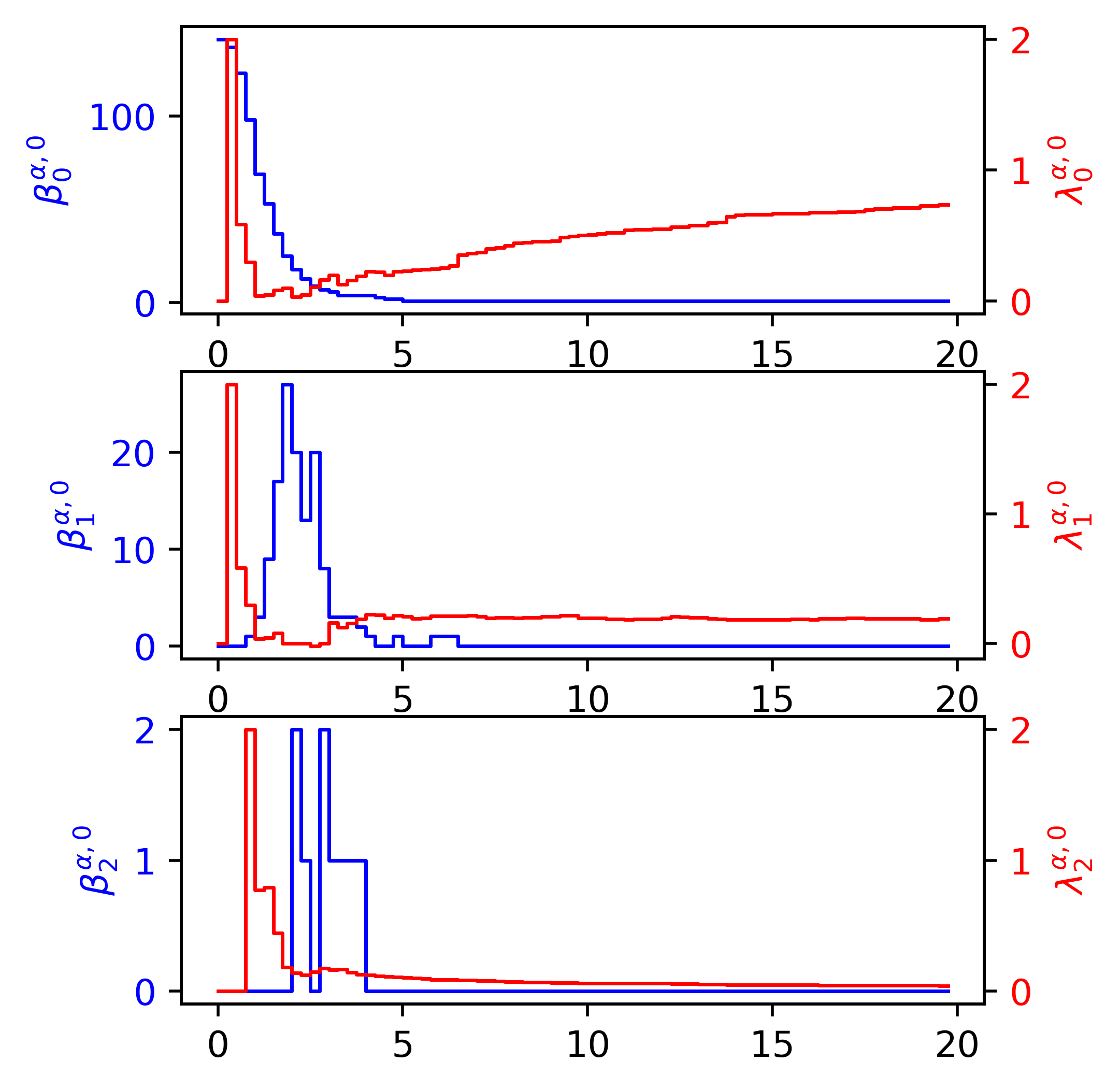}
			\caption{Persistence of TCGA-PANCAN class 4.}
	 	\end{subfigure}
	\end{minipage}
	\caption{Shape of data, R-S and persistence visualization of TCGA-PANCAN class 4 data. CCP was used to reduce the data to $N=3$. (a) Shape of data  was visualized with isovalue  0.1 in ChimeraX \cite{pettersen2021ucsf}.   Yellow color indicates the correctly classified data. (b) R-S plot of class 4.  Yellow  circle is the correct label. The $x$ and $y$-axes correspond to the residue and similarity scores, respectively. (c) Visualization of the smallest non-zero eigenvalue curves along the filtration (indicated by red color) $\lambda_0^{\alpha, 0}, \lambda_1^{\alpha, 0},$ and $\lambda_2^{\alpha, 0}$ , and the harmonic spectral curves (indicated by blue color) $\beta_0^{\alpha, 0}, \beta_1^{\alpha, 0},$ and $  \beta_2^{\alpha, 0}$ for    class 4. HERMES package \cite{wang2021hermes} with the $\alpha$ complex was used to calculate the harmonic and non-harmonic spectra. The $x$-axis is the filtration radius. The left $y$-axis corresponds to the $\beta_0^{\alpha, 0}, \beta_1^{\alpha, 0}, $and $ \beta_2^{\alpha, 0}$ from top to bottom, and the right $y$-axis corresponds to  $\lambda_0^{\alpha, 0}, \lambda_1^{\alpha, 0}, $ and $ \lambda_2^{\alpha, 0}$ from top to bottom.}
	\label{fig: tcgapancan class4}
\end{figure}

In \autoref{fig: tcgapancan class4}, the $\beta_0^{\alpha, 0}$ curve suggests 150 samples in class 4 (yellow). Some samples are misclassified as class 1 (red), class 3 (blue), and  class 5 (purple) as shown in shape and R-S plots. The topological persistence indicates many topological invariants along the filtration axis, which can be a faithful representation of the data \cite{chen2022persistent}. Specifically, all data points overlap at radius 5, as shown by $\beta_0^{\alpha, 0}$. 
However, $\lambda_0^{\alpha, 0}$ still indicates a discontinuity at radius 7. 
All cycles disappear after radius 7 as revealed by $\beta_1^{\alpha, 0}$. The last cycle persists from radius 6 to 7. The $\beta_2^{\alpha, 0}$ curve becomes flat at radius 4. The misclassified red samples show low S scores.


\begin{figure}[H]
	\centering
	\begin{minipage}{0.48\linewidth}
		\begin{subfigure}{\linewidth}
			\centering
			\includegraphics[height=5cm,width=6cm]{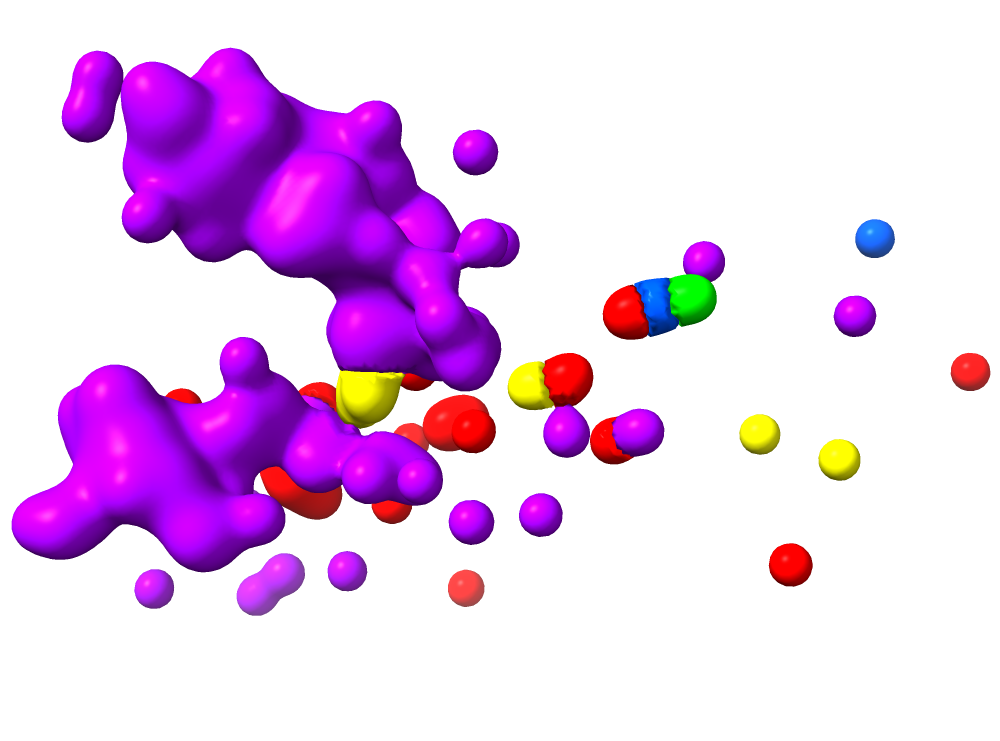}
			\caption{Gaussian surface of TCGA-PANCAN class 5.}
		\end{subfigure}
		
		\begin{subfigure}{\linewidth}
			\centering
			\includegraphics[height=5cm,width=6cm]{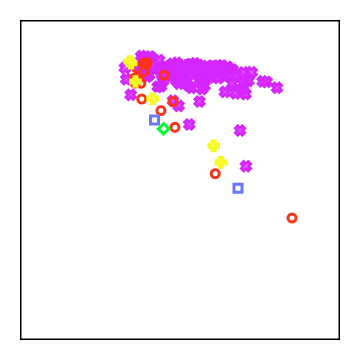}
			\caption{R-S plot of TCGA-PANCAN class 5.}
		\end{subfigure}
	\end{minipage}
	\begin{minipage}{0.48\linewidth}
		 \begin{subfigure}{\linewidth}
			\centering
			\includegraphics[height=10.7 cm,width=8cm]{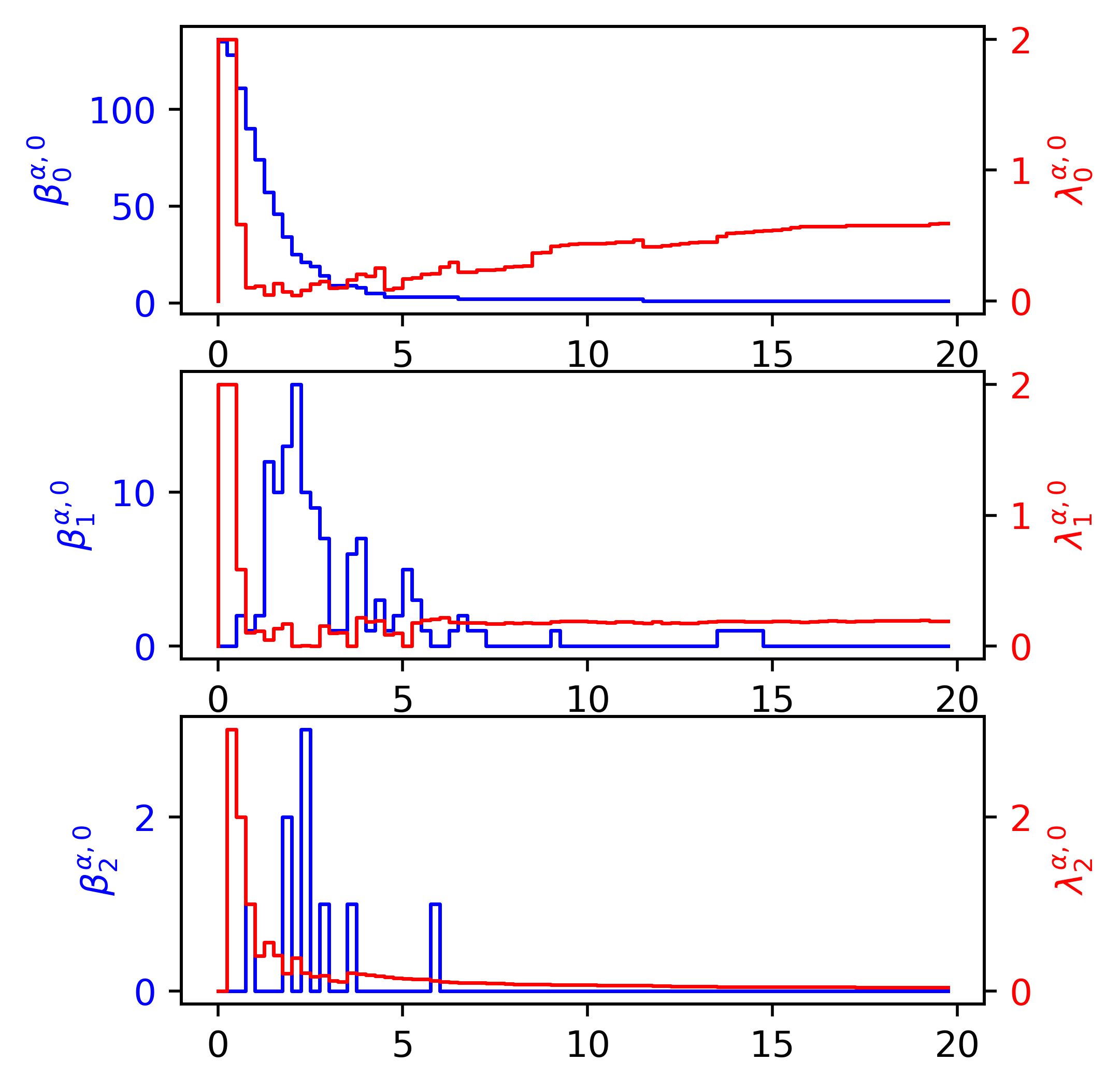}
			\caption{Persistence of TCGA-PANCAN class 5.}
	 	\end{subfigure}
	\end{minipage}
	\caption{Shape of data, R-S and persistence visualization of TCGA-PANCAN class 5 data. CCP was used to reduce the data to $N=3$. (a) Shape of data  was visualized with isovalue  0.1 in ChimeraX \cite{pettersen2021ucsf}.   Purple color indicates the correctly classified data. (b) R-S plot of class 5.  Purple  circle is the correct label. The $x$ and $y$-axes correspond to the residue and similarity scores, respectively. (c) Visualization of the smallest non-zero eigenvalue curves along the filtration (indicated by red color) $\lambda_0^{\alpha, 0}, \lambda_1^{\alpha, 0},$ and $\lambda_2^{\alpha, 0}$ , and the harmonic spectral curves (indicated by blue color) $\beta_0^{\alpha, 0}, \beta_1^{\alpha, 0},$ and $  \beta_2^{\alpha, 0}$ for    class 5. HERMES package \cite{wang2021hermes} with the $\alpha$ complex was used to calculate the harmonic and non-harmonic spectra. The $x$-axis is the filtration radius. The left $y$-axis corresponds to the $\beta_0^{\alpha, 0}, \beta_1^{\alpha, 0}, $ and $ \beta_2^{\alpha, 0}$ from top to bottom, and the right $y$-axis corresponds to  $\lambda_0^{\alpha, 0}, \lambda_1^{\alpha, 0}, $ and $ \lambda_2^{\alpha, 0}$ from top to bottom.}
	\label{fig: tcgapancan class5}
\end{figure}

 \autoref{fig: tcgapancan class5} illustrates our shape, R-S, and topological analyses of class 5. Although $\beta_0^{\alpha, 0}$ indicates there are only about 140 samples,  class 5 is very rich in its topological persistence. The $\beta_1^{\alpha, 0}$ shows the data points did not connect before the filtration radius reached 12. The $\beta_1^{\alpha, 0}$ curve indicates a large one-dimensional hole from radius 13.5 to 15. The $\beta_2^{\alpha, 0}$ curve shows 9 short-living cavities in the data. It is clear from the R-S plot that samples having low R-S scores are more likely to be mislabeled.

\begin{figure}[H]
	\centering
	\begin{subfigure}{0.245\textwidth}
		\centering
		\includegraphics[width = \textwidth]{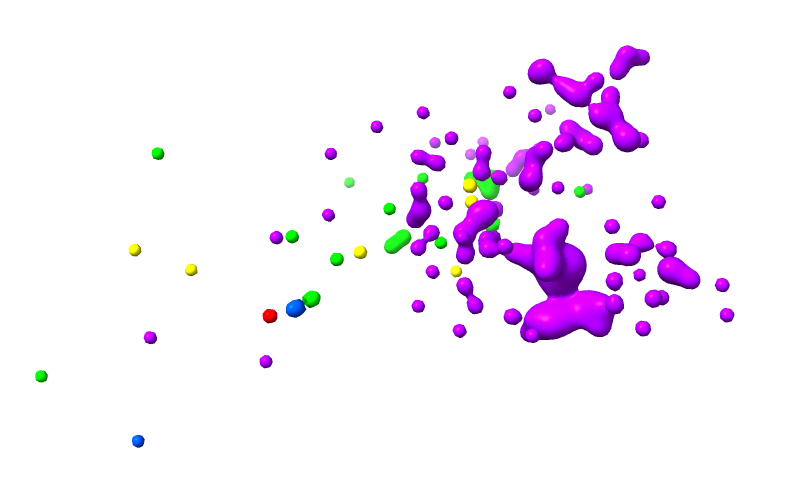}
		\caption{Isovalue 0.235}
	\end{subfigure}
	\begin{subfigure}{0.245\textwidth}
		\centering
		\includegraphics[width = \textwidth]{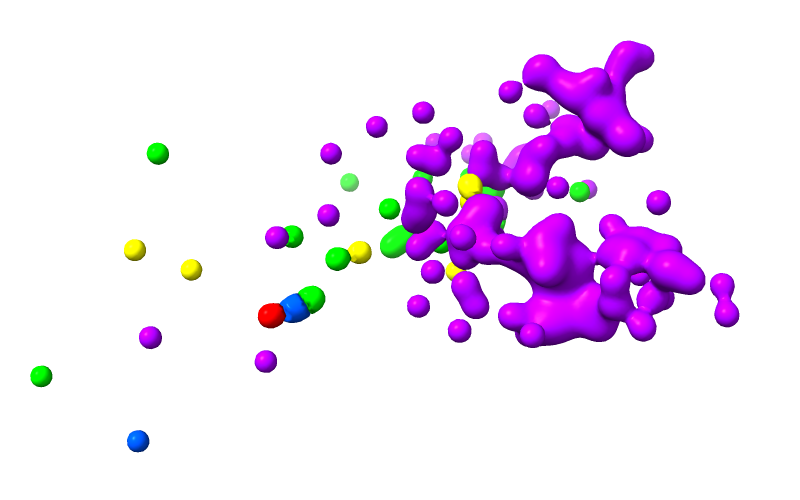}
		\caption{Isovalue 0.139}
	\end{subfigure}
	\begin{subfigure}{0.245\textwidth}
		\centering
		\includegraphics[width = \textwidth]{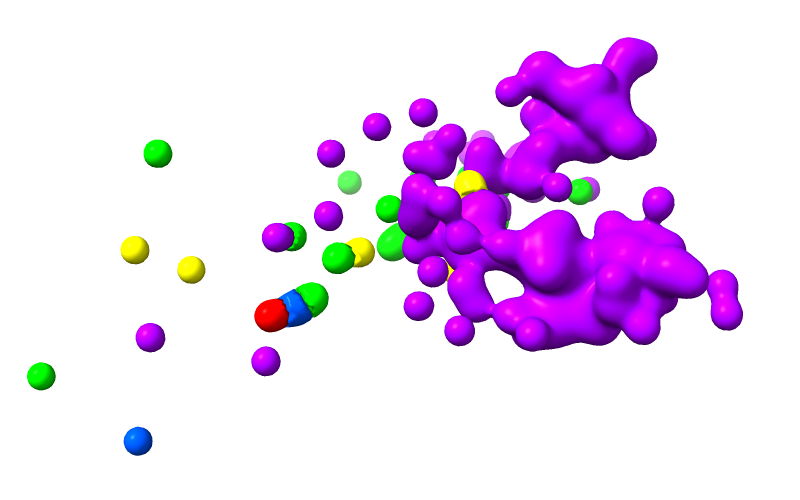}
		\caption{Isovalue 0.0963}
	\end{subfigure}
	\begin{subfigure}{0.245\textwidth}
		\centering
		\includegraphics[width = \textwidth]{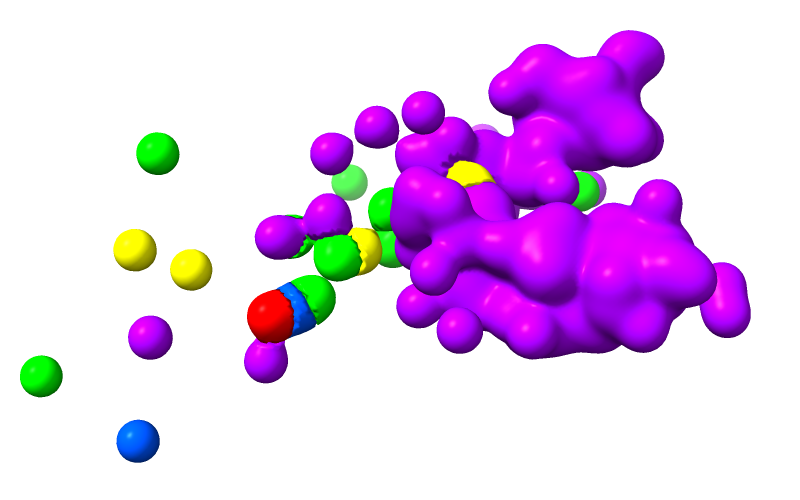}
		\caption{Isovalue 0.0426}
	\end{subfigure}
	\caption{Shape of class 5 of TCGA-PANCAN dataset visualized in multiscale using  ChimeraX \cite{pettersen2021ucsf} when isovalues were varied. The different colors indicated the predicted labels, and purple is the true label of class 5.
	}
	\label{fig: tcgapancan class 5 isosurface}
\end{figure}

Because \autoref{fig: tcgapancan class5}'s persistence plot indicates interesting topological features, and class 5 was further visualized in \autoref{fig: tcgapancan class 5 isosurface} with varying isovalues or  scales. We can see that from isovalues 0.235 to 0.139, 2 holes form in the bottom right corner. We can also see another hole beginning to form in the bottom center.  From isovalues 0.139 to 0.0963, the two holes are no longer visible, but the hole that was forming is now completed. This corresponds to  $\beta_1^{\alpha, 0}$. The voids are short-lived, as shown by $ \beta_2^{\alpha, 0}$ in \autoref{fig: tcgapancan class5} and cannot be visible in the isosurface. Decreasing the isovalue further to 0.0426 shows the combination of two main parts of the data, which would stabilize the structure. This corresponds to the decrease of $\beta_0^{\alpha, 0}$ because the number of components is decreasing, but the increase in $\lambda_0^{\alpha ,0}$ indicates that the structure is more stable.

\subsection{Comparison with other dimensionality reduction methods}
 
In this section, we compare CCP's performance with  UMAP, PCA, LLE, and Isomap on ALL-AML, TCGA-PANCAN, Coil-20, and Coil-100 datasets. For each dataset, we performed 5-fold or 10-fold cross-validation depending on the size of the dataset to test the accuracy using $k$-nearest neighbors ($k$-NN). Results of all 10 random seeds were used in performance evaluation.

\subsubsection{ALL-AML}

\begin{figure}[H]
	\centering
	\includegraphics[scale = 0.45]{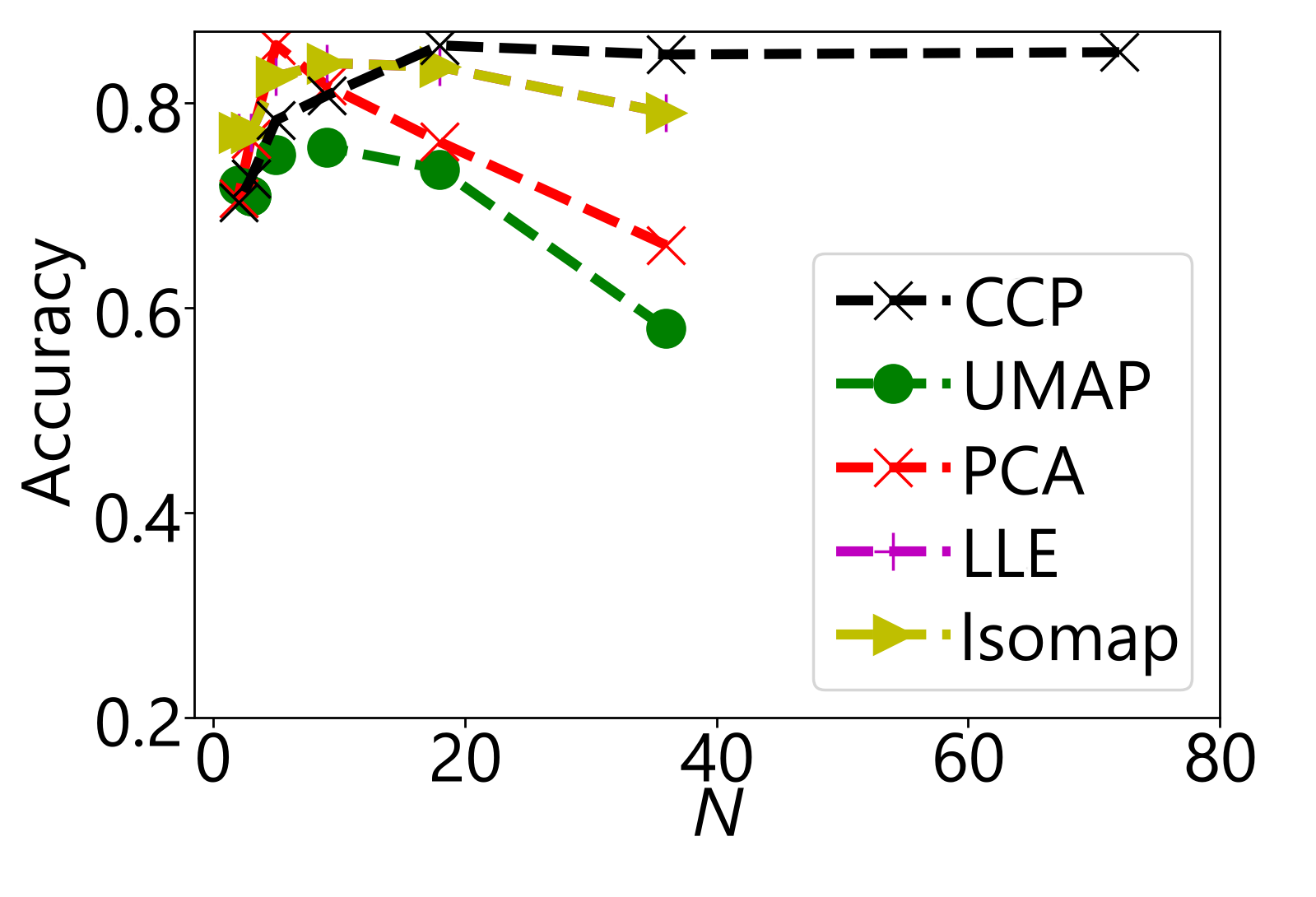}
	\caption{Accuracy of  $k$-NN classification of the ALL-AML dataset when the dimension is reduced to $N$ by using CCP, UMAP, PCA, LLE, and Isomap. Here, a 5-fold cross-validation with 10 random seedings was used. Test-train split was done prior to the dimensionality reduction. For CCP, exponential kernel with $\kappa = 1$ and $\tau = 2.0$ was used. The sample size, feature size, and the number of classes of the ALLAML dataset are 72, 7129, and 2, respectively.}
	\label{fig: ALLAML acc}
\end{figure}

\begin{figure}[H]
	\centering
	\includegraphics[scale = 0.45]{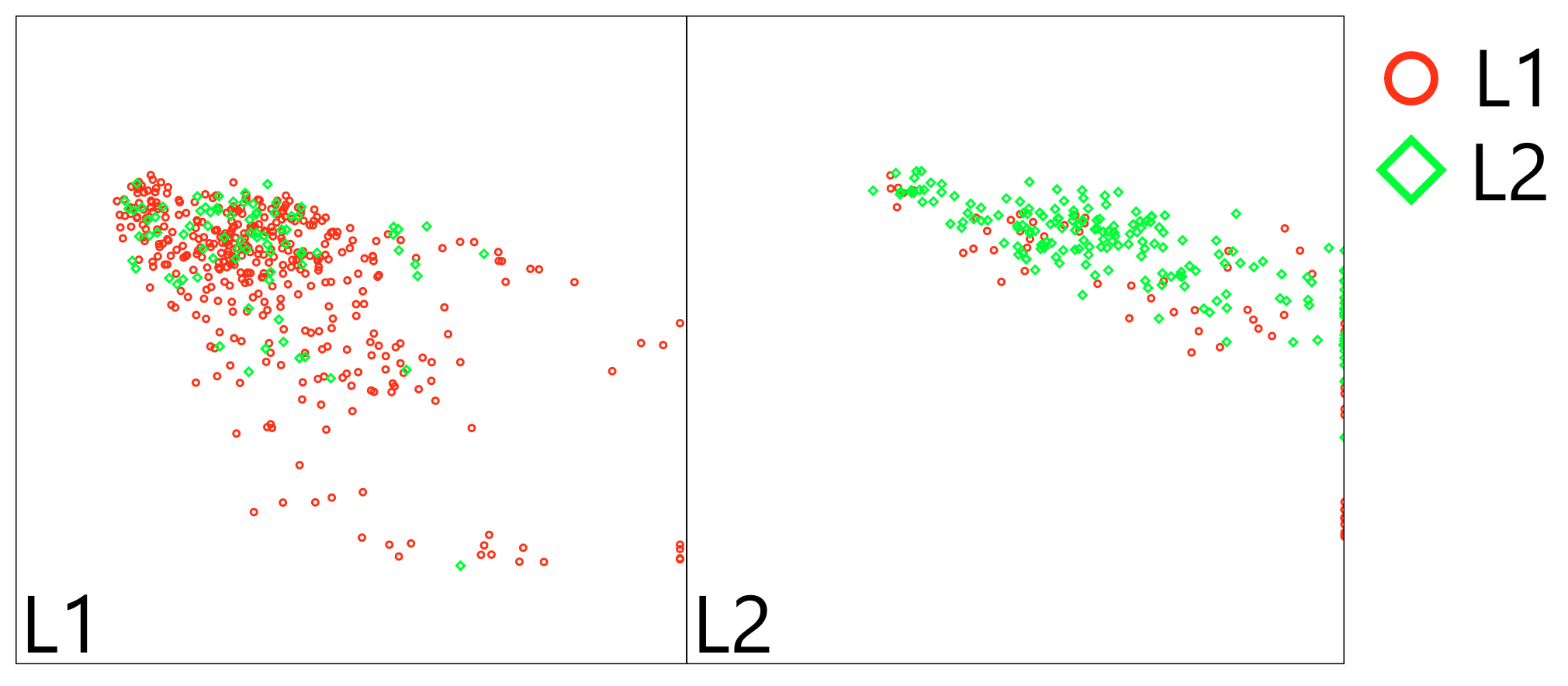}
	\caption{Visualization of the ALL-AML dataset when dimension reduced by CCP with exponential kernel, $\kappa = 1$ and $\tau = 2.0$ to $N = 36$. Each section represents a class and the samples were colored according to their predicted labels from the $k$-NN classification via 5-fold cross-validation. Results of all 10 seeds were used for the visualization. The $x$ and $y$ axes are the residue and similarity scores, respectively. The sample size, feature size, and the number of classes of the ALL-AML dataset are 72, 7129, and 2, respectively.}
	\label{fig: ALLAML CCP}
\end{figure}

The dimension of the ALL-AML dataset was reduced using an exponential kernel with $\kappa = 1$ and $\tau = 2.0$. \autoref{fig: ALLAML acc} shows the performance of CCP, UMAP, PCA, Isomap, and LLE. Here, a 5-fold cross-validation with 10 random seeds was used. CCP performs better than the other algorithms do and is stable with a wide range of $N$ values. All other methods show a drop in their accuracy beyond dimension $N = 36$. Since the ALL-AML dataset only has 72 samples, UMAP, PCA, LLE, and Isomap cannot reduce the ALL-AML dimension to $N > 72$ because their dimension is limited by the size of the matrix diagonalization.

\begin{figure}[H]
	\centering
	\includegraphics[scale = 0.45]{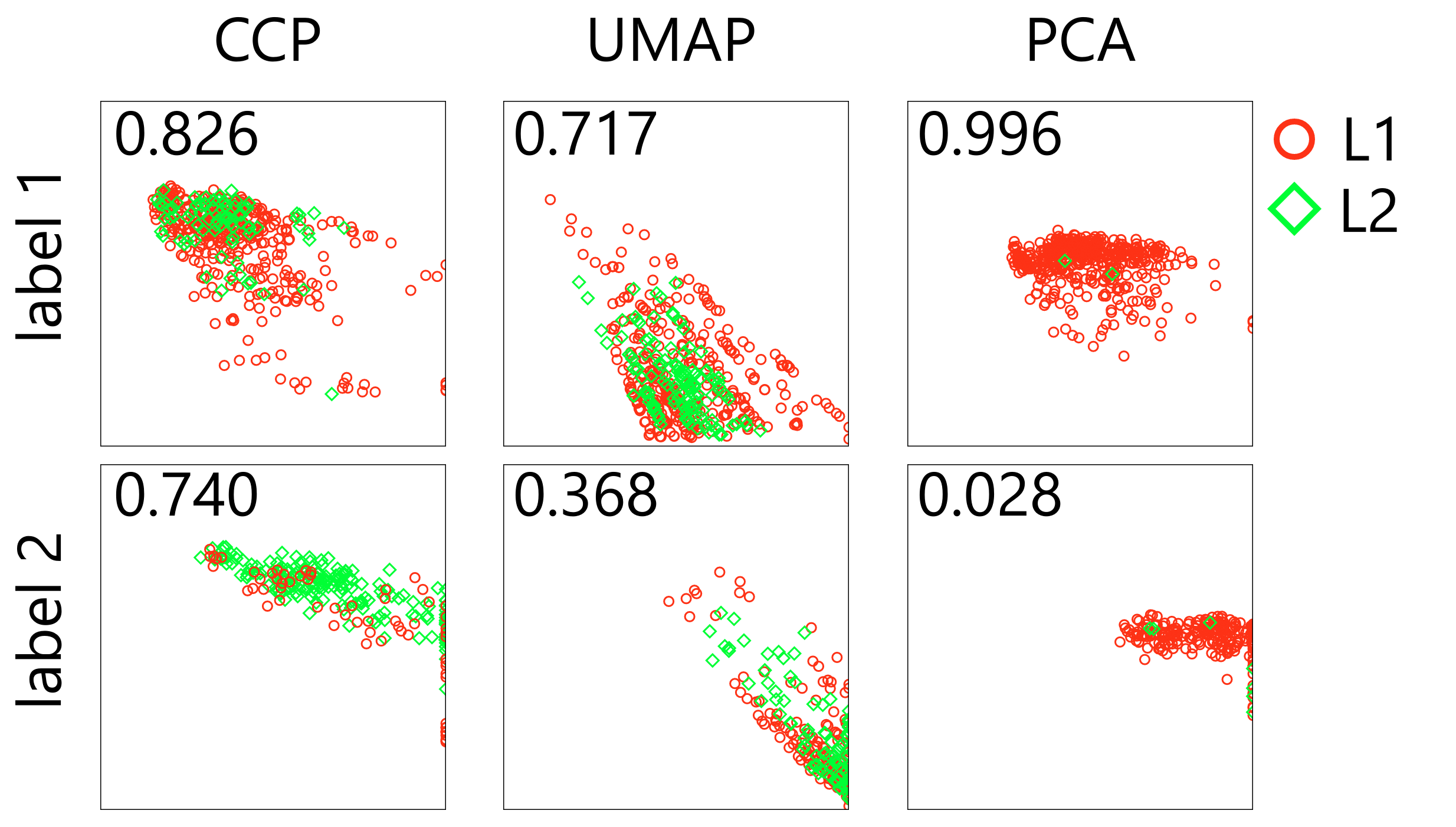}
	\caption{Visualization of the ALL-AML dataset when the dimension was reduced to $N = 36$ by using CCP, UMAP, and PCA. Since there are only 72 samples in the ALL-AML dataset, results from all 10 seeds were plotted, leading to 720 sample points in the plot.  For CCP, exponential kernel with $\kappa = 1$ and $\tau = 2.0$ was used. The $x$ and $y$ axes are the residue and similarity scores, respectively. Each row is for one class, and the data points are colored based on the predicted labels from the $k$-NN classifier, using 5-fold cross-validation. The sample size, feature size, and the number of classes of the ALL-AML are 72, 7129, and 2, respectively.}
	\label{fig: ALLAML visual}
\end{figure}

\autoref{fig: ALLAML CCP} shows the R-S plot of the ALL-AML dataset when the dimension is reduced by CCP to $N = 36$. The left and right sections correspond to the 2 classes. Samples were plotted according to their 36 features and colored with the predicted labels from $k$-NN. Results of all 10 seeds were plotted into one chart (i.e., 720 samples), and the residue and the similarity scores were calculated separately for each random seed. The $x$ and the $y$-axes are the residue and similarity scores, respectively. Class 2 has a better R-S distribution.

\autoref{fig: ALLAML visual} shows the R-S plot of ALL-AML when the feature dimension is reduced to $N = 36$ by using CCP, UMAP, and PCA. Results of all 10 random seeds were used in the visualization to obtain a better understanding of the performance, and the residue and similarity scores were computed separately for each seed. In each class, the samples were colored according to their predicted labels obtained from $k$-NN. The $x$-axis and $y$-axis of each R-S plot are the residue and similarity scores, respectively. The top row is class 1 (ALL), and the bottom row is class 2 (AML). The numerical number inside the plot is the accuracy for the class. Notice that UMAP's R-S plot indicates that UMAP's reduction has a low similarity score, which explains its low accuracy. On the other hand, PCA has higher accuracy than that UMAP, but most AML samples are mislabeled. This indicates that PCA is unable to distinguish the difference between ALL and AML when $N=36$.

\subsubsection{TCGA-PANCAN}

\begin{figure}[H]
	\centering
	\includegraphics[scale = 0.45]{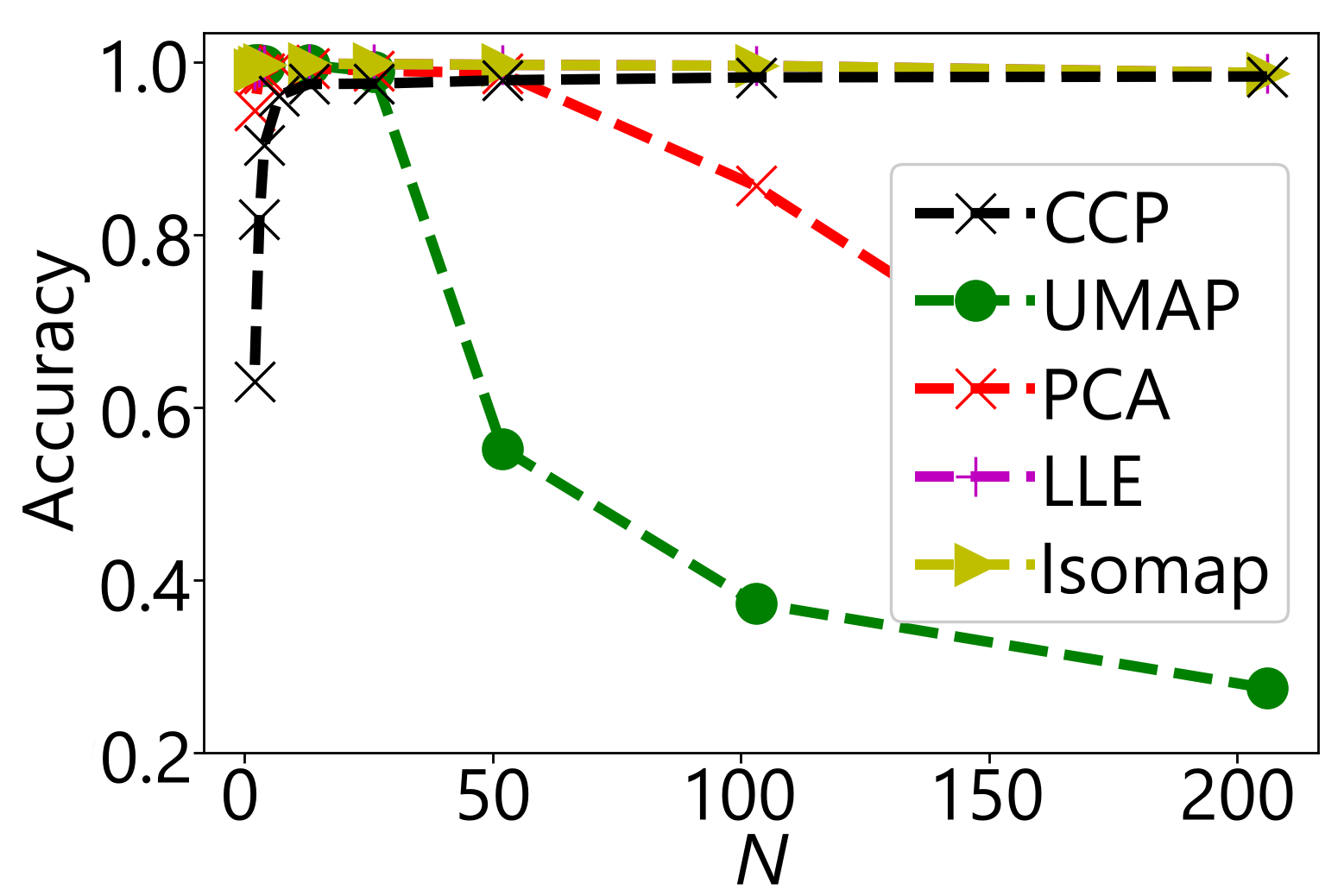}
	\caption{ Accuracy of $k$-NN classification of the TCGA-PANCAN dataset when the dimension is reduced to $N$ by using CCP, UMAP, PCA, LLE, and Isomap. Here, 5-fold cross-validation with 10 random seedings was used, and the test-train split was done prior to the reduction. For CCP, Lorentz kernel with $\kappa = 1$ and $\tau = 1.0$ was used. The sample size, feature size, and the number of classes of the TCGA-PANCAN are 801, 20531, and 5, respectively.}
	\label{fig: tcgapancan acc}
\end{figure}

 For CCP, the dimension of the TCGA-PANCAN dataset was reduced using Lorentz kernel with $\kappa = 1$ and $\tau = 1.0$. \autoref{fig: tcgapancan acc} shows the performance comparison of  CCP, UMAP, PCA, Isomap, and LLE. Here, a 5-fold cross-validation with 10 random seeds was used. Notice that CCP is comparable to Isomap and LLE in accuracy, whereas UMAP and PCA are unstable at higher dimensions.

\begin{figure}[H]
	\centering
	\includegraphics[scale = 0.45]{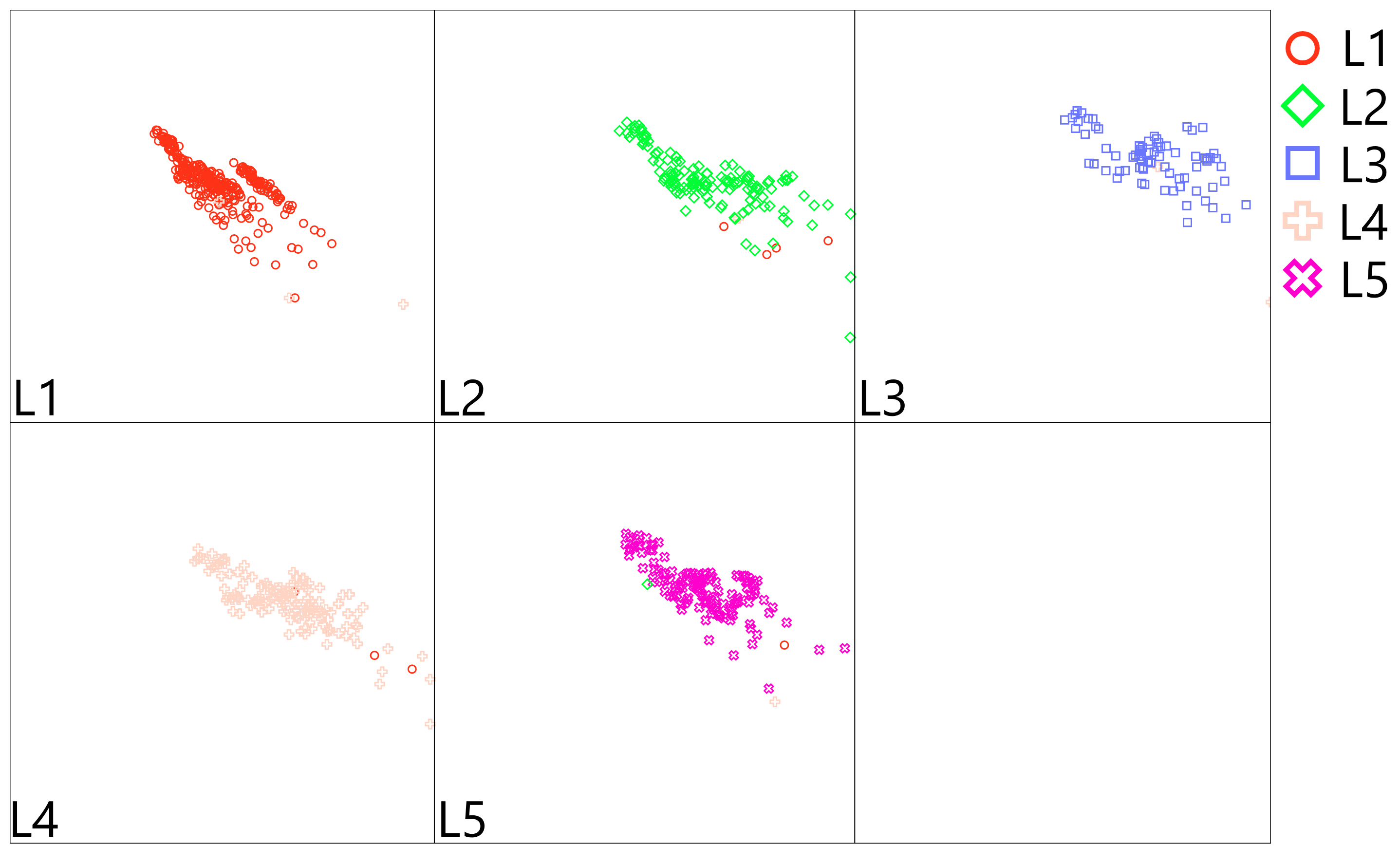}
	\caption{Visualization of the TCGA-PANCAN dataset when the dimension is reduced  to $N = 103$ by using CCP with Lorentz kernel, $\kappa = 1$ and $\tau = 1.0$. Each section represents a different class. The samples were plotted based on 103 features and colored with their predicted labels from $k$-NN classification via 5-fold cross-validation. The $x$ and $y$ axes are the residue and similarity scores, respectively. The sample size, feature size, and the number of classes of the TCGA-PANCAN are 801, 20531, and 5, respectively.}
	\label{fig: tcgapancan CCP}
\end{figure}

\autoref{fig: tcgapancan CCP} shows the R-S plot of the TCGA-PANCAN dataset when the dimension was reduced by CCP to $N = 103$. Each section corresponds to the 5 classes of TCGA-PANCAN. Samples were plotted according to 103 features and colored with the predicted labels from $k$-NN. The $x$ and the $y$-axes are the residue and similarity scores, respectively.

\begin{figure}[H]
	\centering
	\includegraphics[scale = 0.45]{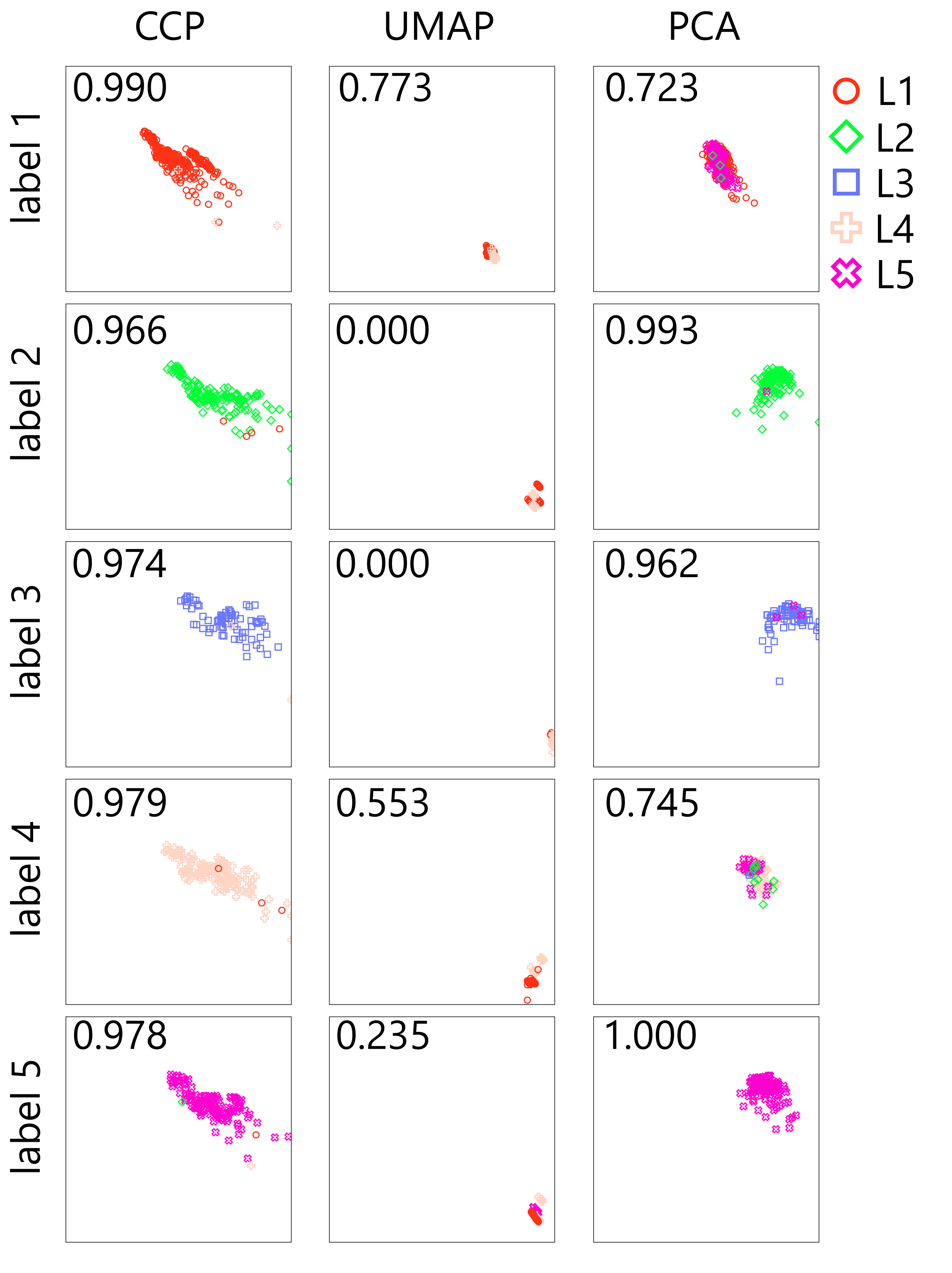}
	\caption{ Visualization of TCGA-PANCAN dataset when the dimension is reduced to $N=103$ by using CCP, UMAP, and PCA. For CCP, Lorentz kernel with $\kappa = 1$ and $\tau = 1.0$ was used. The $x$ and $y$ axes are the residue and similarity scores, respectively. Each row contains a  class. The data is plotted based on 103 features and colored with the predicted labels of the $k$-NN classifier, using 5-fold cross-validation. The sample size, feature size, and the number of classes of the TCGA-PANCAN are 801, 20531, and 5, respectively.}
	\label{fig: tcgapancan visual}
\end{figure}

\autoref{fig: tcgapancan visual} shows the R-S plot of TCGA-PANCAN when the dimension is reduced to $N = 103$ by using CCP, UMAP, and PCA, respectively.   The samples were plotted based on 103 features and colored with their predicted labels from $k$-NN. The $x$-axis and $y$-axis of each plot are the residue and similarity scores, respectively. Each row corresponds to one of the 5 classes, and the number inside the plot is the accuracy for each class. Notice that UMAP has a cluster in each plot, but the cluster has a low similarity score. This means that in UMAP's embedding, the sample within each class is not near each other, which results in low accuracy. PCA has comparable accuracy to CCP, but CCP has a notable improvement for class 1 and class 4.

\subsubsection{Coil-20}
\begin{figure}[H]
	\centering
	\includegraphics[scale = 0.45]{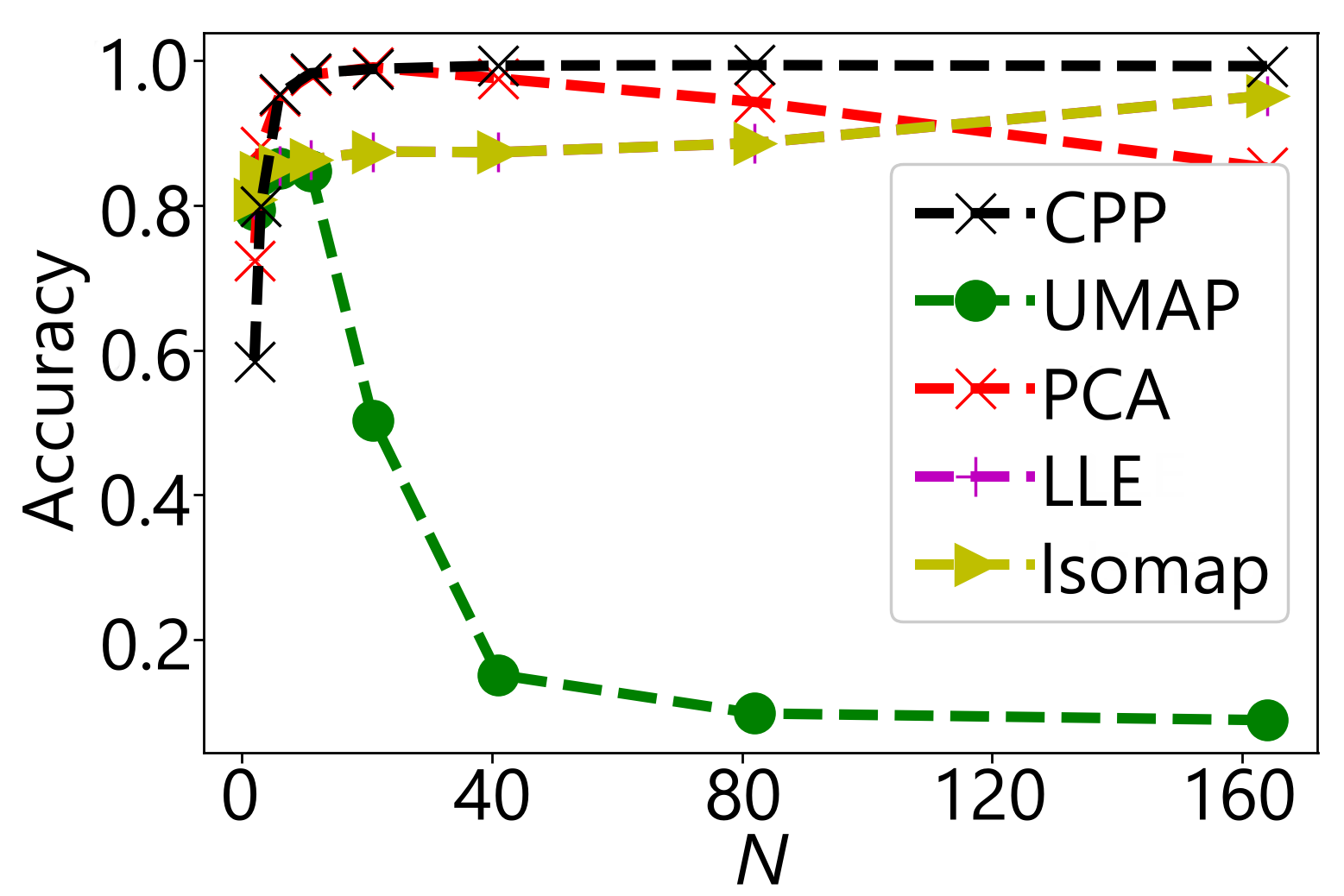}
	\caption{Accuracy of $k$-NN classification of Coil-20 dataset when its dimension is reduced to different dimensions $N$ by using  CCP, UMAP, PCA, LLE, and Isomap. The 10-fold cross-validation with 10 random seedings was used, and the test-train split was done prior to the dimensionality reduction. For CCP, Lorentz kernel with $\kappa = 1$ and $\tau = 6.0$ was used. The sample size, feature size, and the number of classes of the Coil-20 are 1440, 16384, and 20, respectively.}
	\label{fig: coil20 acc}
\end{figure}

The dimension of the Coil-20 dataset was reduced using Lorentz kernel with $\kappa = 1$ and $\tau = 6.0$. \autoref{fig: coil20 acc} shows the performance of CCP, UMAP, PCA, Isomap, and LLE. The 10-fold cross-validation with 10 random seeds was used. CCP has the best performance out of the 5  algorithms and maintains its accuracy in higher dimensions. PCA also has high accuracy but loses its accuracy in higher dimensions.

\begin{figure}[H]
	\centering
	\includegraphics[scale = 0.45]{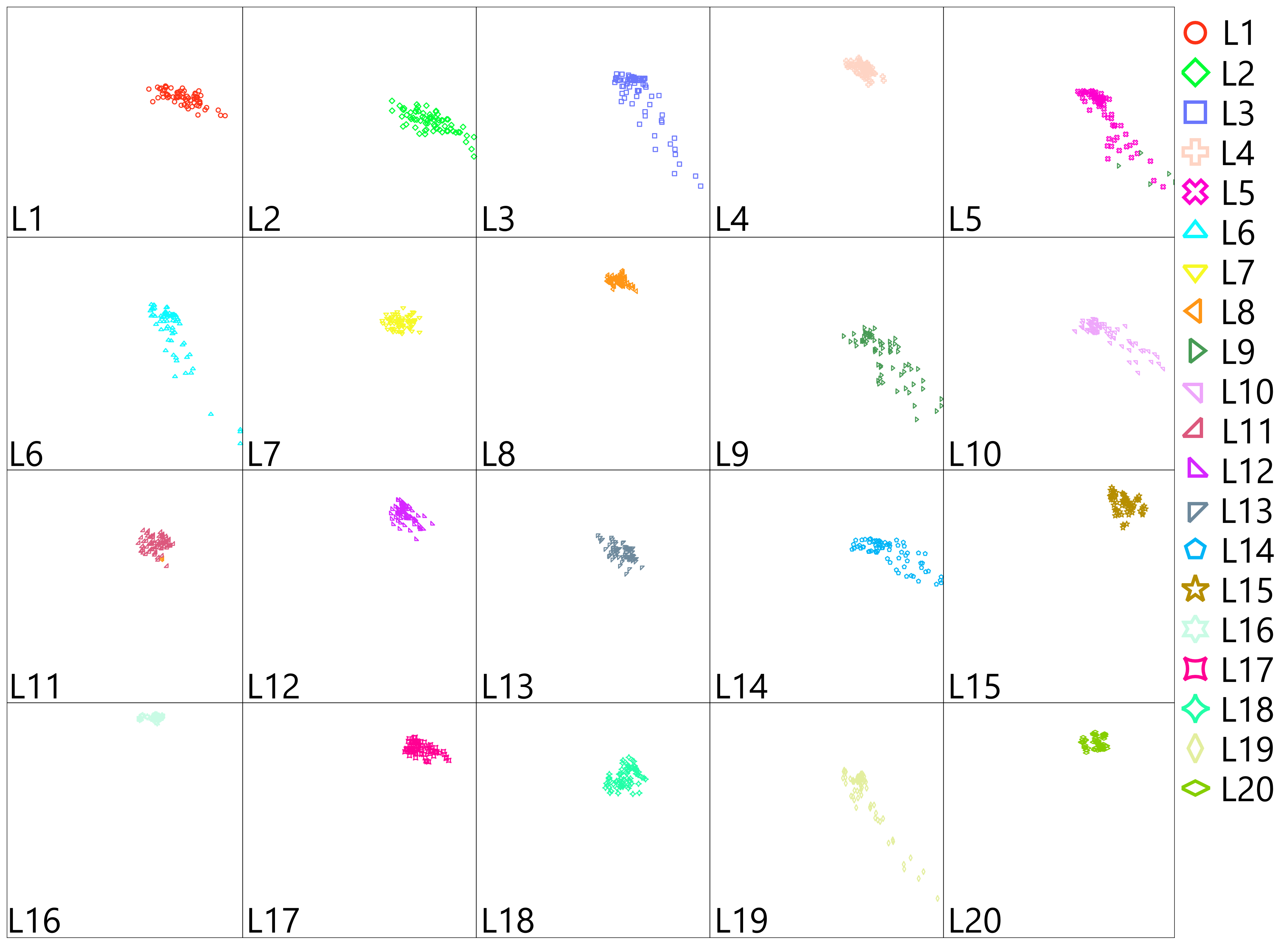}
	\caption{Visualization of Coil-20 dataset when the dimension is reduced  to $N = 82$ by using CCP with Lorentz kernel, $\kappa = 1$ and $\tau = 6.0$. Each section represents a different class, and the samples were plotted based on 82 features and colored with their predicted labels from the $k$-NN classification via 10-fold cross-validation. The $x$ and $y$ axis are the residue and similarity scores, respectively. The sample size, feature size, and the number of classes of the Coil-20 are 1440, 16384, and 20, respectively.}
	\label{fig: coil20 CCP}
\end{figure}

 \autoref{fig: coil20 CCP} shows the R-S plot of the Coil-20 dataset when the dimension is reduced by CCP to $N = 82$. Each section corresponds to the 20 classes of Coil-20. Samples were plotted based on 82 features and colored with the predicted labels from $k$-NN. The $x$ and the $y$-axes are the residue and similarity scores, respectively.

\begin{figure}[H]
	\centering
	\includegraphics[scale = 0.4]{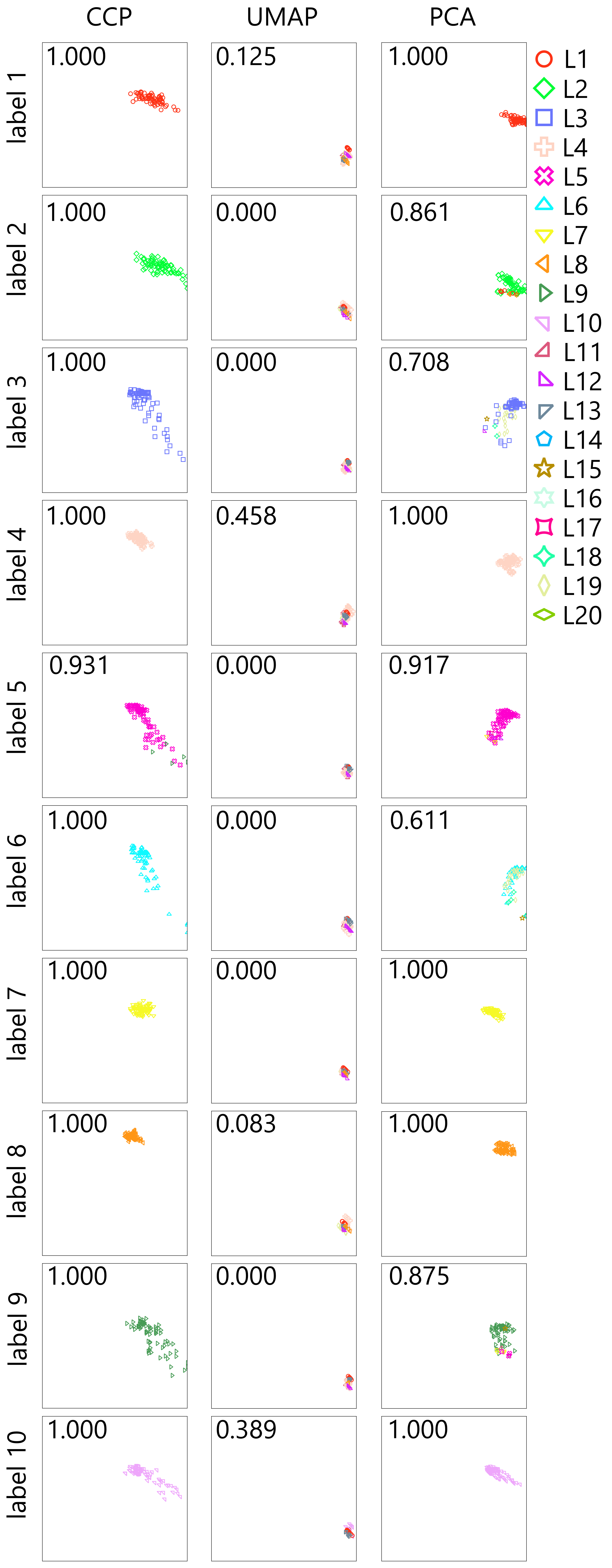}
	\caption{Visualization of Coil-20 dataset for classes 1 to 10 when the dimension is reduced to $N=82$ by using CCP, UMAP, and PCA. For CCP, Lorentz kernel with $\kappa = 1$ and $\tau = 6.0$ was used. The $x$ and $y$ axis are the residue and similarity scores, respectively. Each row visualizes each class. The data is plotted based on 82 features and colored based on the predicted labels of the $k$-NN classifier, using 10-fold cross-validation. The sample size, feature size, and the number of classes of the Coil-20 are 1440, 16384, and 20, respectively.}
	\label{fig: coil20 visual1}
\end{figure}

\begin{figure}
	\centering
	\includegraphics[scale = 0.4]{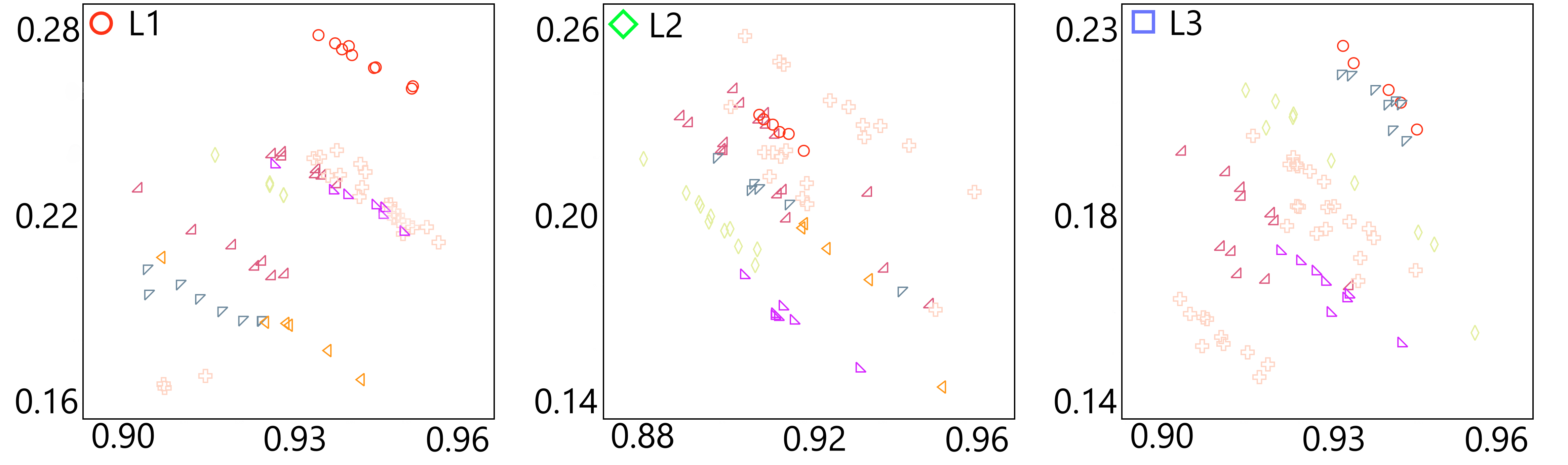}
	\caption{Visualization of Coil-20 dataset class 1, 2, and 3, when the data dimension is reduced to $N=82$ by UMAP. The figures are zoomed-in view. The data were plotted based on 82 features and colored according to their predicted labels from the $k$-NN classifier using 10-fold cross-validation. Label 1 has an accuracy of 0.125, whereas labels 2 and 3 have accuracy 0.000.}
	\label{fig: coil20 umap}
\end{figure}

\autoref{fig: coil20 visual1} shows the R-S plot of Coil-20 when its dimension is reduced to $N = 82$ by using CCP, UMAP, and PCA. Samples in each class are plotted according to their 82 features and colored according to their predicted labels from $k$-NN. The $x$-axis and $y$-axes of each plot are the residue and similarity scores, respectively. Each row corresponds to one of the 20 classes, and the number inside the plot is the classification accuracy for each class. Notice that all of UMAP's visualizations show a poor distribution in the bottom right, indicating that the residual score is high and the similarity score is low, which gives rise to poor performance in the classification. In order to further investigate the performance, labels 1, 2, and 3 were visualized in \autoref{fig: coil20 umap}. We can see that in the zoomed-in view, there are small subclusters within each plot, which come from different folds of the cross-validation.

\begin{figure}[H]
	\centering
	\includegraphics[scale = 0.4]{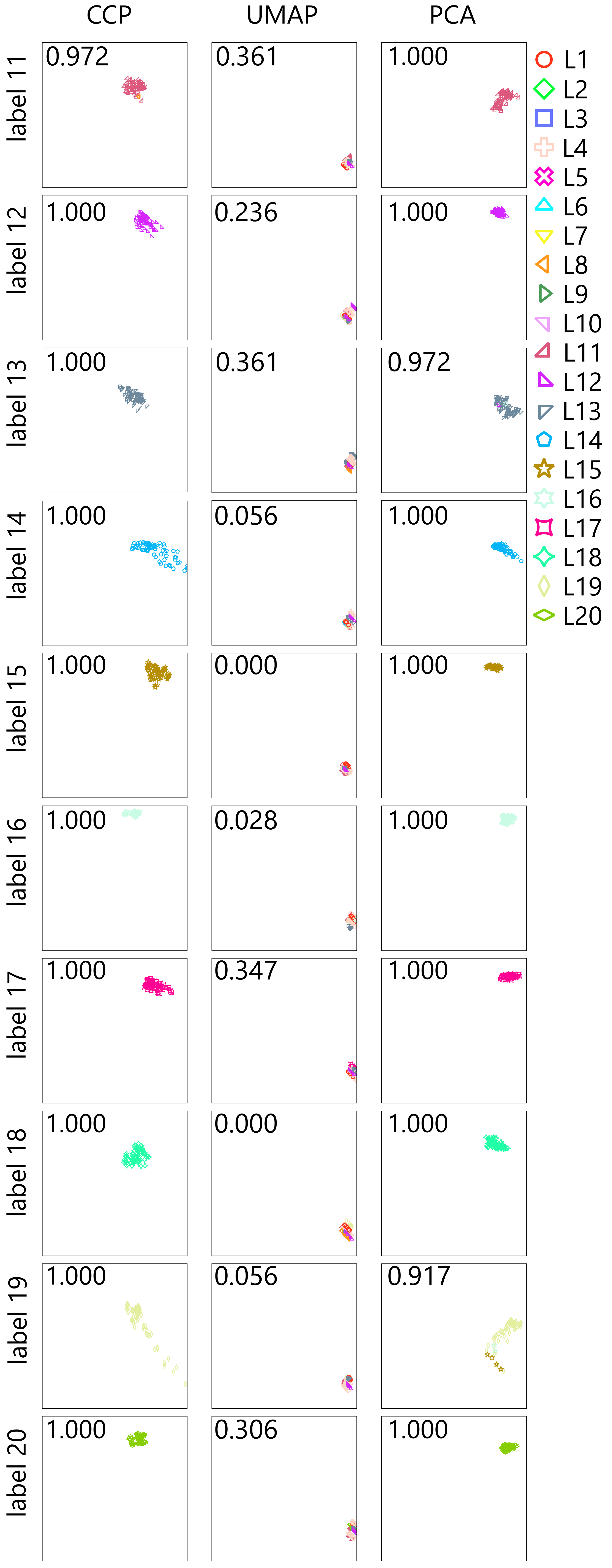}
	\caption{Visualization of Coil-20 dataset for classes 11 to 20 when the dimension is reduced to $N=82$ by using CCP, UMAP, and PCA. For CCP, Lorentz kernel with $\kappa = 1$ and $\tau = 6.0$ was used. The $x$ and $y$ axis are the residue and similarity scores, respectively. Each row visualizes each class. The data is plotted based on 82 features and colored based on the predicted labels of the $k$-NN classifier, using 10-fold cross-validation. The sample size, feature size, and the number of classes of the Coil-20 are 1440, 16384, and 20, respectively.}
	\label{fig: coil20 visual2}
\end{figure}

\subsubsection{Coil-100}

\begin{figure}[H]
	\centering
	\includegraphics[scale = 0.45]{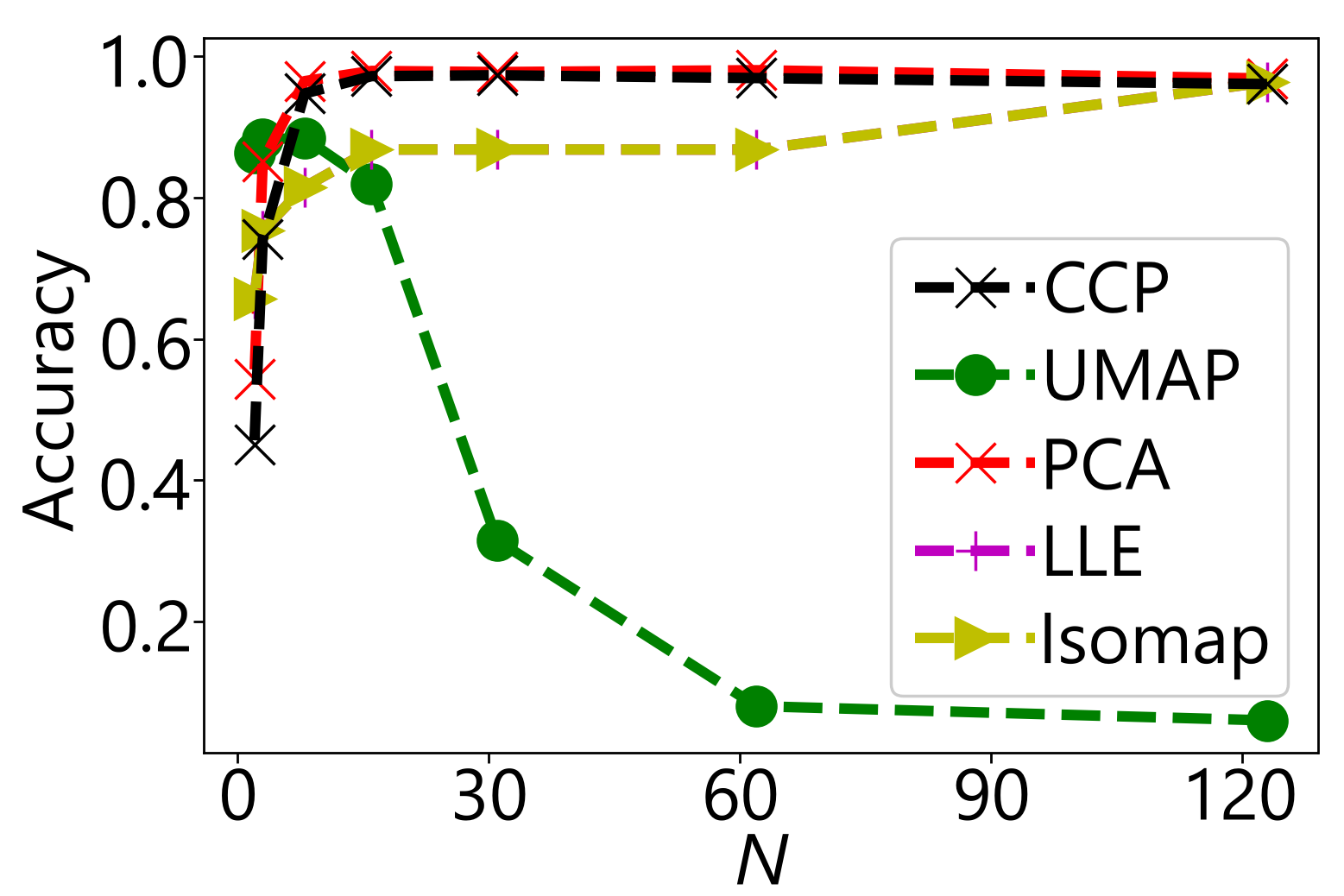}
	\caption{ Accuracy of $k$-NN classification of Coil-100 dataset when the dimension is reduced to $N$ by using CCP, UMAP, PCA, LLE, and Isomap. The 10-fold cross-validation with 10 random seedings was used, and a test-train split was done prior to the reduction. For CCP,Lorentz kernel with $\kappa = 1$ and $\tau = 6.0$ was used. The sample size, feature size, and the number of classes of the Coil-20 are 1440, 16384 and 20, respectively.}
	\label{fig: coil100 acc}
\end{figure}

The dimension of the Coil-100 dataset was reduced using exponential kernel with $\kappa = 1$ and $\tau = 6.0$. \autoref{fig: coil100 acc} shows the performance of CCP, UMAP, PCA, Isomap, and LLE. Here,  10-fold cross-validation with 10 random seeds was used. CCP, PCA, LLE, and Isomap have comparable results, whereas UMAP is unstable at a higher dimension $N$. The best performance of UMAP was not as good as those of CCP and PCA. This indicates that Coil-100 has a high intrinsic dimension, for which UMAP has poor performance.

\section{Discussion}

\subsection{Centrality based CCP}

CCP used FRI to project a group of correlated features into a 1D representation. If we observe the projection in a graph setting, the FRI projection can be viewed as computing the degree centrality of the graph. That is, let $Z \in \mathbb{R}^{M\times I}$ be the data, with $M$ samples and $I$ features. For each partition, we can define a graph $G^{n} = (V^{n}, E^{n}, W^{n})$, $n=1,2,...,N$, where $V^{n}$, $E^{n}$ and $W^{n}$ are the vertex, edge and weight sets of the graph of the $n$th component, respectively. The weights are precisely the kernels defined in Eq. (\ref{weights}). Then, the FRI projection for $\mathbf{x}^{n}_i$ can be viewed as the degree centrality ($C_d$) of a weighted graph
\begin{align}
	C_d(\mathbf{z}^{\mathcal{S}^n}_i) = \sum_{\mathbf{z}^{\mathcal{S}^n}_j} \Phi^{n}(|\mathbf{z}^{\mathcal{S}^n}_i - \mathbf{z}^{\mathcal{S}^n}_j|; \tau, \eta^n, \kappa),
\end{align}
where $C_d(\mathbf{z}^{S_n}_i)$ is the degree centrality of vertex $\mathbf{z}^{S_n}_i$. In this case, we treat each data $\mathbf{z}^{\mathcal{S}^n}_i$ as a vertex.

Instead of using the FRI projection, we can impose a traditional graph-based approach, setting the edge weight $\omega^{n}_{ij} =  1$ for all $1\le i,j\le M$ and $1 \le n \le N$, when the node-node distance satisfies a cutoff. That is, instead of applying Eq. (\ref{weights}), we take 
\begin{align}
	A^n = \{A^n_{ij}\}, \quad A^n_{ij} = \begin{cases}
		1, & \text{if $\|\mathbf{z}^{\mathcal{S}^n}_i-\mathbf{z}^{\mathcal{S}^n}_j\| < r^n_c$ } \\
		0, & \text{otherwise}
	\end{cases}, \quad 1 \le i,j \le M.
\end{align}
Here, instead of writing $C^n_{ij}$ as in Eq. (\ref{weights}), we use $A^n_{ij}$ to denote the adjacency matrix of the graph, and $r^n_c$ is the cutoff distance. Then, the reduced new variables $x^n_i$ can be computed by replacing  $\Phi^{n}(\|\mathbf{z}_i^{\mathcal{S}^n} - \mathbf{z}_m^{\mathcal{S}^n}\|; \tau,  \eta^n, \kappa)$ in Eq. (\ref{xcomponent}) with $A^n_{im}$.
In such a manner, we can implement other centrality formulations, such as the degree centrality, closeness centrality \cite{freeman1978centrality}, betweenness centrality \cite{freeman1977set}, and eigenvector centrality \cite{bonacich1987power} in CCP.

\begin{figure}[H]
	\centering
	\includegraphics[scale = 0.45]{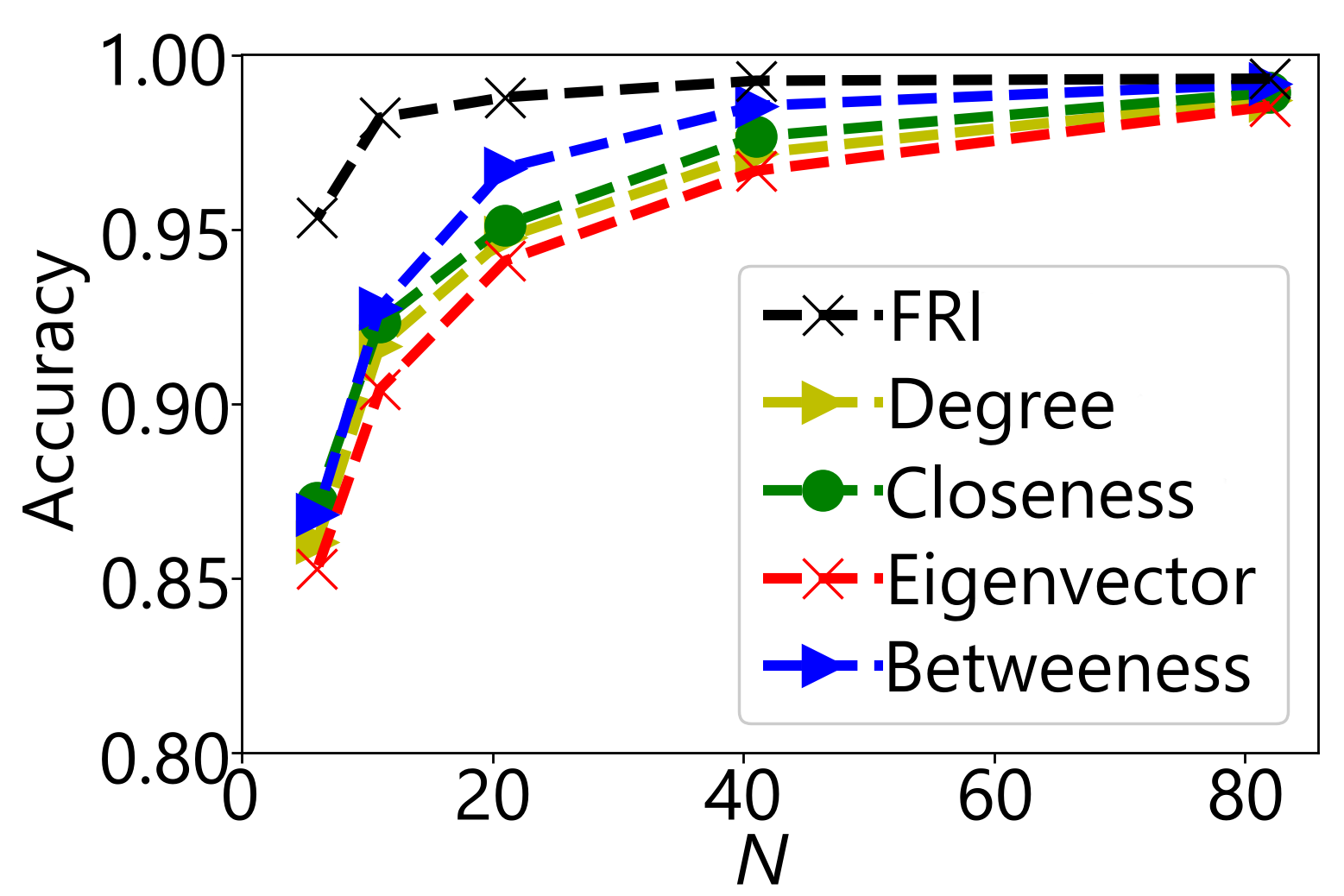}
	\caption{The coil-20 dataset was reduced using centrality formulations, instead of the FRI projection. Degree, closeness, eigenvector centrality, and betweenness centrality were tested, with $r_c = 0.7 d_{\max}$. The accuracy was calculated from 10-fold cross-validation with 10 random seeds. The sample size, feature size, and the number of classes of the Coil-20 are 1440, 16384, and 20, respectively.}
	\label{fig: coil20 centrality}
\end{figure}

\autoref{fig: coil20 centrality} shows the accuracy of using different centrality formulations instead of the FRI projection. Using the adjacency matrix, degree centrality, closeness centrality, betweenness centrality, and eigenvector centrality were computed with $r_c = 0.7d_{\max}$, where $d_{\max}$ is the maximum pairwise distance between the input data. The performance of all methods is quite similar. However, the stability of computing the centrality is heavily reliant on $r_c$. Moreover, if the data is well clustered within each class, the graph may not be connected, which may affect the stability of centrality computations.

\subsection{Correlation distance based CCP}

\begin{figure}[H]
	\centering
 \includegraphics[scale = 0.45]{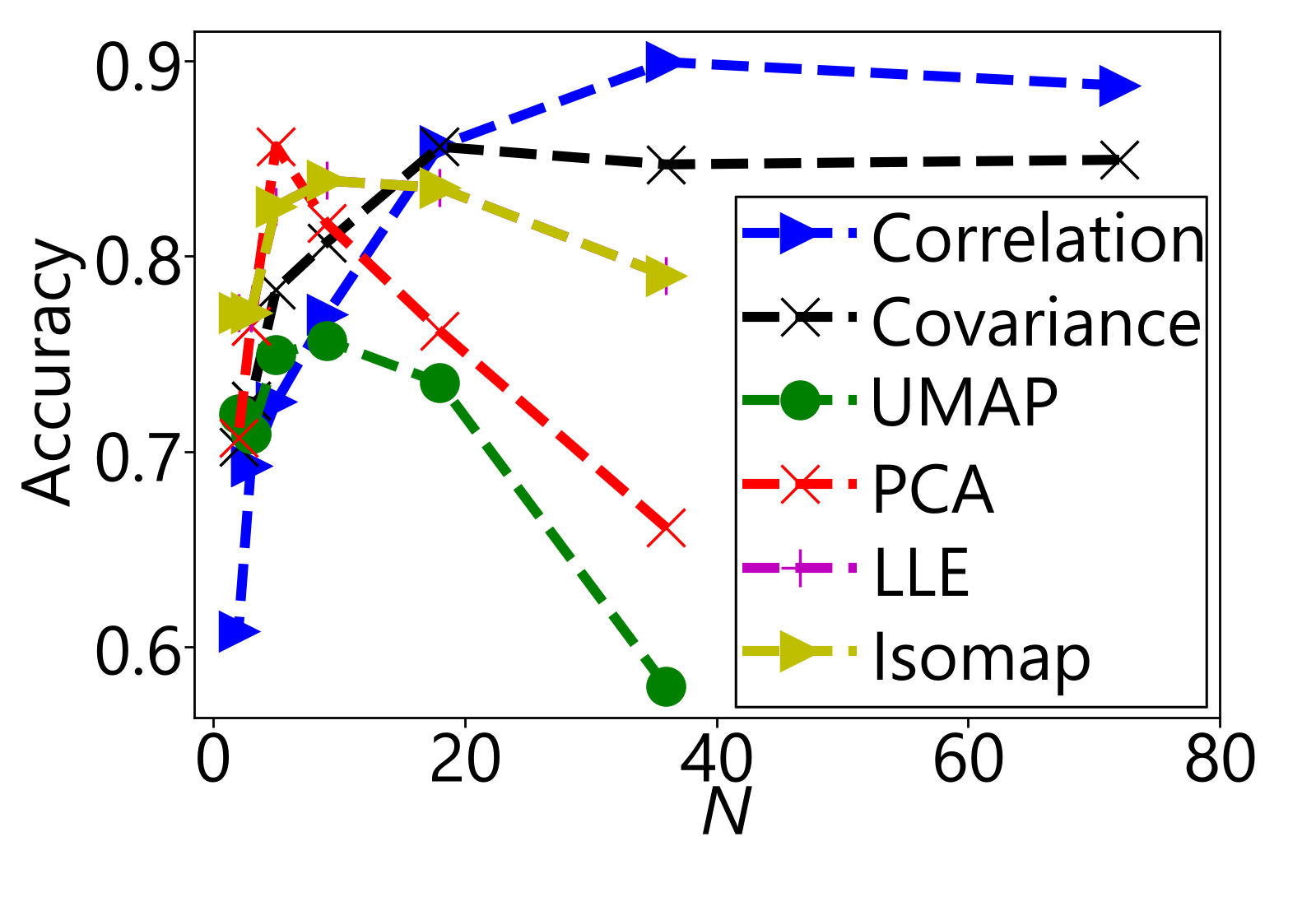}
	\caption{Comparison between correlation distance-based partitioning and covariance distance-based partitioning of ALL-AML dataset. $k$-NN with 10-fold cross-validation was used to compute the accuracy.  The sample size, feature size, and the number of classes of ALL-AML are 72, 7129, and 2, respectively.}
	\label{fig: ALLAML cor}
\end{figure}

CCP utilizes covariance distance in clustering to partition features. However, other distance metrics can be used in clustering as well, depending on the size of the data and the relationship between the features. In particular, correlation distance can be used instead of covariance distance, when the relationship between features is highly nonlinear. \autoref{fig: ALLAML cor} shows the effectiveness of correlation distance-based CCP when compared to covariance distance-based CCP and other DR algorithms. Notice that the correlation distance-based CCP significantly outperforms covariance-based CCP and other DR algorithms.  Therefore, correlation distance-based CCP can be employed if high accuracy is desirable. However, it is noted that correlation distance-based CCP is very time-consuming and memory-demanding. This limitation may constrain the use of correlation distance-based CCP in high-dimensional data with large data sizes.

\subsection{Parameter-free CCP}

The performance of the proposed two-step CCP depends on a few parameters, such as the dimension $N$, kernel type (i.e.,  generalized exponential and generalized Lorentz), power ($\kappa$), and scale ($\tau$). Among them, the dimension may be chosen by the user.  Although a set of default parameters is prescribed, it may not be optimal for different datasets. It will be a burden for users to select parameters. Fortunately, CCP is very stable under subsampling. Therefore, we can use subsampling to search the optimal parameter range for a given dataset automatically.  

\begin{figure}[H]
	\centering
	\includegraphics[width = 0.8\textwidth]{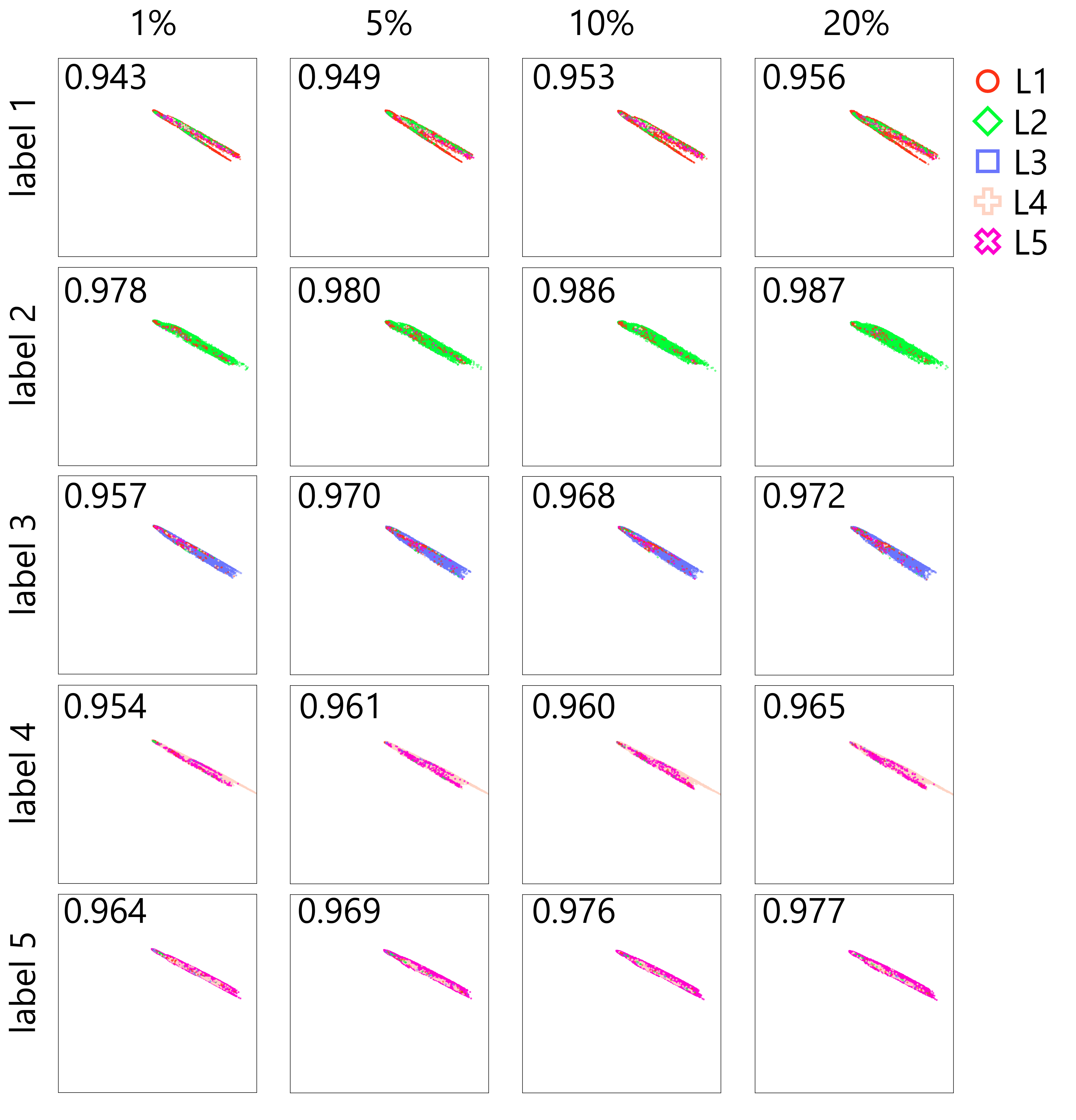}
	\caption{R-S plot visualization of Smallnorb classification using CCP with the reduction ratio of 400 (47 dimensions) at $1\%, 5\%, 10\%$ and $20\%$ subsampling. Each row represents the data plotted based on 47 features and colored with the predicted labels from the $k$-NN classifier, using 10-fold cross-validation. The number in each plot shows the accuracy within each label obtained with subsampling-generated kernel parameters. 
	The $x$ and $y$ axis are the residue score and the similarity scores, respectively.}
	\label{fig: smallnorb visual}
\end{figure}

In this subsection, we show that CCP is stable under subsampling. To verify this claim, we test CCP on the Smallnorb dataset, which has 46,800 samples and 5 classes. Each sample consists of a binocular picture of an object of size 96$\times$96 pixels, taken from different radial and azimuthal angles. We flattened each image and combined the images to make an 18,432-dimensional feature vector. We subsample $1\%, 5\%, 10\%$ and $20\%$ samples to optimize CCP kernel parameters, respectively. Then, based on these CCP parameter sets, we carry out the CCP dimensionality reduction of the whole dataset for classification. The resulting 10-fold cross-validation accuracies of classification for the Smallnorb dataset are shown in  \autoref{fig: smallnorb visual} for subsampled at $1\%, 5\%, 10\%$ and $20\%$. It is clear that the accuracy increases as the subsampling size is increased from 1\% to 20\%. However, the accuracy difference between 1\% subsampling and 20\% subsampling is under 2\% for all classes.  It is seen that under different subsampling ratios, CCP can capture the structure of the data. Even at $1\%$ subsampling, CCP is still very accurate.

Since CCP is very stable under subsampling, one can make CCP a parameter-free method by using a relatively small amount of dataset to determine CCP parameters automatically.  

CCP's stability under subsampling implies that CCP can be used in the dynamic data acquisition of excessively large datasets. Newly collected data can be added to the existing data without the need to restart the CCP calculation from the very beginning.

\subsection{R-S plot vs 2D plot}

\begin{figure}[H]
	\centering
	\begin{subfigure}{\textwidth}
		\centering
		\includegraphics[scale = 0.4]{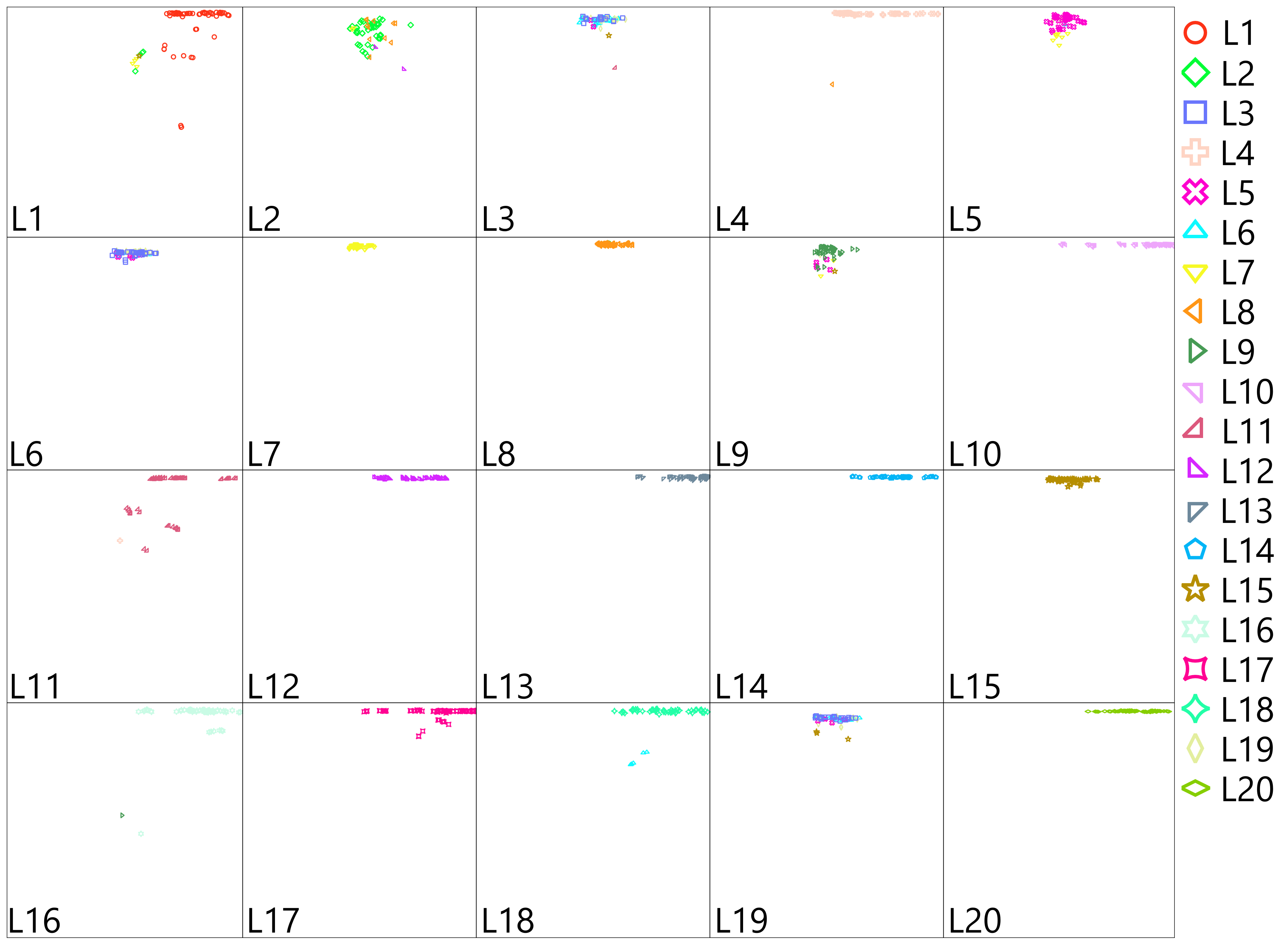}
		\caption{R-S plot of Coil-20 with 2 dimension.}
	\end{subfigure}
	\begin{subfigure}{\textwidth}
		\centering
		\includegraphics[scale = 0.4]{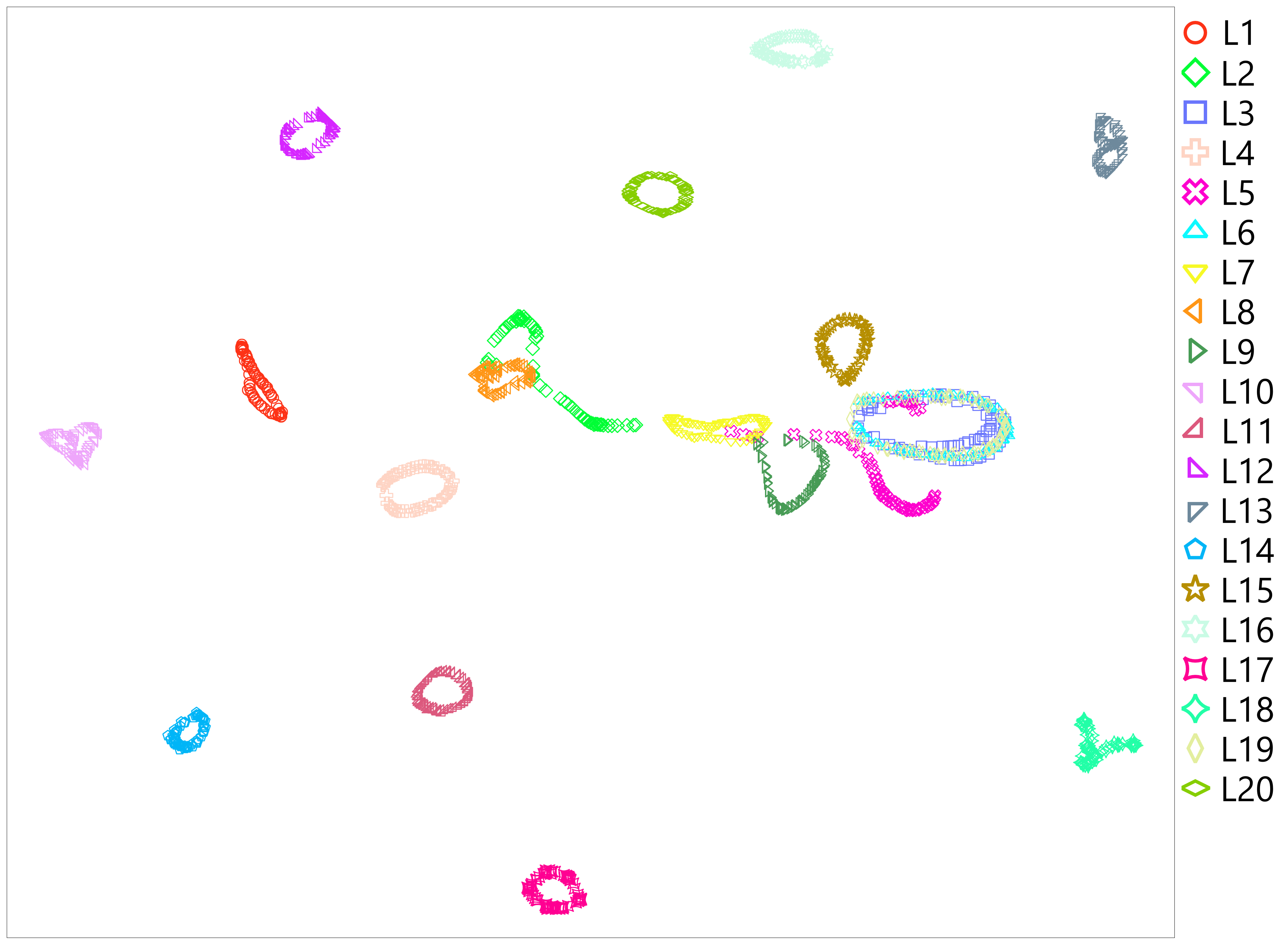}
		\caption{Traditional 2d plot of Coil-20 using UMAP.}
	\end{subfigure}
	\caption{ (a) shows the R-S plot of the Coil-20 dataset when reduced by UMAP to $N=2$. Each section represents a different class, where the data points were colored according to their predicted labels from $k$-NN classification via 10-fold cross-validation. (b) Coil-20 dataset reduced to $N = 2$ by UMAP. The data points were colored according to the ground truth.}
	\label{fig: coil20 rs umap}
\end{figure}

R-S plot is an effective tool to visualize the performance of classification in general. \autoref{fig: coil20 rs umap} shows the comparison of the R-S plot and the traditional 2D visualization of the Coil-20 dataset when the dimension is reduced to 2 by UMAP. For the traditional 2D plot, each data point was colored by the ground truth, and for the R-S plot, each section represents one of the 20 different classes, and data points were colored by the predicted labels from the $k$-NN classification. We can see that in the traditional 2D visualization, labels 3, 9, and 19 are located in the same region. It is interesting to see that this situation is reflected in our R-S plot as three labels mixed up. In the R-S plot, Labels 3, 9, and 19 have a high similarity score but a low residue score, meaning that the data points are not separated well among different classes and show the limitation of preserving the local structure of a high dimensional data represented in the 2D space. Essentially, some data lay in an intrinsically high-dimensional space that cannot be well-described in the 2D representation. 

Note that 2D plots work best when the data dimension is reduced to 2, whereas the R-S plots can be applied to arbitrarily high dimensions.  

\subsection{R-S indices}

We have shown in the previous section that R-S plots can be used as an alternative method of visualizing the data. In this section, we  illustrate the utility of the residue index, similarity index, and R-S index.

\begin{figure}[H]
	\centering
	\makebox[\textwidth][c]
	{\includegraphics[width = 0.9\textwidth]{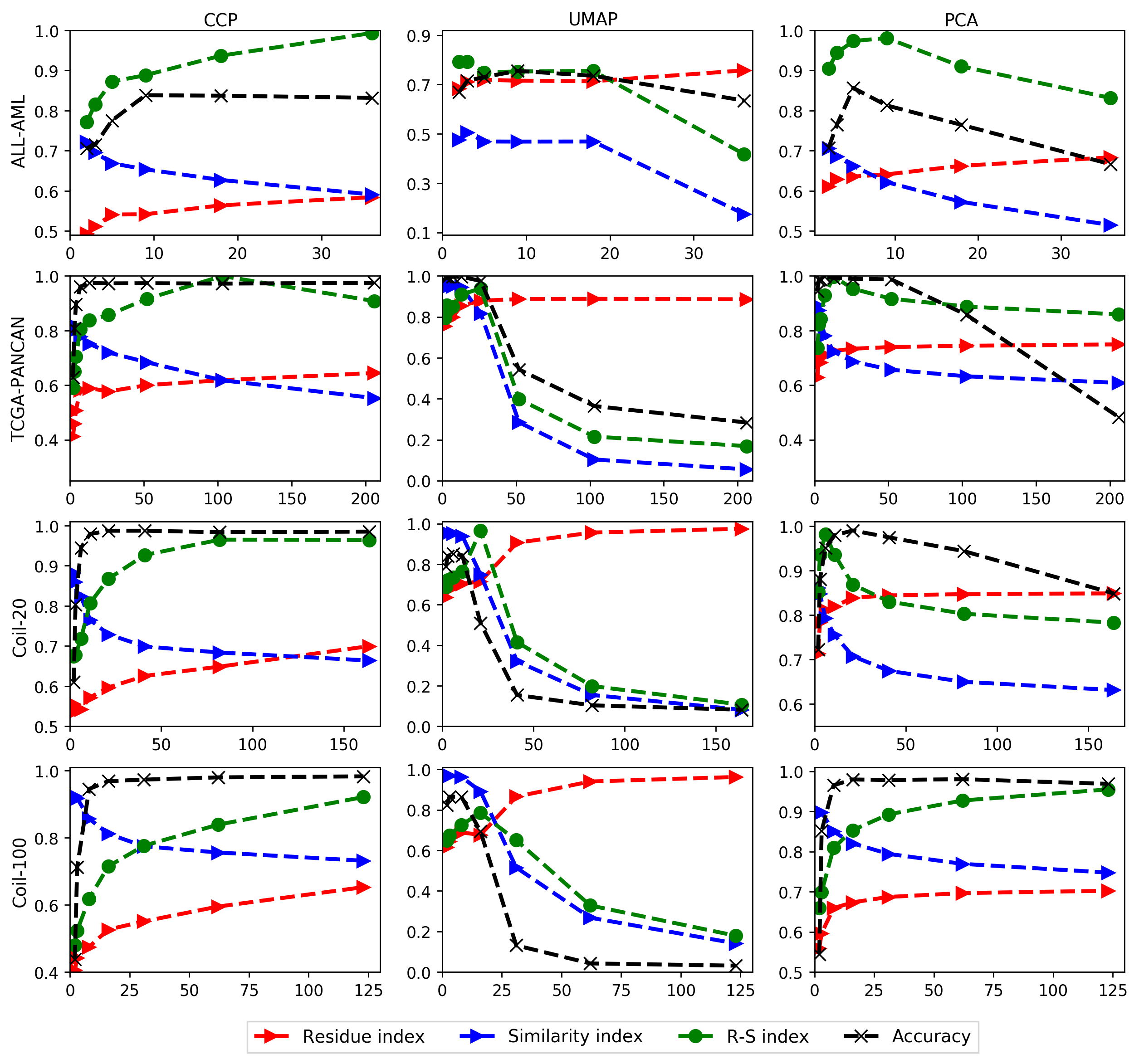} }
	\caption{ Illustration of residue index, similarity index, R-S index, and accuracy from $k$-NN classification. Residue index, similarity index, R-S index were calculated for each seed individually, and the average values were taken over 10 random seeds. Accuracy was obtained from taking the average of cross-validation with 10 random seeds. The rows correspond to the 4 datasets, ALL-AML, TCGA-PANCAN, Coil-20, and Coil-100 from top to bottom, and the columns correspond to CCP, UMAP, and PCA from right to left. The $x$ axis is the reduced dimension $N$, and the $y$ axis is the accuracy and/or indices. } 
	\label{fig: acc rs comparison}
\end{figure}

\autoref{fig: acc rs comparison} demonstrates the residue index, similarity index, R-S index, and tradition accuracy. The red and the blue lines are the residue and similarity indices, and the green line is the R-S index. The black line shows the accuracy from the $k$-NN classification. Rows correspond  to different datasets and columns are associated with CCP, UMAP, and PCA.

First, we noticed that the residue index and similarity index have opposite trends for all datasets and all methods as the reduced dimension $N$ increases. Most notably, for all methods over all datasets, there is a strong correlation between the R-S index and accuracy, indicating its utility in performance evaluation.

\subsection{Accuracy comparison using four classifiers}

We have shown the effectiveness of CCP on various datasets. However, all the aforementioned analysis was based on the $k$-NN classifier.  It is important to know whether the same pattern returns if other classification algorithms are employed. 
To this end, we compare CCP with other dimensionality reduction methods using $k$-NN, support vector machine (SVM), random forest (RF), and gradient boost decision tree (GBDT).

\begin{figure}[H]
	\makebox[\textwidth][c]
	{\includegraphics[width = 1.0\textwidth]{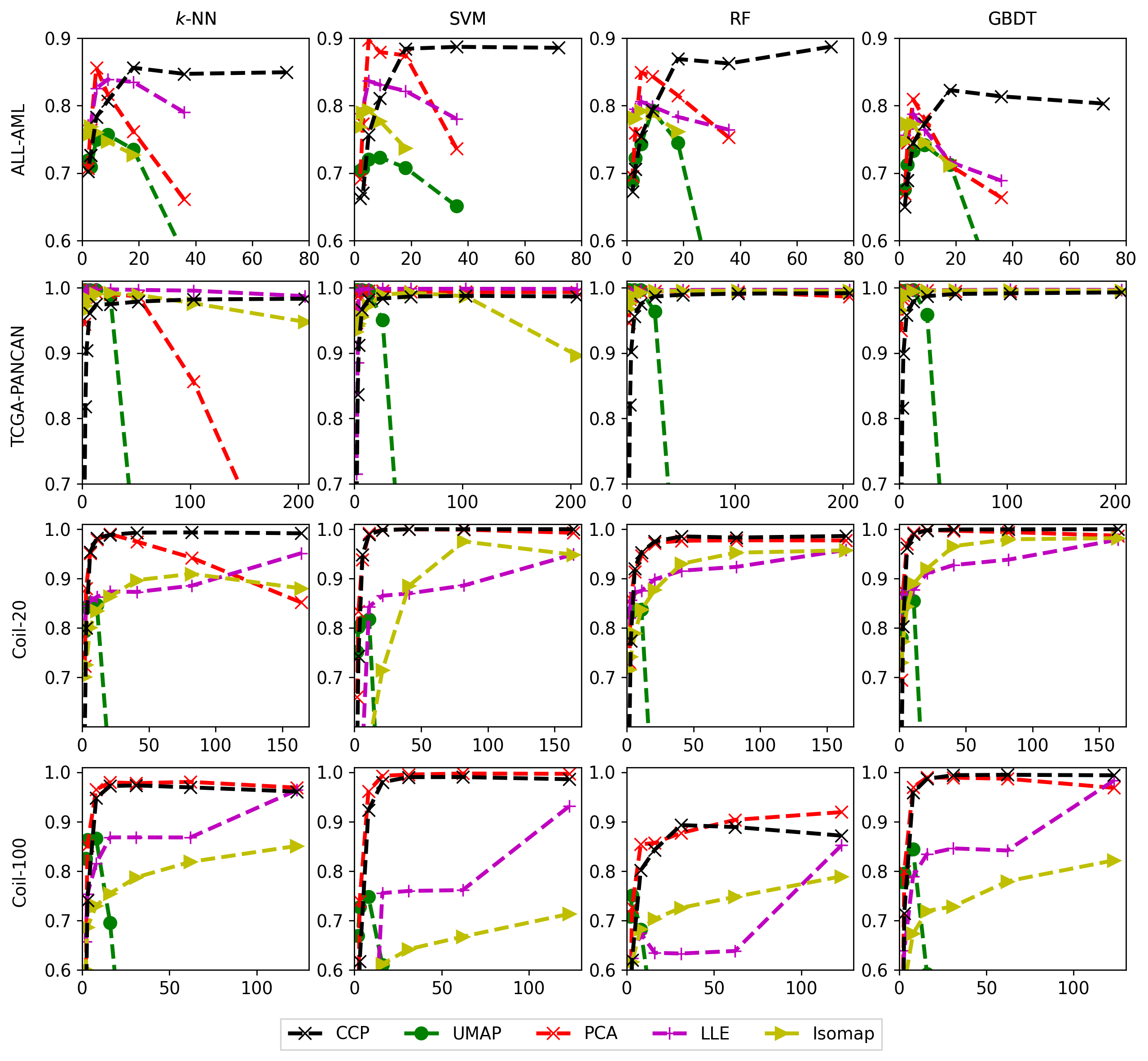}}
	\caption{Comparison of the accuracy of CCP on a variety of datasets and classification algorithms. The rows represent four datasets,  from the top to the bottom: ALL-AML, TCGA-PANCAN, Coil-20, and Coil-100. The columns are for four classification algorithms, namely,  $k$-NN, SVM, RF, and GBDT. The $x$-axes are  the reduced dimension $N$ and the $y$-axes are accuracy. }	
	\label{fig: ccp method comparison}
\end{figure}

\autoref{fig: ccp method comparison} shows the comparison of CCP when utilizing $k$-NN, SVM, RF, and GBDT on ALL-AML, TCGA-PANCAN, Coil-20, and Coil-100 datasets. The rows are the 4 datasets, and the columns are 4 classification methods. For all the tests, sklearn's classification package was utilized. For $k$-NN and SVC, default parameters were used. For RF and GBDT, \{n\_estimators=1000, max\_depth = 7, min\_samples\_split = 3, max\_features = 'sqrt', n\_jobs = -1 \} were used. For all tests, standard scaling was used after the reduction.

 First, CCP remains very competitive against all other dimensionality reduction methods over all datasets when other classifiers are employed. The relative behaviors of all dimensionality reduction methods did not change much under different classifiers. Therefore, our earlier comparison is fair and our findings remain correct.  

Second, SVM appears to slightly improve the performance of CCP and PCA. However, LLE and Isomap  do not work well with SVM.
 
Third,  UMAP did not perform well on ALL-AML, Coil-20, and Coil-100 when the $k$-NN method was used. However, its performance does not improve much with SVM, RF, and GBDT. Its instability with relatively large reduced dimension $N$ persists over different classifiers. In fact, its best results have never reached those of other methods for these three datasets. A possible reason is that UMAP does not work well for data having moderately large intrinsic dimensions. 

Fourth, LLE had some instability in TCGA-PANCAN and Coil-100 datasets. Because the input data led the computed matrix to become singular, some of the tests from the cross-validation were not computed. For these cases, the average was taken over the working tests.

Finally, we noticed that all dimensionality reduction methods underperformed with RF for the Coil-100 dataset and with GBDT for the ALL-AML dataset. This behavior might be due to the fact that for a given classifier, a uniform set of parameters was used for all datasets and RF does not work well for large datasets.

\subsection{Efficiency comparison}

Although accuracy is very important, computational cost can be a crucial factor for huge datasets. In this section, we assess the computational times of various methods with elementary computer resource allocations. Specifically, 4 central processing units (CPUs) with 64GB of memory from the High-Performance Computing Center (HPCC) of Michigan State University were used for all methods and all datasets.

\begin{figure}[H]
	\centering
	\makebox[\textwidth][c]
	{\includegraphics[width = 1.0\textwidth]{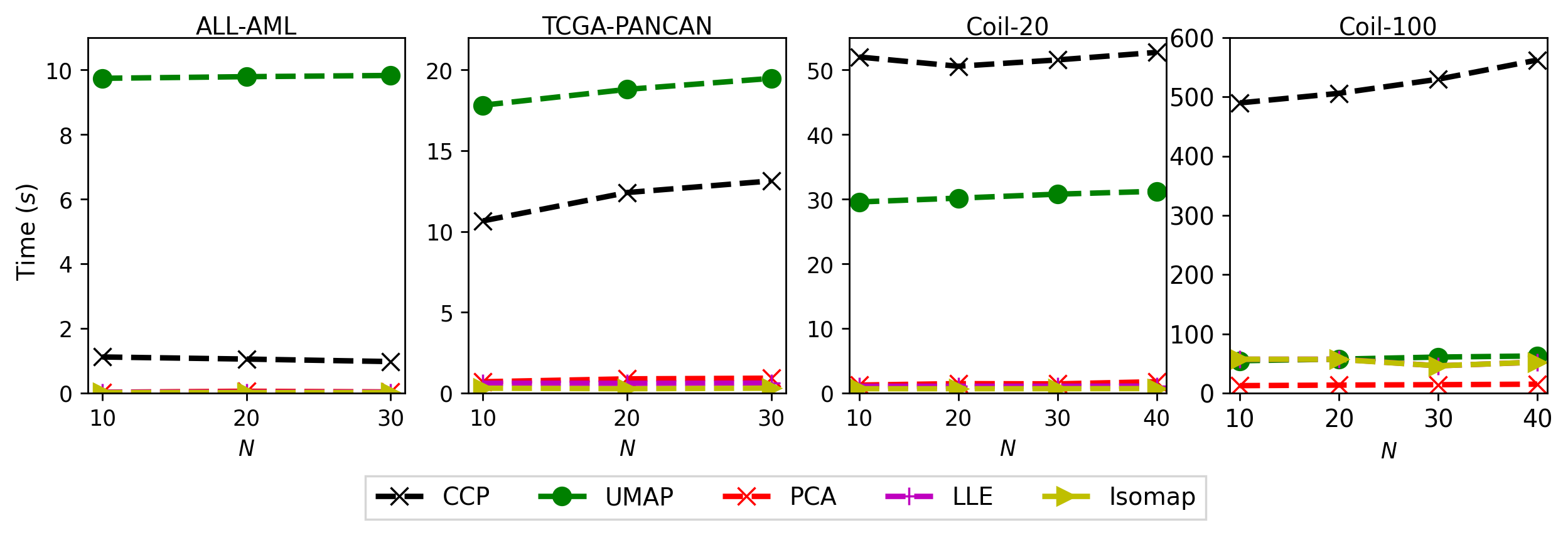} }
	\caption{CPU run time comparison among CCP, UMAP, and PCA on  ALL-AML, TCGA-PANCAN, Coil-20, and Coil-100 datasets. For ALL-AML and TCGA-PANCAN, computational times for $N=10, 20,$ and $30$ were calculated by taking the average of the 5-fold cross-validation over 10 random seeds. For Coil-20 and Coil-100 computational times for $N=10, 20, 30,$ and $40$ were calculated by taking the average of the 10-fold cross-validation over 10 random seeds. In each chart, the $x$ axis corresponds to the reduced dimension $N$, and the $y$ axis is the average time ($s$).} 
	\label{fig: time comparison}
\end{figure}

\autoref{fig: time comparison} shows the computational time of the three-dimensionality reduction methods on ALL-AML, TCGA-PANCAN, Coil-20, and Coil-100. For ALL-AML and TCGA-PANCAN datasets, the average time from the 5-fold cross-validation over 10 random seeds was computed. For Coil-20 and Coil-100, the average time from the 10-fold cross-validation over 10 random seeds was recorded.

PCA shows essentially the fastest computation for all datasets.  Isomap and LLE have very similar behaviors for all datasets. Their time efficiencies are quite similar to that of PCA.   

UMAP is faster than CCP for Coil-20 and Coil-100. For ALL-AML and TCGA-PANCAN, CCP is faster because of the small data size. Note that \autoref{xcomponent} indicates the summation over all samples that satisfies cutoff of within 3 standard deviations of the average pairwise distance. This cutoff can be reduced for faster computation. However, it reduces the overall accuracy. 

For CCP, because clustered features are projected independently, each reduced dimension can be computed independently using the parallel architecture. Similar parallel computations can be applied to different samples. Therefore, CCP can be further accelerated by using parallel and graphics processing unit (GPU) algorithms in practical applications.

\section{Concluding remarks}

Like other dimensionality reduction algorithms, CCP has its advantage and disadvantages. First, CCP is a unique data-domain method and its features are highly interpretable. Because CCP partitions feature into clusters according to some metric, such as covariance distance or correlation distance, features with high correlation will perform better. One limitation for many methods relying on matrix diagonalization is that pairwise distance computation can encounter the ``curse of dimensionality'', where distance computation of high dimensional data could become unreliable. By clustering features, CCP can more reliably compute distances because the dimension in each cluster will be much lower. Moreover, CCP performs better for data with a large number of features, such as TCGA-PANCAN, Coil-20, and Coil-100datasets. Therefore, CCP is suitable for the dimensionality reduction of data with relatively large intrinsic dimensions, for which many other popular methods may not work well.   

However, for datasets with a smaller number of features, CCP may not be as good as other methods. In this case, dimensionality reduction is unnecessary anyway. Also, we noticed that CCP might not be as good as UMAP and some other frequency-domain methods for extremely low final dimensions, say $N=2$ or 3.   

In addition to doing well for data having moderately large intrinsic dimensions, CCP allows embedding for streamlined datasets, such as molecular dynamics generated transient data. We have shown that CCP is stable under subsampling, which enables users to optimize the CCP model with a small portion of initial data, and allows subsequent data to be embedded with the initial set. We noticed that dimensionality reduction algorithms that rely on matrix diagonalization have instability when dealing with streamlined data. 

Because CCP does not compute the nearest neighbors graph and does not diagonalize, a traditional 2D plot does not give a meaningful visualization. However, each dimension of CCP is computed by projecting the partitioned features. Hence we can easily interpret each dimension of CCP. In tree-based classification algorithms, such as random forest and gradient boost decision trees, feature importance can be computed for each feature component, which gives a rank on how much impact each component has on the classification. For CCP, feature importance may be interpreted as how meaningful a set of highly correlated features is in the classification.

CCP can be further optimized in various ways. It allows a wide variety of alternative data-domain embedding strategies in each of its two steps: clustering and projection. For example, in the clustering step, one might select alternative distance metrics, clustering algorithms, and loss functions to optimize feature vector partition for a given dataset.  In the project step, one might choose alternative distance metrics based on Riemannian geometry or statistical theories and select alternative projections based on linear/nonlinear, orthogonal/non-orthogonal, and Grassmannian considerations.    

A wide variety of multistep  dimensionality reduction methods can be developed. Unlike frequency-domain dimensionality reduction techniques, CCP renders a data-domain representation of the original high-dimensional data. Therefore, the resulting low-dimensional data can be reused as an input for a Secondary Dimensionality Reduction (SDR) with a frequency-domain technique to achieve specific goals. For example, one can use CCP as an initializer for local methods to capture global patterns \cite{kobak2021initialization}. The combination of CCP with UMAP and t-SNE, called  CCP-UMAP and CCP-t-SNE, respectively, may generate better 2D visualizations for datasets with global structures. Additionally, for real-world problems, better accuracy is always desirable. New hybrid methods, such as three-step  CCP-UMAP and CCP-Isomap, may achieve better dimensionality reduction performance for clustering, classification, and regression. 

Finally,  the R-S scores, R index, S index, R-S disparity, and R-S index introduced in this work can be used for general-purpose data visualization and analysis. The shape of data and persistent Laplacian discussed in this work offer new geometric, topological, and spectral tools for data analysis and visualization.

\section*{Server availability} 
The CCP online server is available at \url{https://weilab.math.msu.edu/CCP/}. 

\section*{Acknowledgment}
This work was supported in part by NIH grants  R01GM126189 and  R01AI164266, NSF grants DMS-2052983,  DMS-1761320, and IIS-1900473,  NASA grant 80NSSC21M0023,    MSU Foundation,  Bristol-Myers Squibb 65109, and Pfizer. The computational assistance from MSU High Performance Computing Center (HPCC) is acknowledged.

\hspace{3cm}

\bibliographystyle{abbrv}

\end{document}